\documentclass[accepted, mathfont=newtx]{uai2025}
                        
\usepackage{natbib}

\usepackage[american]{babel}
\usepackage[utf8]{inputenc} 
\usepackage[T1]{fontenc}    
\usepackage{hyperref}       
\usepackage{url}            
\usepackage{booktabs}       
\usepackage{amsfonts}       
\usepackage{nicefrac}       
\usepackage{microtype}      
\usepackage{xcolor}         
\usepackage[font=footnotesize]{subfig}
\usepackage[inkscapearea=page]{svg}
\usepackage{amsmath} 
\usepackage{mathtools}
\usepackage{algorithm}
\usepackage{algpseudocode}
\usepackage{dsfont}
\usepackage{rotating}
\usepackage{wrapfig}
\usepackage{adjustbox}
\usepackage{xspace}
\usepackage{makecell}
\usepackage{tikz}
\usepackage[capitalise, nameinlink]{cleveref}
\usepackage{fancyhdr}
\fancypagestyle{firstpage}{%
  \fancyhf{}

  \fancyfoot[C]{\footnotesize
    \textit{Proceedings of the 41$^{st}$ Conference on Uncertainty in Artificial Intelligence} (UAI 2025), PMLR 244:398--420.}%
}

  \renewcommand{\a}{\alpha}
  \renewcommand{\b}{\beta}
  \renewcommand{\d}{\delta}

  \renewcommand{\l}{\lambda}
  \renewcommand{\L}{\Lambda}
    \newcommand{\m}{\mu}
    
    \newcommand{\N}{\nabla}
  \renewcommand{\k}{\kappa}
  \renewcommand{\o}{\omega}
  \renewcommand{\O}{\Omega}

  \renewcommand{\r}{\varrho}
    \newcommand{\s}{\sigma}
  
  \renewcommand{\t}{\theta}

    \newcommand{\BB}{{\cal B}}

    \newcommand{\FF}{{\cal F}}
    
    \newcommand{\HH}{{\cal H}}

    \newcommand{\NN}{{\cal N}}

  \renewcommand{\SS}{{\cal S}}

  \newcommand{\NNN}{{\mathbb{N}}}

  \newcommand{\RRR}{{\mathbb{R}}}

  \newcommand{\VVV}{{\mathbb{V}}}

  \newcommand{\vx}{\mathbf{x}}
  \newcommand{\vy}{\mathbf{y}}
  \newcommand{\vC}{\mathbf{C}}
  \newcommand{\vmu}{\boldsymbol{\mu}}
  \newcommand{\vxi}{\boldsymbol{\xi}}
  \newcommand{\vp}{\mathbf{p}}

  \renewcommand{\[}{\Big[}

  \renewcommand{\|}{\,|\,}
  
  \renewcommand{\=}{\!=\!}

  \newcommand{\Id}{{\rm\bf I}}

  \newcommand{\tsum}{\textstyle\sum}

  \renewcommand{\vec}{\boldsymbol}

  \providecommand{\href}[2]{{\color{blue}USE PDFLATEX!}}
  \providecommand{\url}[2]{\href{#1}{{\color{blue}#2}}}
  \newcommand{\anchor}[2]{\begin{picture}(0,0)\put(#1){#2}\end{picture}}

  \newcommand{\fullhide}[1]{\message{^^JHIDE--Warning (hidden)!^^J}}
  \newcommand{\mytitle}[1]{\title{#1}\newcommand{\thetitle}{#1}}
  \newcommand{\header}{\begin{document}\mytitle\cleardefs}
  \newcommand{\contents}{{\tableofcontents}\renewcommand{\contents}{}}
  \newcommand{\footer}{\small\bibliography{marc,bibs}\end{document}}
  \newcommand{\widepaper}{\usepackage{geometry}\geometry{a4paper,hdivide={25mm,*,25mm},vdivide={25mm,*,25mm}}}
  \newcommand{\moviex}[2]{\movie[externalviewer]{#1}{#2}} 
  \newcommand{\rbox}[1]{\fboxrule2mm\fcolorbox[rgb]{1,.85,.85}{1,.85,.85}{#1}}
  \newcommand{\mpage}[2]{{\begin{minipage}{#1\columnwidth}#2\end{minipage}}}
  \newcommand{\redbox}[2]{\fboxrule1mm\fcolorbox[rgb]{1,.7,.7}{1,.7,.7}{\begin{minipage}{#1\columnwidth}\center#2\end{minipage}}}
  \newcommand{\onecol}[2]{
    \begin{minipage}[c]{#1\columnwidth}#2\end{minipage}
    }

\newcommand{\norm}[1]{|\!|#1|\!|}

\newcommand{\Jwi}{J^\sharp_W}
\newcommand{\THi}{T^\sharp_H}
\newcommand{\Jci}{J^\natural_C}
\newcommand{\hJi}{{\bar J}^\sharp}
\renewcommand{\|}{\,|\,}
\renewcommand{\=}{\!=\!}
\newcommand{\1}{{\myminus1}}
\newcommand{\2}{{\myminus2}}
\newcommand{\3}{{\myminus3}}
\newcommand{\po}{{\myplus1}}
\newcommand{\pt}{{\myplus2}}
\newcommand{\T}{{\!\top\!}}
\newcommand{\xT}{{\underline x}}
\newcommand{\uT}{{\underline u}}
\newcommand{\zT}{{\underline z}}
\newcommand{\Sum}{\textstyle\sum}
\newcommand{\Int}{\textstyle\int}
\newcommand{\Prod}{\textstyle\prod}

\newcommand{\old}{{\text{old}}}
\newcommand{\new}{{\text{new}}}
\newcommand{\MAP}{{\text{MAP}}}
\newcommand{\ML}{{\text{ML}}}

\newcommand{\redArrow}{\quad\anchor{0,-1}{\includegraphics[scale=.5]{figs/redArrow}}}
\newcommand{\pub}[1]{{\color{green}\helvetica{8}{1.3}{m}{n}#1\\}}
\newcommand{\KL}[2]{\opKL\big(#1\,\big|\!\big|\,#2\big)} 

\renewcommand{\show}[2][.7]{\centerline{\includegraphics[width=#1\columnwidth]{#2}}}
\newcommand{\showh}[2][.8]{\includegraphics[width=#1\columnwidth]{#2}}
\newcommand{\shows}[2][.8]{\centerline{\includegraphics[scale=#1]{#2}}}
\newcommand{\showhs}[2][.8]{\includegraphics[scale=#1]{#2}}
\newcommand{\mov}[2]{\movie[externalviewer]{{\color{blue}\small #1}}{movies/#2}}
\newcommand{\movex}[2]{\movie[externalviewer]{#1}{#2}} 
\newcommand{\movh}[3][autostart,loop]{
\movie[#1]{\showh[#2]{#3.jpg}}{#3.mp4}%
\movie[externalviewer]{$\circ$}{#3.mp4}
}
\newcommand{\movc}[3][autostart,loop]{\centerline{\movh[#1]{#2}{#3}}}
\newcommand{\mymovie}[3][]{\centerline{\movie[autostart,loop,#1]{\includegraphics[#1]{#2}}{#3}}}

\newcommand{\cen}[1]{\centerline{#1}}

\newcommand{\citing}[1]{
{\color{citcol}\ttiny#1\par}
}

\newcommand{\cit}[3]{
\par\smallskip
{\color{citcol}\ttiny #1: \emph{#2}. #3 \par}
}

\newcommand{\citurl}[4]{
\par\smallskip
{\color{citcol}\ttiny #1: \protect{\href{#4}{\color{blue}{#2.}}} #3 \par}
}

\newcommand{\cito}[3]{
\par\smallskip
{\color{bluecol}\ttiny #1: \emph{#2}. #3 \par}
}

\newcommand{\redoMacrosInProof}{
  \renewcommand{\d}{\delta}
  \renewcommand{\=}{\!=\!}
}

\newcommand{\pdflatex}{
  \definecolor{bluecol}{rgb}{0,0,.5}
  \usepackage[
    colorlinks,
    urlcolor=bluecol,
    citecolor=black,
    linkcolor=bluecol,
    pdfborder={0 0 0},
    pdfpagemode=UseOutlines, 
    pdfauthor={Marc Toussaint}
  ]{hyperref}
  \DeclareGraphicsExtensions{.pdf,.png,.jpg,.eps}
  \renewcommand{\r}{\varrho}
  \renewcommand{\l}{\lambda}
  \renewcommand{\L}{\Lambda}
  \renewcommand{\s}{\sigma}
  \renewcommand{\b}{\beta}
  \renewcommand{\d}{\delta}
  \renewcommand{\k}{\kappa}
  \renewcommand{\t}{\theta}
  \renewcommand{\O}{\Omega}
  \renewcommand{\o}{\omega}
  \renewcommand{\SS}{{\cal S}}
  \renewcommand{\=}{\!=\!}
}

\newcommand{\mypause}{\pause}
\newcommand{\defi}[1]{\textbf{#1}}
\newcommand{\red}[1]{\emph{\color{red}#1}}
\newcommand{\pos}{{\textsf{pos}}}
\newcommand{\eff}{{\textsf{eff}}}
\newcommand{\rot}{{\textsf{rot}}}
\newcommand{\veC}{{\textsf{vec}}}
\newcommand{\quat}{{\textsf{quat}}}
\newcommand{\col}{{\textsf{col}}}
\newcommand{\de}[2]{\frac{\partial #1}{\partial #2}}
\newcommand{\target}{{\text{target}}}
\newcommand{\near}{{\text{near}}}
\newcommand{\qfree}{Q_{\text{free}}}
\renewcommand{\vec}{\boldsymbol}
\newcommand{\lft}{\text{left}}
\newcommand{\rgh}{\text{right}}
\newcommand{\prev}{{\text{prev}}}
\newcommand{\TR}[2]{T_{{#1}\to{#2}}}
\newcommand{\RO}[2]{R_{{#1}\to{#2}}}
\newcommand{\liter}{\helvetica{8}{1.1}{m}{n}\parskip 1ex}
\newcommand{\Fc}{\color{green}F}
\newcommand{\muc}{\color{blue}\mu}
\newcommand{\Astar}{A$^*$}

\providecommand{\defn}[1]{\textbf{#1}}

\newcommand{\adec}{\r_\a^-}
\newcommand{\ainc}{\r_\a^+}
\newcommand{\ldec}{\r_\l^-}
\newcommand{\linc}{\r_\l^+}
\newcommand{\minc}{\r_\m^+}
\newcommand{\mdec}{\r_\m^-}
\newcommand{\step}{{\delta}}
\newcommand{\lsstop}{\r_{\text{ls}}}
\newcommand{\stepmax}{\step_{\text{max}}}
\newcommand{\Min}{\text{min}}
\newcommand{\Max}{\text{max}}
\DeclareMathOperator*{\argmin}{arg\,min}
\DeclareMathOperator*{\EEE}{\mathbb{E}}

\definecolor{amaranth}{rgb}{0.9, 0.17, 0.31}
\definecolor{BLUE}{rgb}{0,0,1}
\definecolor{algblue}{HTML}{125DA6} 
\newcommand{\rred}[1]{\textcolor{amaranth}{#1}}
\newcommand{\bblue}[1]{\textcolor{algblue}{#1}}

\newcommand{\wrt}{w.r.t.\@\xspace}
\newcommand{\cma}{\Delta \vx_{\textsc{cma}}}

\title{Stein Variational Evolution Strategies}

\author[1]{Cornelius V.\ Braun}
\author[2]{Robert T.\ Lange}
\author[1]{Marc Toussaint}
\affil[1]{%
    TU Berlin
}
\affil[2]{%
    Sakana AI
}

\begin{document}
\maketitle
\thispagestyle{firstpage}

\begin{abstract}
Efficient global optimization and sampling are fundamental challenges, particularly in fields such as robotics and reinforcement learning, where gradients may be unavailable or unreliable.
In this context, jointly optimizing multiple solutions is a promising approach to avoid local optima.
While Stein Variational Gradient Descent (SVGD) provides a powerful framework for sampling diverse solutions, its reliance on first-order information limits its applicability to differentiable objectives. 
Existing gradient-free SVGD variants often suffer from slow convergence and poor scalability.
To improve gradient-free sampling and optimization, we propose Stein Variational CMA-ES, a novel gradient-free SVGD-like method that combines the efficiency of evolution strategies with SVGD-based repulsion forces.
We perform an extensive empirical evaluation across several domains, which shows that the integration of the ES update in SVGD significantly improves the performance on multiple challenging benchmark problems.
Our findings establish SV-CMA-ES as a scalable method for zero-order sampling and blackbox optimization, bridging the gap between SVGD and evolution strategies.
\end{abstract}

\begin{figure*}[!ht]
    \centering
    \begin{minipage}[t]{0.29\textwidth}
        \centering
        \large \makecell[t]{ \textbf{Stein Variational CMA-ES}\\ \textbf{(SV-CMA-ES)}: Multiple ES\\ Populations Optimize Fitness \\and Diversity }\\[.5em]
        \includegraphics[page=6, width=\linewidth, trim={0cm 6.5cm 33.5cm 3cm}, clip]{imgs/sves_plot.pdf}\\[0.25em]
        \includegraphics[page=6, width=.875\linewidth, trim={0cm 2cm 33.5cm 16.5cm}, clip]{imgs/sves_plot.pdf}
    \end{minipage}\hspace{1mm}
    \begin{minipage}[t]{0.29\textwidth}
        \centering
        \large \textbf{Quantitative Performance} \\
        Density Approximation\\[.1em]
        \hspace{-2mm} \includegraphics[width=0.65\linewidth]{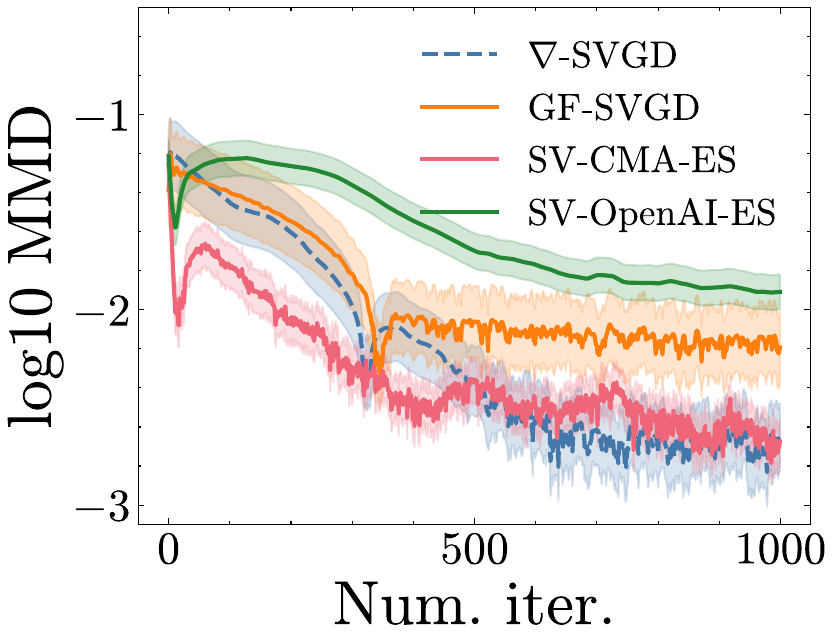}\\
        Reinforcement Learning\\[.1em]
        \hspace{-2mm} \includegraphics[width=0.65\linewidth]{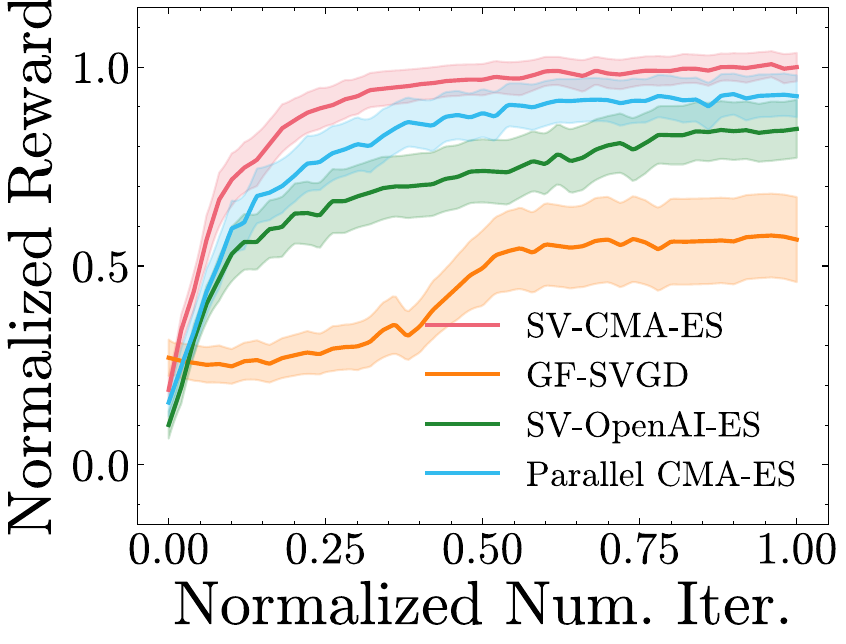}
    \end{minipage}\hspace{1mm}
    \begin{minipage}[t]{0.29\textwidth}
        \centering
        \large \makecell[t]{\textbf{Qualitative Performance}\\ Solution Quality \& Diversity}\\[0.5em]
        \begin{tabular}{cc}
        \includegraphics[width=0.35\linewidth]{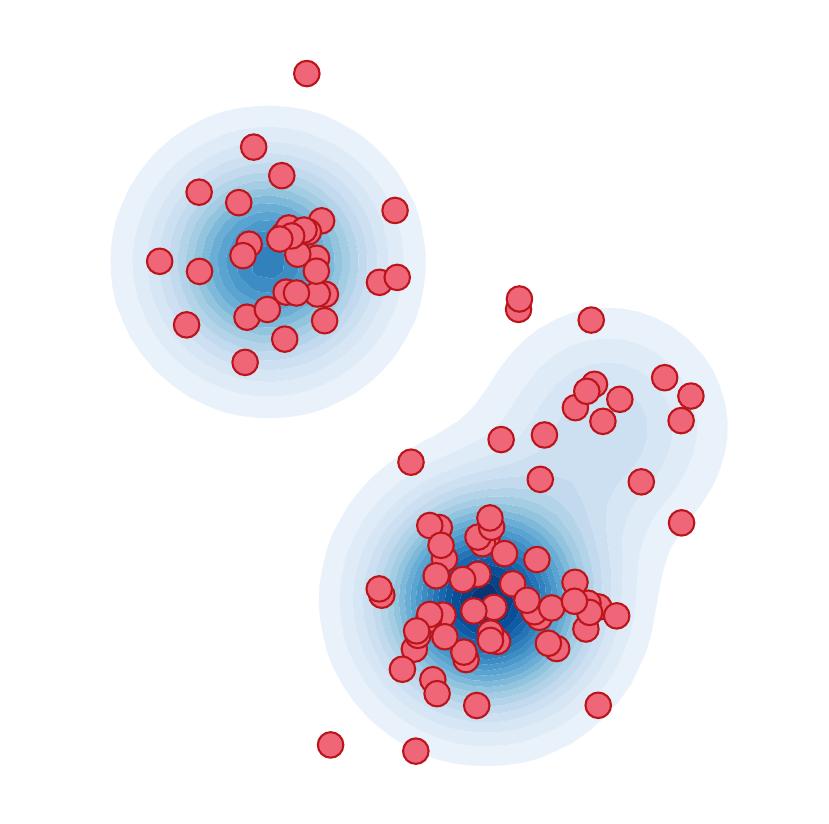}
        & 
        \includegraphics[width=0.35\linewidth]{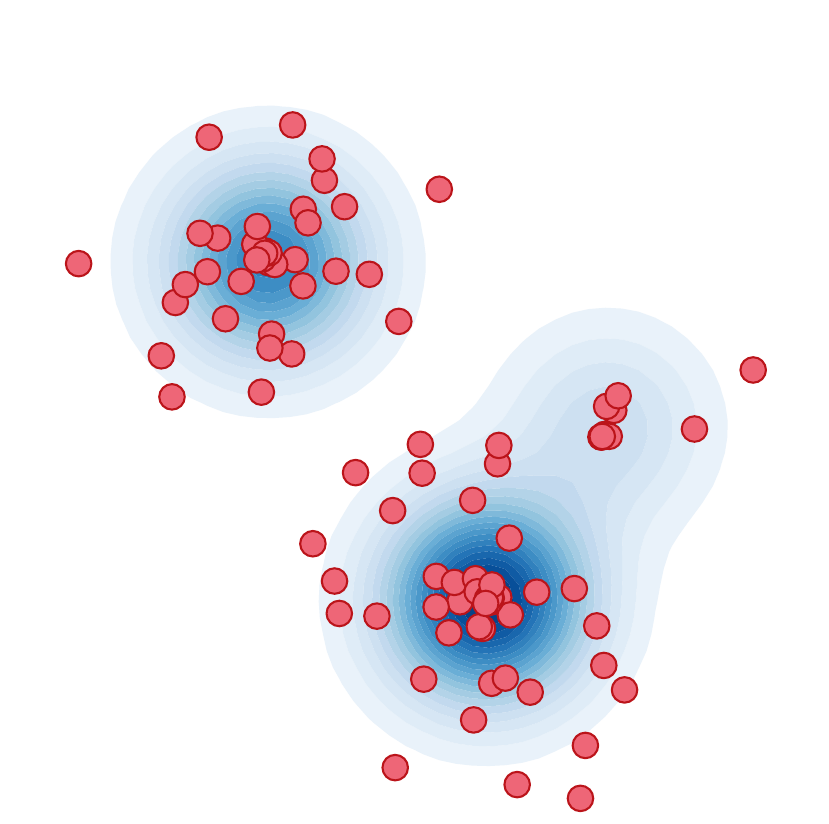}
        \\
        \small Ground Truth & \small SV-CMA-ES\\[0.5em]
        \includegraphics[width=0.35\linewidth]{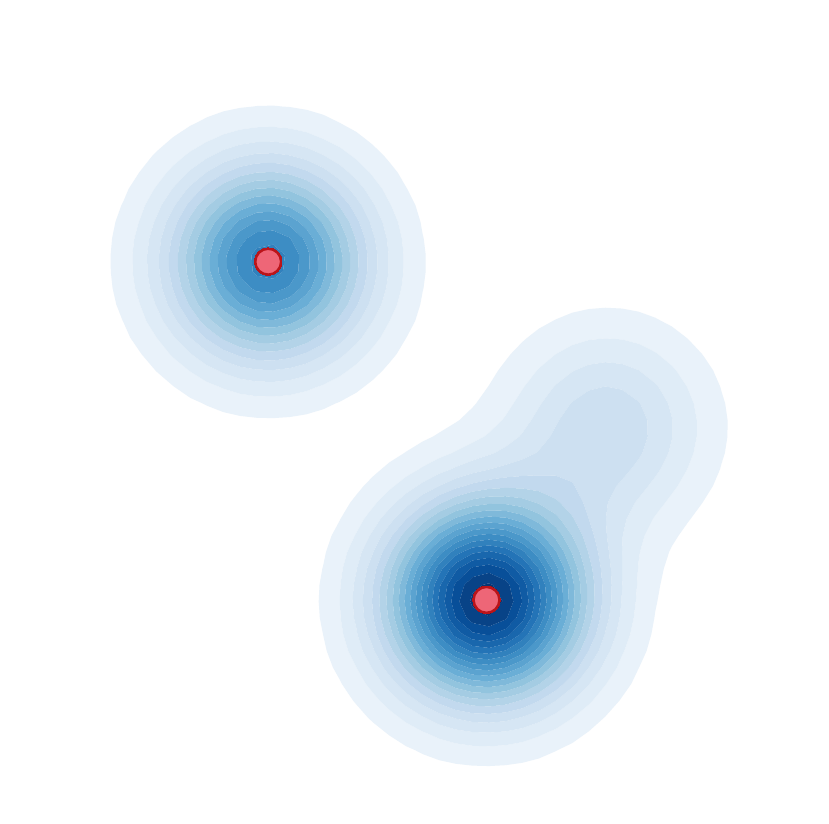}
        &
        \includegraphics[width=0.35\linewidth]{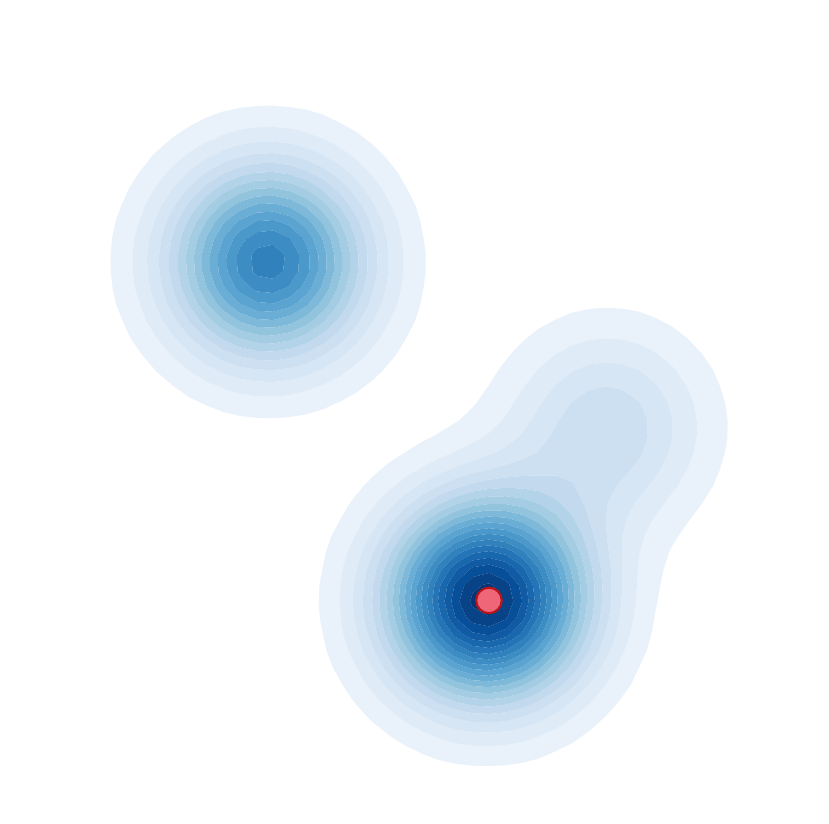}\\
        \makecell{\small Multi-Pop. CMA-ES \\[-1mm] \small (kernel-free)} & \small CMA-ES\\
        \end{tabular}
    \end{minipage}

    \caption{\textbf{Left}: Illustration of Stein Variational CMA-ES. 
    Multiple ES search distributions are updated in parallel, similar to the SVGD step. \textbf{Middle}: Quantitative comparison of different methods for sampling and RL control tasks. SV-CMA-ES obtains higher quality solutions than existing gradient-free SVGD-based approaches. \textbf{Right}: Qualitative comparison of different CMA-ES-based methods unveils that SV-CMA-ES generates more diverse samples than other methods. The full experimental details can be found in \cref{secExp}.
    }
    \label{fig:sves-overview}
\end{figure*}

\section{Introduction}\label{secIntro}
Many optimization problems -- such as neural network parameter search -- involve highly non-convex objective functions, which makes the optimization process very sensitive to its initialization \citep{sullivan2022cliff, li2018visualizing}.
Thus, these hard optimization problems are commonly approached by generating multiple solution candidates from which the best is selected \citep{toussaint2024nlp, parker2020ridge}.
This allows to frame the task of minimizing a function $f: \RRR^d \to \RRR$ as an approximate inference problem which can be formulated as follows:
$$\min_{q} D (q~ \Vert~ p) \qquad p(\vx) = \frac{e^{-f(\vx)}}{Z},$$
where the normalization constant $Z = \int_{\RRR^d}e^{-f(\vx)} \text{d}\vx$ is typically intractable, $p$ and $q$ are probability distributions supported on $\RRR^d$, and $D$ is a suitable divergence, such as the Kullback-Leibler (KL) divergence.

Stein Variational Gradient Descent (SVGD) is a powerful algorithm to solve this optimization problem through iteratively updating a particle set \citep{liu2016stein}.
As the approach is non-parametric and does not require the lengthy burn-in periods of Markov chain Monte Carlo (MCMC) methods \citep{andrieu2003introduction}, it is a computationally efficient method to approximate complex distributions.
Due to these properties, SVGD is an increasingly popular first-order method for sampling and non-convex optimization \citep{zhang2019bayesian, maken2021stein, pavlasek2023ready}.

Unfortunately, the reliance of SVGD on the score function limits its applicability to differentiable objectives.
In many real-world problems -- such as robotics and chemistry -- however, the energy function $f$ may not yield reliable gradients or be non-differentiable altogether \citep{lambert2020stein, englert2018learning, maus2023discovering}. 
To facilitate \textit{gradient-free} Stein variational inference, prior work introduced a zero-order version of SVGD that uses analytical gradients from a surrogate distribution \citep[GF-SVGD]{han2018stein}.
While the algorithm provably minimizes the KL divergence, fitting the surrogate to the objective function is challenging in practice, especially in higher dimensions (cf. Fig.~\ref{fig:samples}).
Alternatively, other works used simple Monte Carlo (MC) gradients in the SVGD update \citep{liu2017stein, lambert2020stein, lee2023stamp}.
Again, this approach comes with limitations as the MC step estimate has high variance, which often leads to noisy updates and thus poor computational efficiency.

To address the aforementioned shortcomings of existing gradient-free SVGD methods, we propose a novel approach, \textit{Stein Variational CMA-ES (SV-CMA-ES)}.
Our method bridges the fields of Evolution Strategies (ES) and distribution approximation by updating multiple ES search distributions in parallel.
The idea of SV-CMA-ES is to perform the distribution updates in a coordinated manner using a kernel-based repulsion term, which ensures an inter-population diversity similar to that in SVGD.
We motivate our work based on prior results that established ES as a competitive alternative to gradient-based optimization algorithms, achieving higher performance and robustness on difficult objectives due to their inherent exploration capabilities \citep{salimans2017evolution, wierstra2014natural}.
In particular, the Covariance Matrix Adaptation Evolution Strategy \citep[CMA-ES]{hansen2001} is one of the most popular ES across many domains \citep{hansen2010comparing, jankowski2023vp}, due to its adaptive and efficient search process, which leverages a dynamic step-size adaptation mechanism to increase convergence speeds \citep{akimoto2012theoretical}.

We evaluate our proposed approach on a wide range of challenging problems from multiple domains, such as robot trajectory optimization and reinforcement learning.
Our experimental results demonstrate that SV-CMA-ES improves considerably over existing gradient-free SVGD approaches.
Fig.~\ref{fig:sves-overview} summarizes our findings.
Not only can our method be used to sample from challenging densities efficiently, but also as a blackbox optimizer on non-convex objectives.
We outline our contributions as follows:
\begin{enumerate}
    \item We introduce a novel zero-order method for diverse sampling and global optimization that combines ideas of SVGD with gradient-free ES, thus bypassing the need for a surrogate distribution required by previous gradient-free SVGD approaches (Section \ref{secMain}).

    \item We validate our method, SV-CMA-ES, on a range of problems and demonstrate that it improves over prior gradient-free SVGD approaches in sampling and optimization tasks (Fig. \ref{fig:sves-overview} middle; Sec.\ \ref{secExpSynth}-\ref{secExpRl}).
    
    \item We show that our presented method improves over prior CMA-ES-based methods because it combines the fast convergence rate of CMA-ES with the entropy-preserving optimization dynamics of SVGD (Fig. \ref{fig:sves-overview} right; Sec.\ \ref{secExpRl}). 
\end{enumerate}

\section{Preliminaries}\label{secPrelim}
\subsection{Stein Variational Gradient Descent}
Stein Variational Gradient Descent \citep[SVGD]{liu2016stein} is a non-parametric inference algorithm that approximates a target distribution with a set of $\varrho \in \NNN^+$ particles $X = \{\vx_i\}^\varrho_{i=1}$ as $q(\vx) = \sum_{\vx_i \in X} \delta (\vx - \vx_i) / \varrho$, where we use $\delta (\cdot)$ to denote the Dirac delta function.
Given an initial set of particles, the goal is to determine an optimal particle transformation $\phi^*: \RRR^d \to \RRR^d$ that maximally decreases the KL divergence $D_{\text{KL}} (q~ \Vert~ p)$:
\begin{align}
    \begin{split}
        &\vx_i \gets \vx_i + \epsilon \phi^*(\vx_i), \quad \forall \vx_i \in X \label{eq:svgd1}\\
        \text{s.t.\ } &\phi^* = \argmin_{\phi \in \FF} \left\{ \left. \frac{d}{d\epsilon} D_{\text{KL}}(q_{[\epsilon \phi]}~ \Vert~ p) \right|_{\epsilon = 0} \right\},
    \end{split}
\end{align}
where $\epsilon \in \RRR$ is a sufficiently small step-size, $q_{[\epsilon \phi]}$ denotes the distribution of the updated particles, and $\FF$ is a set of candidate transformations.

The main result by \citet{liu2016stein} is the derivation of a closed form solution to this optimization problem.
By choosing $\FF$ as the unit sphere $\BB_k$ in a vector-valued reproducing kernel Hilbert space $\HH_k^d$, i.e.\, $\FF_k = \BB_k  \{\phi \in \HH_k^d: \lVert \phi \rVert_{\HH_k^d} \leq 1\}$, with its kernel function $k(\cdot, \cdot): \RRR^d \times \RRR^d \to \RRR$, the authors show that the solution to Eq.~\eqref{eq:svgd1} is:
\begin{align}
    \phi^*_k (\cdot) \propto \EEE_{\vx \sim q} \bigl[ \underbrace{\nabla_\vx \log p(\vx)k(\vx, \cdot)}_{\text{driving force}} + \underbrace{\nabla_\vx k(\vx, \cdot)}_{\text{repulsive force}} \bigr].\label{eq:svgd2}
\end{align}
This result can be used to update the particle set iteratively using Eq.~\eqref{eq:svgd1} and \eqref{eq:svgd2}, where the expectation is estimated via MC approximation over the entire particle set $X$.
Intuitively, the particle update balances likelihood maximization and particle repulsion:
the first term drives particles toward regions of higher probability, while the second term counteracts this by repulsing particles based on the kernel gradient \citep{d2021annealed, ba2021understanding}.

Because vanilla SVGD is prone to the initialization of particles and mode collapse \citep{zhuo2018message, ba2021understanding, zhang2020stochastic}, prior work proposed \textit{Annealed SVGD} \citep{liu2017stein, d2021annealed}.
This extension of SVGD, reweighs the terms in the update based on the optimization progress \citep{d2021annealed}. 
Given the timestep-dependent temperature parameter $\gamma(t) \in \RRR$, the annealed update is:
\begin{align}
    \phi^*_k (\cdot) \propto \EEE_{\vx \sim q} \left[ \nabla_\vx \log p(\vx)k(\vx, \cdot) + \gamma(t) \nabla_\vx k(\vx, \cdot) \right].\label{eq:asvgd}
\end{align}

\subsection{Covariance Matrix Adaptation Evolution Strategy}
The Covariance Matrix Adaptation Evolution Strategy \citep[CMA-ES]{hansen2001} is one of the most popular ES algorithms. 
We therefore choose it as the starting point for our ES-based SVGD method.
The core idea of the CMA-ES algorithm is to iteratively optimize the parameters of a Gaussian search distribution $\NN (\vx, \sigma^2 \vC)$ from which the candidate solutions are sampled.
While the algorithmic intuition of CMA-ES is similar to MC gradient approaches \citep{salimans2017evolution}, CMA-ES updates the search distribution following natural gradient steps, which has been shown to produce more efficient steps than standard gradient descent on multiple problems \citep{martens2020new, akimoto2012theoretical, glasmachers2010exponential}.

We note that our notation in the following deviates from the default notation in the ES literature, as some of its variable names are typically associated with a different meaning compared to the variational inference (VI) literature. 
For instance, $\mu$ commonly is the symbol for the mean of a Gaussian in VI literature, while it refers to the number of elites in CMA-ES.
To improve clarity, we thus use the variable $n$ to denote the size of a sampled CMA-ES population and $m$ for the number of selected elites.
In our notation, CMA-ES is therefore an $(m, n)$ strategy.
The CMA-ES algorithm relies on multiple hyperparameters which we fix to the default values from \citet{hansen2016cma}.
For completeness, we include the definitions of these variables -- $w_i, \alpha_1, \alpha_m, \alpha_\sigma, h_\sigma, d(h_\sigma)$ and $\bar{w}_i$ -- in the \hyperref[secAppendix]{Appendix}.
Further, we slightly overload notation by using $\vp$ to denote the evolution path updates following \citet{hansen2016cma} (unlike pdf's which we denote by $p$).

Given a population of $n$ candidate samples $\vxi_i \sim \NN (\vx, \sigma^2 \vC)$, each iteration of CMA-ES updates the search parameters as follows.
First, the samples $\vxi_i$ are evaluated and ranked by their fitness $f(\vxi_i)$ in ascending order.
To simplify notation, we assume ranked solutions in the following, i.e., we assume that index $i < j \rightarrow f(\vxi_i) \leq f(\vxi_j)$.
This allows to assign each sample to a mutation weight $w_i$, where weights for better solutions are higher.
For details on the exact computation of the weights, we refer to \citet{hansen2016cma} and to Appendix \ref{secAlgSuppl} of our work.
The mean of the search distribution is then updated by mutating the $m \leq n$ best samples from the current generation of candidates:
\begin{align}
    &\vx \gets \vx + \cma \label{eq:cma-mu}\\ 
    \intertext{where }
    &\cma = \sigma \tsum_{i=1}^m w_i \vy_i, \text{ and } \vy_i = (\vxi_i - \vx) / \sigma. \label{eq:cma-step}
\end{align}
Next, the parameters of the search distribution are updated.
First, the step-size $\sigma$ is updated based on the history of prior steps.
Given
\begin{align}
    &m_{\text{eff}} = (\Sigma_{i=1}^m w_i^2)^{-1}, \text{ and }\label{eq:mueff}\\
    &\vp_{\sigma} \leftarrow (1 - \alpha_{\sigma}) \vp_{\sigma} + \sqrt{\alpha_\sigma (2 - \alpha_\sigma)~ m_{\text{eff}}}~\vC^{-\frac{1}{2}} \cma/\sigma \label{eq:cma_evo_trace},\\
\intertext{we define}
    &\sigma \leftarrow \sigma \times \exp\Bigl(\frac{\alpha_{\sigma}}{d_{\sigma}} \Bigl(\frac{\lVert \vp_{\sigma}\rVert}{\EEE \lVert \NN (0, \Id) \rVert} - 1\Bigr)\Bigr),\label{eq:cma-sig}
\end{align}
where $\vp_{\sigma}$ is a moving average over the optimization steps, which unprojects the steps using $C^{-1/2}/\sigma$, so the resulting vector follows a standard normal.
Thus, Eq.~\eqref{eq:cma-sig} automatically adapts the step-size based on the expected length of steps, similar to momentum-based optimizers \citep{kingma2014adam, nesterov1983method}, with the hyperparameters $\alpha_\sigma$ and $d_\sigma$ governing the rate of the step-size changes.
Finally, the covariance $\vC$ is updated based on the covariance of the previous steps and current population fitness values:
\begin{align}
    \begin{split}
       &\vC \leftarrow (1 + \alpha_1 d(h_\sigma)- \alpha_1 - \alpha_m \tsum_{j=1}^n w_j)\vC\\
       &\qquad + \alpha_1 \vp_c {\vp_c}^T + \alpha_m \tsum_{i=1}^n \bar{w}_i \vy_i \vy_i^T,
    \end{split}\label{eq:cma-cov}
\intertext{with } 
    &\vp_c \leftarrow (1 - \alpha_c) \vp_c + h_{\sigma} \sqrt{\alpha_c (2-\alpha_c) m_{\text{eff}}}~ \cma/\sigma.\label{eq:cma-pc}
\end{align}
In words, the covariance update performs smoothing over the optimization path to update $\vC$ based on the within- and between-step covariance of well-performing solutions.
Thus, Eq.~\eqref{eq:cma-cov} scales $\vC$ along directions of successful steps to make the search converge faster.

\section{A Stein Variational Evolution Strategy}\label{secMain}
This section introduces a novel framework of using a multi-population ES for efficient discovery of multiple high-quality solutions to an optimization problem. 
The idea of this work is to represent each SVGD particle by the mean of an ES search distribution and use the estimated steps of the ES algorithm as the \textit{driving force} in the SVGD particle update. 
Hence, our approach exploits the CMA-ES step-size adaptation mechanism to make gradient-free inference more efficient. 
Intuitively, the reformulated update permits larger particle updates, similar to momentum, especially in flat regions of the target.
Since ES are easily parallelizable on modern GPUs \citep{lange2023evosax, tang2022evojax}, this approach comes at a small additional runtime cost.
In the following, we use $\varrho \in \NNN^+$ to refer to the number of ES search distributions, $n\in \NNN^+$ to denote the size of each sampled population, and $m\in \NNN^+$ for the number of elite samples.
This amounts to a total population size of $\varrho n$ for ES-based algorithms.

Based on the SVGD update in Eq.~\eqref{eq:svgd2} and the CMA-ES update of the search distribution mean in Eq.~\eqref{eq:cma-mu}, we now define \textit{Stein Variational CMA-ES (SV-CMA-ES)}.
The full algorithm is listed in Algorithm \ref{alg:sv-cmaes}.
SV-CMA-ES is a multi-population version of CMA-ES, where $\varrho$ search distributions are updated in parallel, each representing an SVGD particle $\vx_i$ via their distribution mean.
In other words, for each particle, there is a corresponding Gaussian search distribution that is centered at the particle and parametrized as $\NN (\vx_i, \sigma_i^2 \vC_i)$.
Given the standard CMA-ES distribution update step $\cma$ from Eq.~\eqref{eq:cma-step} and a sampled population $\vxi_{ij} \sim \NN (\vx_i, \sigma_i^2 \vC_i)$, we propose the following SVGD-based update:
\begin{align}
    \vx_i &\gets \vx_i + \epsilon~\phi (\vx_i) \quad \text{with} \label{eq:sv-cma-m}\\
    \phi(\vx_i) &= \EEE_{\vx_j \sim q} \Bigl[ k(\vx_j, \vx_i)\Delta {\vx_j}_{\textsc{cma}} + \N_{\vx_j} k(\vx_j, \vx_i) \Bigr]\nonumber\\
    = &\frac{1}{\varrho} \sum_{j=1}^\varrho \Biggl[ \underbrace{\Bigl[\sum_{\ell=1}^m w_{j\ell}(\vxi_{j\ell} - \vx_j)\Bigr] k(\vx_j, \vx_i)}_{\text{driving force}} + \underbrace{\vphantom{\Bigl[\sum_{k=1}^m\Bigr]} \N_{\vx_j} k(\vx_j, \vx_i)}_{\text{repulsive force}}\Biggr]\label{eq:sv-cma-step}
\end{align}
where we assume the same sorting by fitness in our sum as in vanilla CMA-ES and $w_{j\ell}$ are the sample weights that are computed based on the fitness values $f(\vxi_{j\ell})$ following \citet{hansen2016cma}.
Further, we use an additional step-size hyperparameter $\epsilon$ for notational consistency with SVGD, but we always fix it to $\epsilon = 1$.

Eq.~\eqref{eq:sv-cma-step} defines how to update each particle search distribution mean.
It now remains to define the remaining SV-CMA-ES parameter updates.
The original CMA-ES step-size update \eqref{eq:cma-sig} is based on the length of the distribution mean update step.
In the particle update in Eq.~\eqref{eq:sv-cma-m}, this quantity corresponds to the effective update step $\phi(\vx_i)$.
Given this particle shift, the smoothened step estimate $\vp_{\sigma_i}$ is computed analogously to the CMA-ES optimization path update in Eq.~\eqref{eq:cma_evo_trace}:
\begin{equation}
    \vp_{\sigma_i} \leftarrow (1 - \alpha_{\sigma}) \vp_{\sigma_i} + \sqrt{\alpha_\sigma (2 - \alpha_\sigma)~ m_{\text{eff}, i}}~\vC_i^{-\frac{1}{2}} \phi(\vx_i) / \sigma_i
\end{equation}
Using the same construction, we update $\vp_{c_i}$ based on $\phi(\vx_i)$, from which the covariance $\vC_i$ can be computed using \Cref{eq:cma-cov}.

\subsection{Practical considerations}
We now discuss some modifications to the algorithm that we found beneficial in practice.
As noted earlier, the update of the particle in Eq.~\eqref{eq:sv-cma-step} smoothens the gradient approximation across all particles.
As a result, the magnitude of the effective steps is reduced compared to standard CMA-ES.
Since CMA-ES reduces the step-size $\sigma$ automatically when small steps are taken, this may lead to premature convergence.
An example that illustrates this problem is a bimodal distribution with both modes far apart, such that for most particles $k(\vx, \vy)$ is close to zero for all pairs $\vx, \vy$ that are sampled from different modes.
In this scenario, the driving force term of the update corresponds to the vanilla CMA-ES update, scaled down by the factor of $1 / \varrho$.
Hence, the proposed steps in this scenario would shrink iteratively.
To address this issue, we propose the following simplified particle update:
\begin{equation}
    \begin{split}
        &\phi(\vx_i) = \frac{1}{\varrho} \sum_{j=1}^\varrho \Biggl[ \Bigl[\sum_{\ell=1}^m w_{i\ell}(\vxi_{i\ell} - \vx_i)\Bigr] + \N_{\vx_j} k(\vx_j, \vx_i)\Biggr].\\
    \end{split}\label{eq:sv-cma-final}
\end{equation}
This update uses only the particle $\vx_i$ to estimate the first term of the update, i.e., the driving force.
We note that this corresponds to a hybrid kernel SVGD setting \citep{d2021stein, macdonaldhybrid}, which uses two separate kernels to compute the repulsion and driving force terms: 
$\phi_{\text{hybrid}}(\vx_i) = \EEE_{\vx \sim q} \left[ \N_\vx f(\vx) k_1(\vx, \vx_i) + \N_\vx k_2(\vx, \vx_i) \right]$ if we choose $k_1(\vx, \vy) = n\mathds{1}(\vx = \vy)$.
This kernel can be approximated by an RBF kernel with small bandwidth $h \rightarrow 0$.

While the update in Eq.~\eqref{eq:sv-cma-final} does not possess the same capabilities of transporting particles ``along a necklace’’ as the vanilla SVGD update (cf. Fig.\ 1 of \citet{liu2016stein}), it has been noted that these SVGD capabilities play a limited role for practical problems in the first place \citep{d2021annealed}.
Instead, prior work proposed the annealed update in Eq.~\eqref{eq:asvgd} to transport the particles to regions of high density \citep{d2021annealed, liu2017stein}.
In practice, we observe that using the annealed version of the above update, i.e.,
\begin{align}
    \phi(\vx_i) &=\frac{1}{\varrho} \sum_{j=1}^\varrho \Biggl[ \Bigl[\sum_{\ell=1}^m w_{i\ell}(\vxi_{i\ell} - \vx_i)\Bigr] + \gamma(t) \N_{\vx_j} k(\vx_j, \vx_i)\Biggr]\nonumber\\
     &=  \sum_{\ell=1}^m w_{i\ell}(\vxi_{i\ell} - \vx_i) + \frac{\gamma(t)}{\varrho}\sum_{j=1}^\varrho \N_{\vx_j} k(\vx_j, \vx_i)\label{eq:asv-cma-final}
\end{align}
ensures sufficient mode coverage to efficiently sample from distributions.

The substitution of the score function with the CMA-ES step introduces a bias in comparison to the SVGD update in Eq.~\eqref{eq:asvgd}, meaning it does not strictly adhere to the canonical SVGD framework and does not inherit its robust convergence properties. 
Still, we find that, in practice, the update makes a useful tradeoff which combines the computational efficiency of CMA-ES with the particle set entropy preservation capabilities of SVGD.
We leave a more in-depth theoretical analysis of the algorithm for future work, and present our empirical findings in the subsequent section.
For an empirical convergence analysis, we refer to Appendix \ref{sec:app_convergence}.

\renewcommand{\algorithmicindent}{.75em}  
\begin{algorithm}[!ht]
    \caption{Stein Variational CMA-ES. Differences to the parallel CMA-ES algorithm are highlighted in \bblue{\textbf{blue}}.
    }
    \label{alg:sv-cmaes}
    
    \textbf{Input:} Kernel $k(\cdot, \cdot)$; num.\ particles $\varrho$; subpop.\ size $n$; num.\ elites $m$; elite weights $w_{i=1\dots m}$; learning rates $\epsilon, \alpha_\sigma, \alpha_1, \alpha_m, \alpha_c$; damping hyperparam.\ $d_\sigma$; dimension $d$; num.\ iterations $T$
    \begin{algorithmic}[1]
        \State Initialize population parameters $\vx_i, \sigma_i, \vC_i$ for particles $i = 1, \dots, \varrho$
        \For{iteration $t = 1, \dots, T$}
            \For{particle $i = 1, \dots, \varrho$}
                \State \textbf{Sample \& evaluate new population}:\vspace{-3mm}
                \begin{flalign*}
                    \hspace{1.5em}&\vxi_{ij} \sim \NN (\vx_i, \sigma_i^2 \vC_i)\text{, for $j = 1, \dots, n$}&\\
                    &\vy_{ij} = (\vxi_{ij} - \vx_i) / \sigma_i&
                \end{flalign*}
                
                \State \textbf{\bblue{Estimate gradient \& shift particle (Eq.~\ref{eq:asv-cma-final})}}:
                \Statex \hspace{1.3em} Sort samples by $f_{ij} = f(\vxi_{ij})$ in ascending order\vspace{-3mm}
                 \begin{flalign*}
                    \hspace{1.5em}&\bblue{
                    \phi(\vx_i) = \sum_{\ell=1}^m w_{i\ell}(\vxi_{i\ell} - \vx_i) + \frac{\gamma(t)}{\varrho}\sum_{j=1}^\varrho \N_{\vx_j} k(\vx_j, \vx_i) 
                    }\hspace{-.5em}&\\
                    &\bblue{\vx_i \gets \vx_i + \epsilon \phi(\vx_i)}&
                \end{flalign*}
        
                \State \textbf{Cumulative step-size adaptation}:\vspace{-2.5mm}
                \begin{flalign*}
                    \hspace{1.5em}&m_{\text{eff}, i} = (\tsum_{\ell=1}^m w_{i\ell}^2)^{-1}&\\
                    \begin{split}
                        &\vp_{\sigma_i} \gets (1 - \alpha_{\sigma}) \vp_{\sigma_i}\\
                        &\qquad + \sqrt{\alpha_\sigma (2 - \alpha_\sigma)~ m_{\text{eff}, i}}~\vC_i^{-\frac{1}{2}} \bblue{\phi(\vx_i) / \sigma_i}
                    \end{split}\\
                    &\sigma_i \gets \sigma_i \times \exp\Bigl(\frac{\alpha_{\sigma}}{d_{\sigma}} \Bigl(\frac{\lVert \vp_{\sigma_i}\rVert}{\EEE \lVert \NN (0, \Id) \rVert} - 1\Bigr)\Bigr)&
                \end{flalign*}
        
                \State \textbf{Covariance matrix adaptation}:\vspace{-3mm}
                \begin{flalign*}
                    \hspace{1.5em}&\bar{h} = \lVert \vp_{\sigma_i} \rVert / \sqrt{1 - (1-\alpha_\sigma)^{2(t+1)}}&\\
                    &h_{\sigma_i} = 1 \text{ if } \bar{h} < (1.4 + \tfrac{2}{d+1})\EEE \lVert \NN (0, \Id) \rVert \text{ else } 0 \hspace{-2em}&\\
                    &d(h_{\sigma_i}) = 1 \text{ if } \alpha_c (1 - h_\sigma)(2-\alpha_c) \leq 1 \text{ else } 0&\\
                    &\bar{w}_{ij} = w_{ij} \text{ if } w_{ij} \geq 0 \text{ else } d / \lVert \vC_i^{-\frac{1}{2}} \vy_{ij} \rVert^2 \hspace{-2em}&\\
                    &\vp_{c_i}\!\gets\!(1 - \alpha_c) \vp_{c_i}\!+\!h_{\sigma_i} \sqrt{\alpha_c (2\!-\!\alpha_c) m_{\text{eff}, i}}~ \bblue{\phi(\vx_i) / \sigma_i}  \hspace{-2em}&\\
                    \begin{split}
                        &\vC_i \gets (1 + d(h_{\sigma_i}) - \alpha_1 - \alpha_m \tsum_{j=1}^n w_{ij}) \vC_i\\
                        &\qquad + \alpha_1 \vp_{c_i} \vp_{c_i}^T + \alpha_m \tsum_{j=1}^n \bar{w}_{ij} \vy_{ij} \vy_{ij}^T
                    \end{split}
                \end{flalign*}
            \EndFor
        \EndFor
    \end{algorithmic}
\end{algorithm}

\section{Related Work}\label{secRelated}
\paragraph{Stein Variational Gradient Descent Extensions} 
SVGD is a popular method for sampling from unnormalized densities.
As such, SVGD has been an active field of research and many extensions have been proposed.
These include approaches to improve the performance in high dimensions, for instance using projections \citep{chen2019projected}
or by adjusting the particle update to reduce its bias \citep{d2021annealed, ba2021understanding}.
Other extensions include non-Markovian steps \citep{ye2020stein, liu2022grassmann}, learning-based methods \citep{langosco2021neural, zhao2023stein}, and domain-specific kernel functions \citep{sharma2023task, barcelos2024path}.
While our focus lies on gradient-free SVGD approaches, most of these ideas could be integrated into our approach, which would be an interesting direction of future research.

\paragraph{Gradient-free sampling}
Many gradient-free sampling methods, like those in the MCMC family, iteratively update a proposal distribution to match the target \citep{andrieu2003introduction}.
A shortcoming of these approaches is their slower sampling procedure compared to SVGD, as they are prone to be trapped in a single mode over long periods of time on multimodal objectives.
Population-based MCMC methods improve over this by running multiple chains in parallel, which exchange information over time \citep{laskey2003population}.
Notably, parallel tempering methods simulate chains with different temperatures in parallel to improve mode coverage \citep{swendsen1986replica}.
Still, these methods commonly require sample rejections and potentially long burning-in periods. 
Gradient-free SVGD \citep[GF-SVGD]{han2018stein} addresses this by estimating the gradient for SVGD on a surrogate distribution, which allows for interactions between all chains at each update step and fast convergence rates.
Further work improved the computational efficiency of this method by fitting the surrogate to a limited set of points \citep{yan2021gradient}.
However, these surrogate-based methods require a well-chosen prior for surrogate initialization, as they lack an explicit exploration mechanism.
Thus, in practical scenarios, a different gradient-free SVGD approach has been presented which relies on MC estimates of the gradient \citep{liu2017stein}.
In this work, we present a novel perspective on gradient-free SVGD, which combines ideas from the literature on ES.
Different from prior work, we propose a particle update that is based on CMA-ES, a highly efficient ES \citep{hansen2004evaluating}.

\paragraph{Evolution Strategies}
ES are a specific class of blackbox optimization methods that iteratively improve a search distribution over solution candidates by implementing specific sampling, evaluation and update mechanisms \citep{rechenberg1978evolutionsstrategien}.
While ES commonly use a single distribution \citep{li2020evolution}, it has been demonstrated that their efficiency can be improved by employing restarts or multiple runs in parallel \citep{auger2005restart, pugh2016quality}.
For instance, restarts with increasing population sizes have been demonstrated to improve the performance of CMA-ES \citep{loshchilov2012alternative}.
A downside of restarting approaches is their sequential nature, which makes them slower and prohibits exploiting the benefits of modern GPUs.
Our method is different as it uses the SVGD update to sample multiple subpopulations in parallel, which naturally enables to explore multiple modes.
In particular, our proposed SVGD-based update is simpler to compute than other distributed updates \citep{wang2019distributed}, yet more informed than uncoordinated parallel runs.

\section{Experiments}\label{secExp}
We compare SV-CMA-ES against the two existing approaches for zero-order SVGD from the literature: \textit{GF-SVGD} as state-of-the-art method for surrogate-based inference, and the MC gradient SVGD as state-of-the-art gradient approximation method. 
We refer to the latter as \textit{SV-OpenAI-ES} throughout the remainder of the paper following the naming convention of the ES community.
Furthermore, we compare against gradient-based SVGD, which we denote as $\N$-SVGD in the following.
All strategies have been implemented based on the evosax library \citep{lange2023evosax}.

To guarantee a fair comparison, we keep the number of function evaluations equal for all methods.
In other words, if the ES-based methods are evaluated for 4 particles, each sampling subpopulations of size 16, we evaluate GF-SVGD and $\N$-SVGD with 64 particles.
For each kernel-based method, we use the standard RBF kernel.
For GF-SVGD, we follow the setup of \citet{han2018stein} and use the same kernel function for the SVGD kernel $k$ and the surrogate kernel $k_\rho$, as well as an isotropic Gaussian prior $\NN(0, \sigma^2 \Id)$.
The optimal scale $\sigma^2$ of the prior is found via a hyperparameter grid search. 
For each method, we implement the annealed version of $\N$-SVGD (cf. Eq.\ \eqref{eq:asvgd}) using a logarithmic schedule $\gamma(t) = \max(\log(T / t), 1)$.
Unless specified differently, this choice is followed in all experiments.
For all methods that require an internal optimizer, we use Adam \citep{kingma2014adam}.
We carefully search for the best hyperparameters for each algorithm separately, to guarantee a fair comparison.
The full details of our experimental setup can be found in the Appendix \ref{secExpDets}.
Moreover, we refer to Appendix \ref{secSupplRes} for additional results including ablation studies and empirical runtime and convergence analyses.

\subsection{Sampling from Synthetic Densities}\label{secExpSynth}

\begin{figure}
    \centering
    \captionsetup[subfloat]{labelfont={scriptsize, bf},textfont={scriptsize, bf},format=plain,justification=centering,singlelinecheck}
    \begin{sideways} {\hspace{1mm} \tiny \textbf{Gaussian Mixture}} \end{sideways}\hspace{-1mm}
    \subfloat{
        \adjustbox{trim=3.5pt 3.5pt 3.5pt 3.5pt, clip}{\includegraphics[width=48pt]{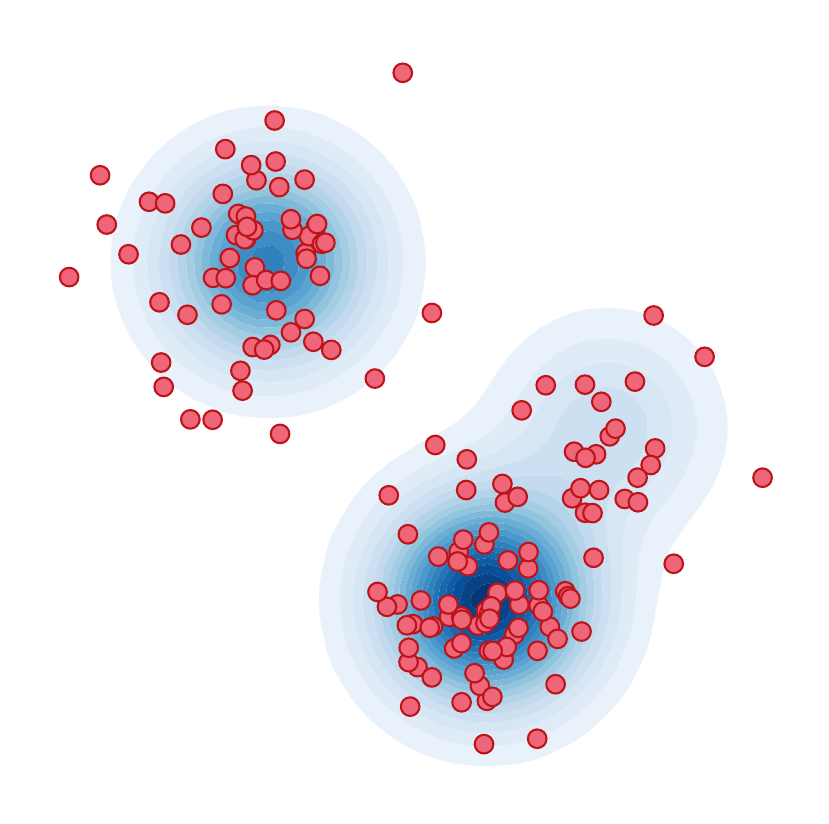}}
    }
    \subfloat{
        \adjustbox{trim=3.5pt 3.5pt 3.5pt 3.5pt, clip}{\includegraphics[width=48pt]{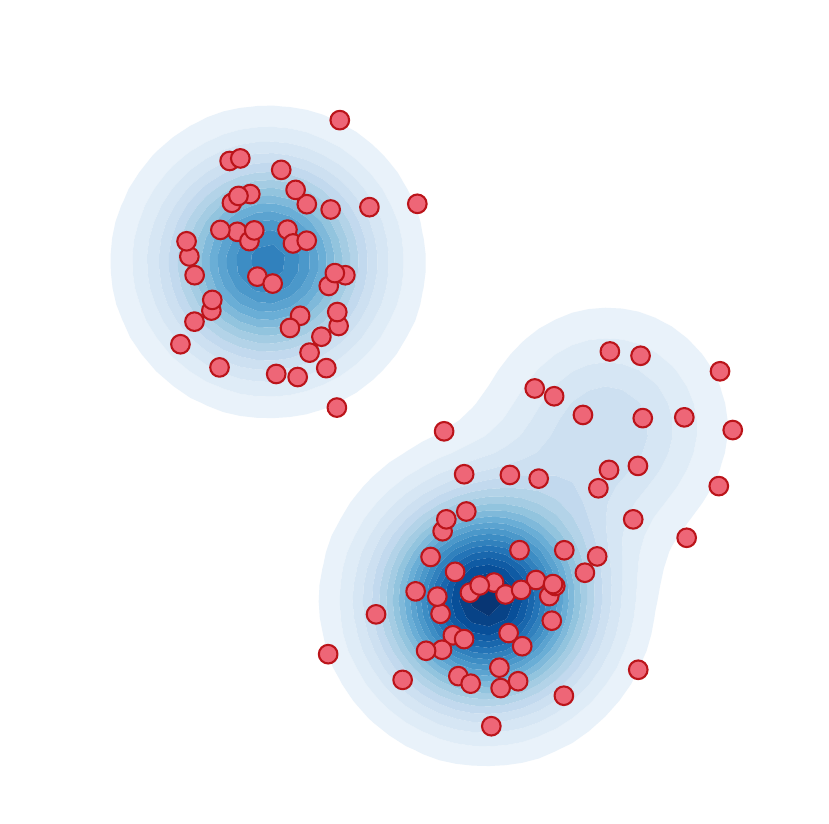}}
    }
    \subfloat{
        \adjustbox{trim=3.5pt 3.5pt 3.5pt 3.5pt, clip}{\includegraphics[width=48pt]{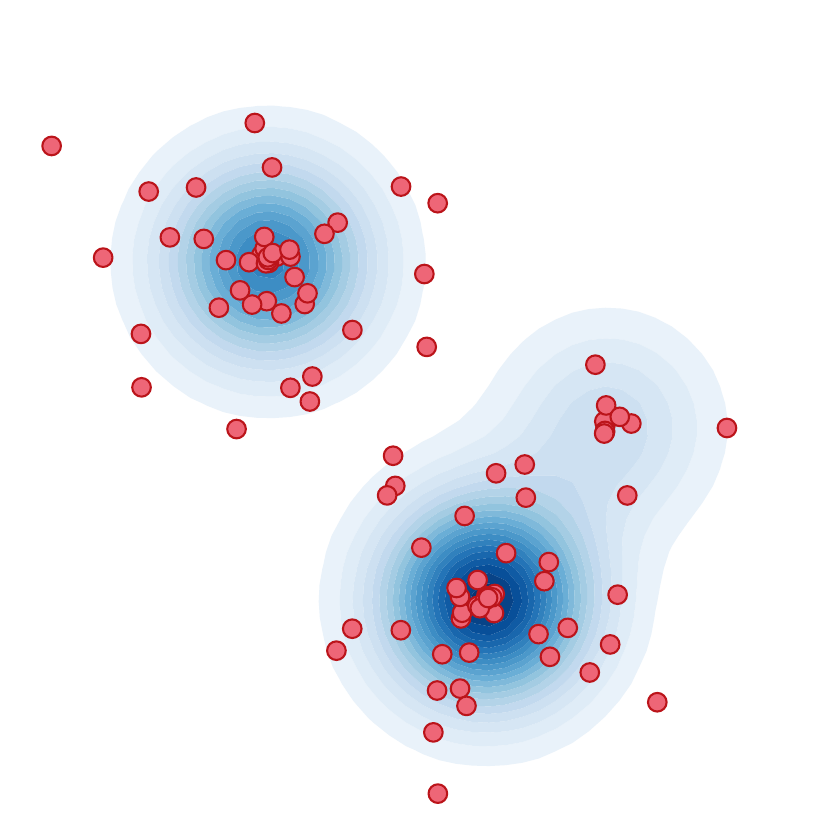}}
    }
    \subfloat{
        \adjustbox{trim=3.5pt 3.5pt 3.5pt 3.5pt, clip}{\includegraphics[width=48pt]{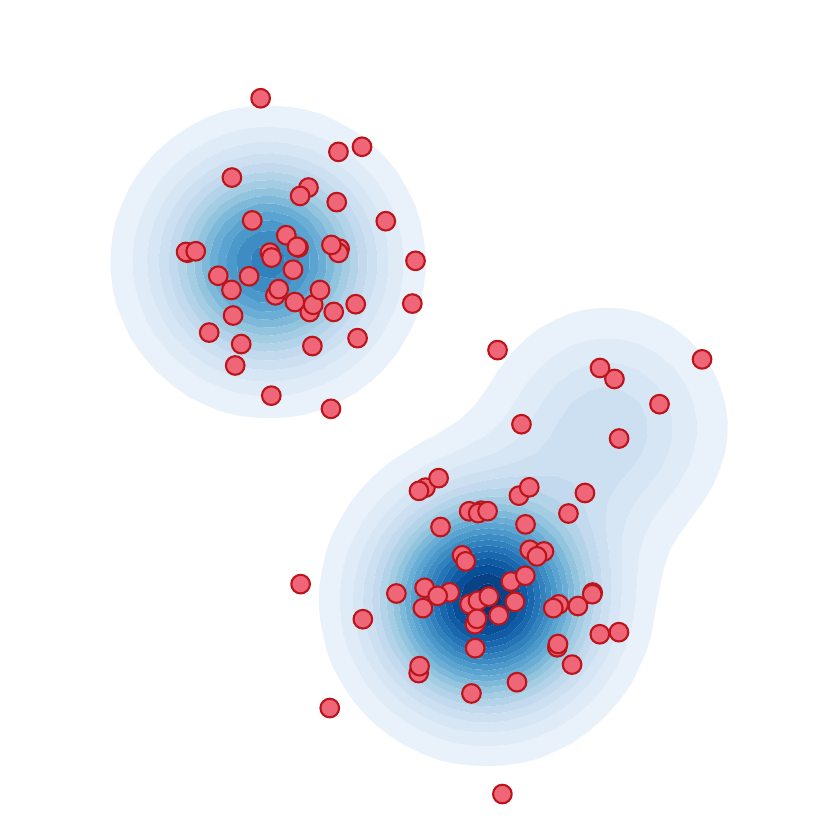}}
    }
    \subfloat{
        \adjustbox{trim=3.5pt 3.5pt 3.5pt 3.5pt, clip}{\includegraphics[width=48pt]{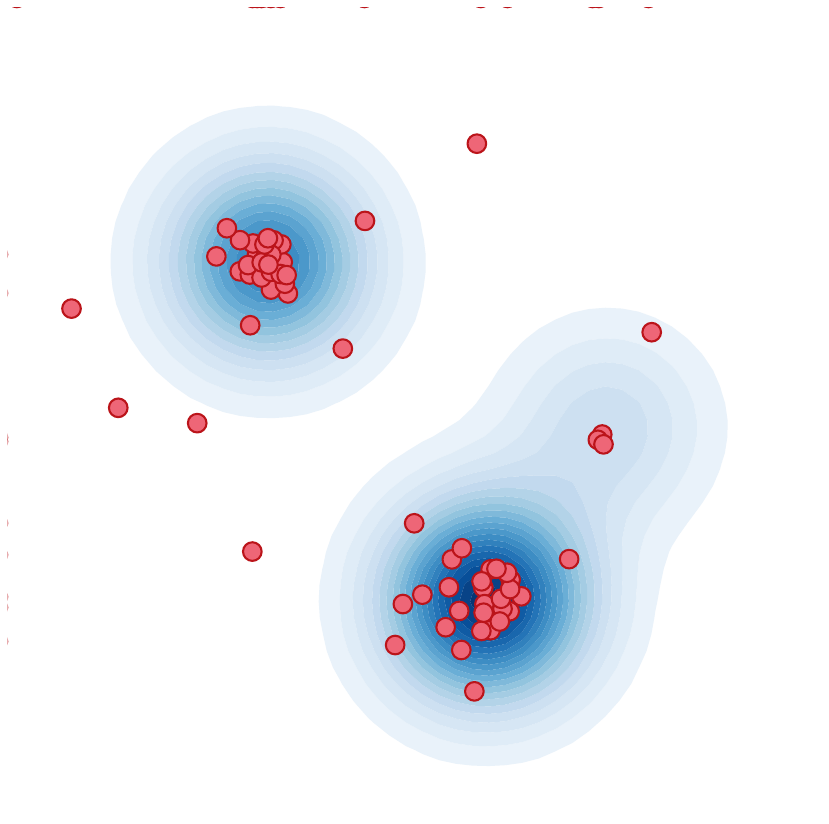}}
    }\\[-2mm]
    \begin{sideways}{\hspace{1.5mm} \tiny \textbf{Double banana}}\end{sideways}\hspace{-1mm}
    \subfloat{
        \includegraphics[width=41pt]{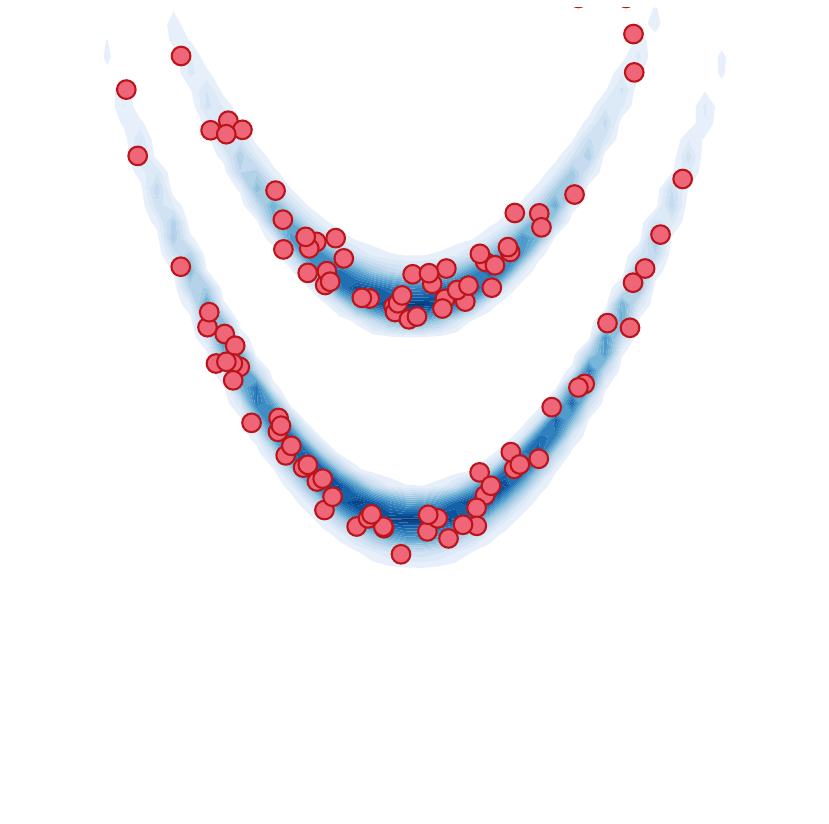}
    }
    \subfloat{
        \includegraphics[width=41pt]{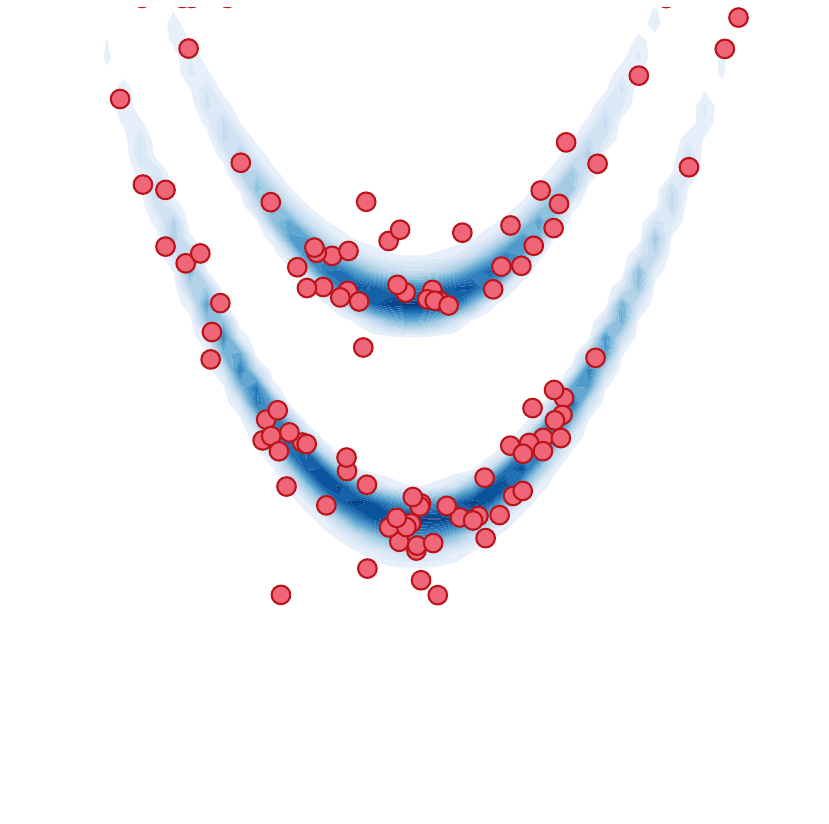}
    }
    \subfloat{
        \includegraphics[width=41pt]{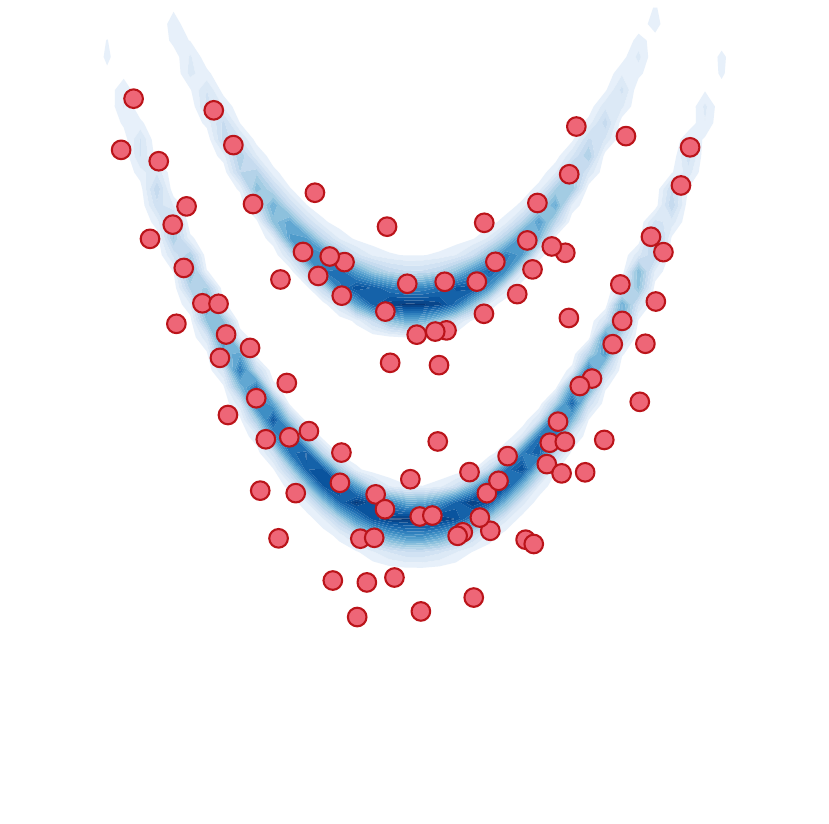}
    }
    \subfloat{
        \includegraphics[width=41pt]{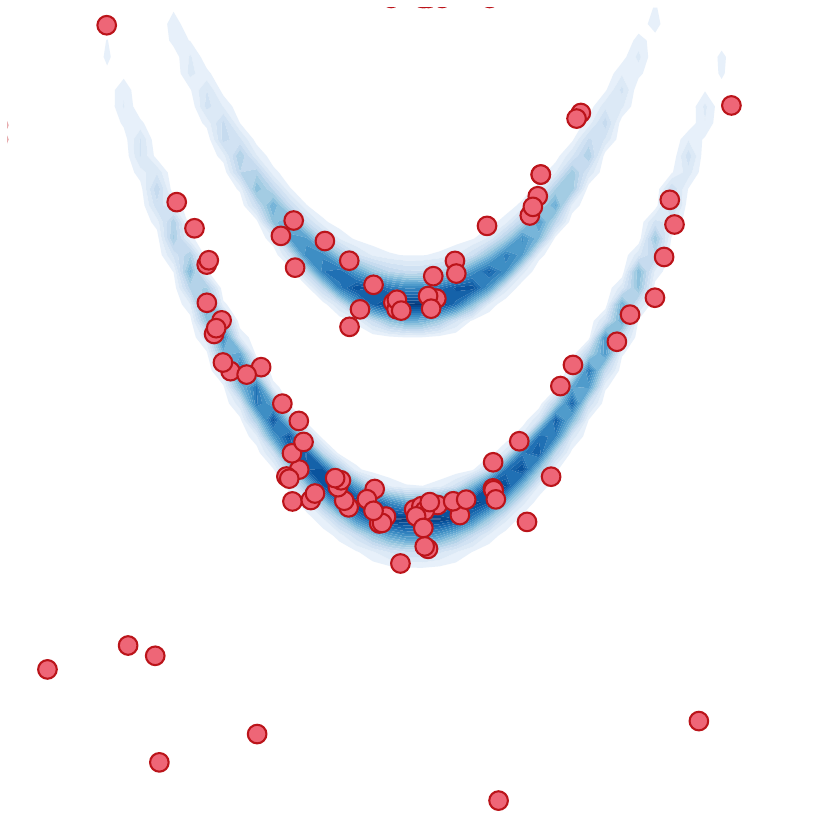}
    }
    \subfloat{
        \includegraphics[width=41pt]{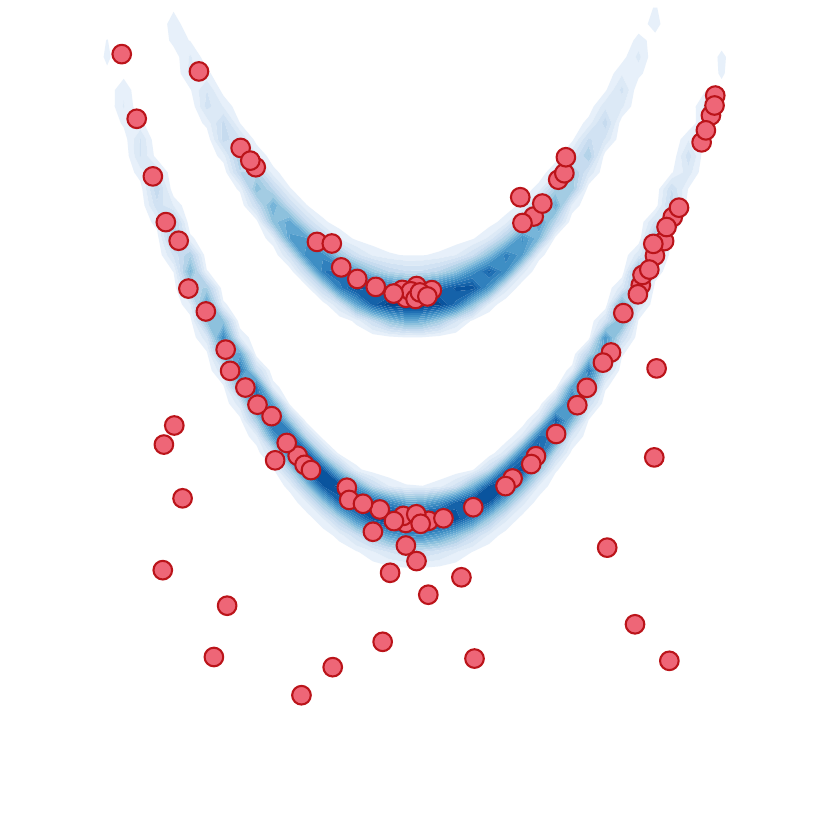}
    }\\[-2mm]
    \setcounter{subfigure}{0}
    \begin{sideways}{\tiny \textbf{Motion planning} }\end{sideways}\hspace{-1mm}
    \subfloat[Ground truth]{
        \includegraphics[width=41pt]{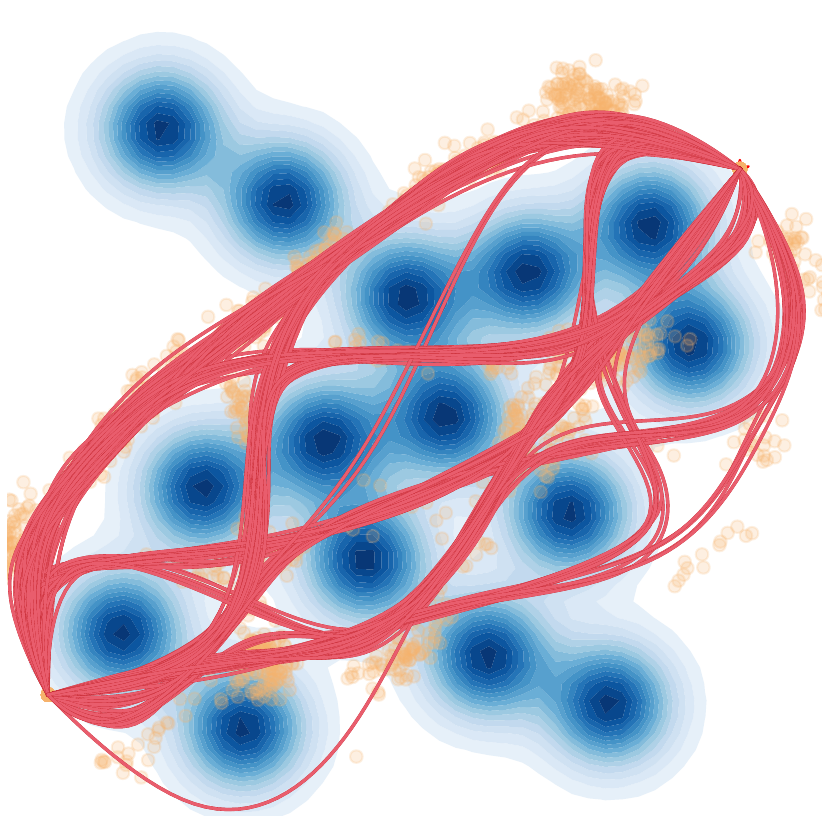}
    }
    \subfloat[$\N$-SVGD]{
        \includegraphics[width=41pt]{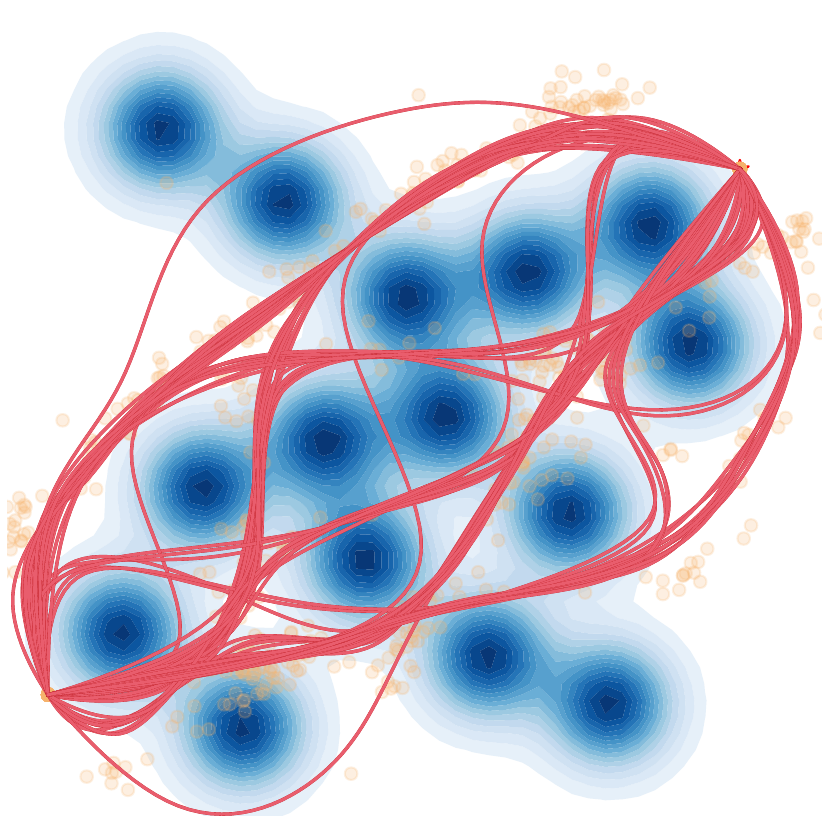}
    }
    \subfloat[SV-CMA-ES (ours)]{
        \includegraphics[width=41pt]{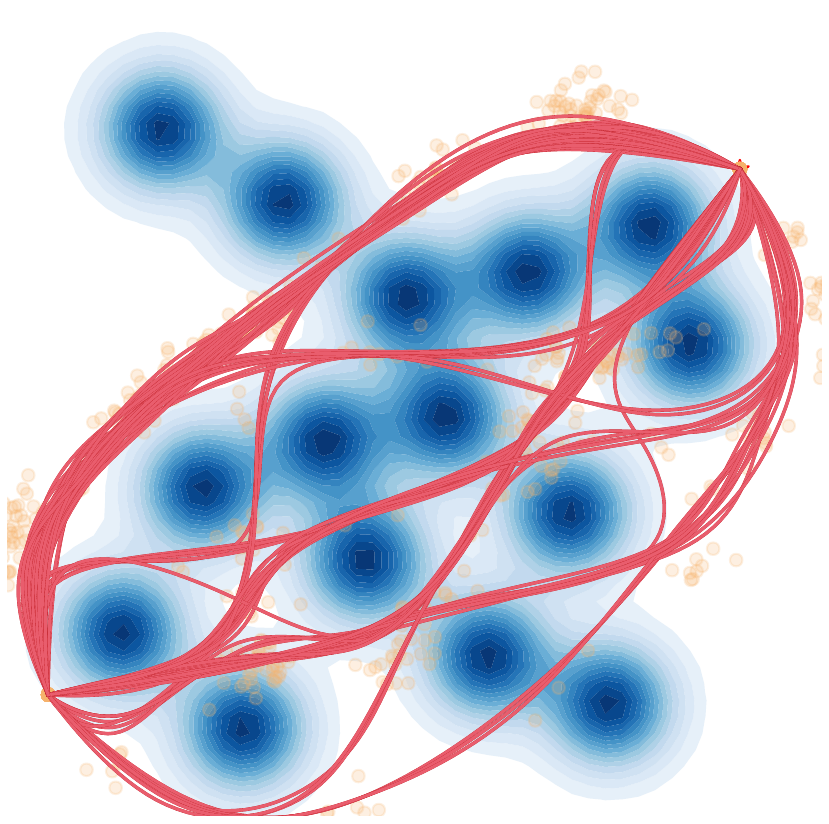}
    }
    \subfloat[GF-SVGD]{
        \includegraphics[width=41pt]{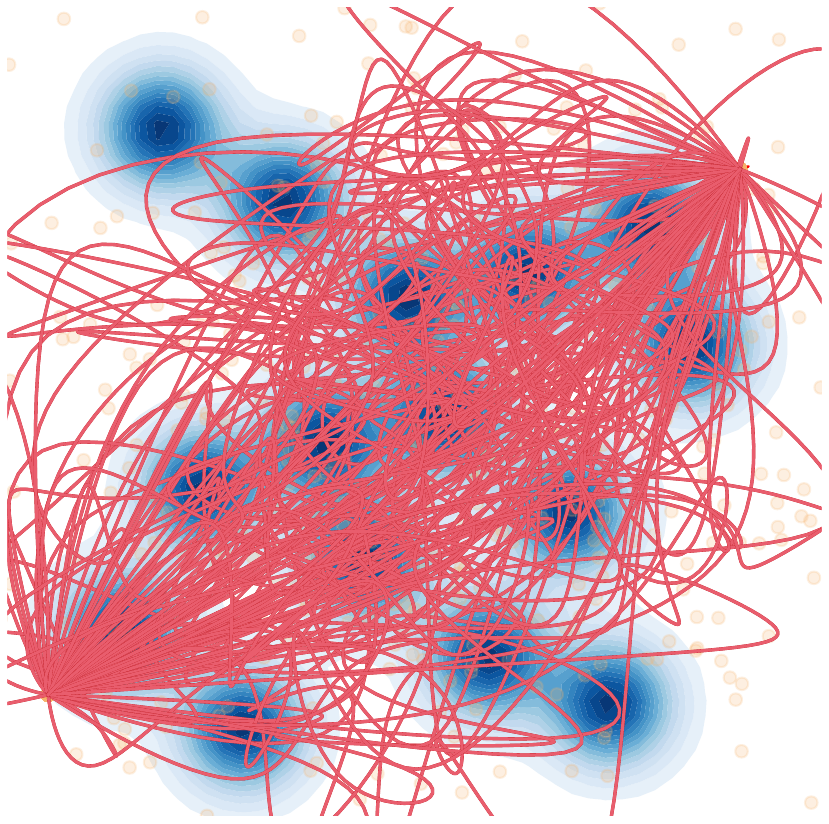}
    }
    \subfloat[SV- OpenAI-ES]{
        \includegraphics[width=41pt]{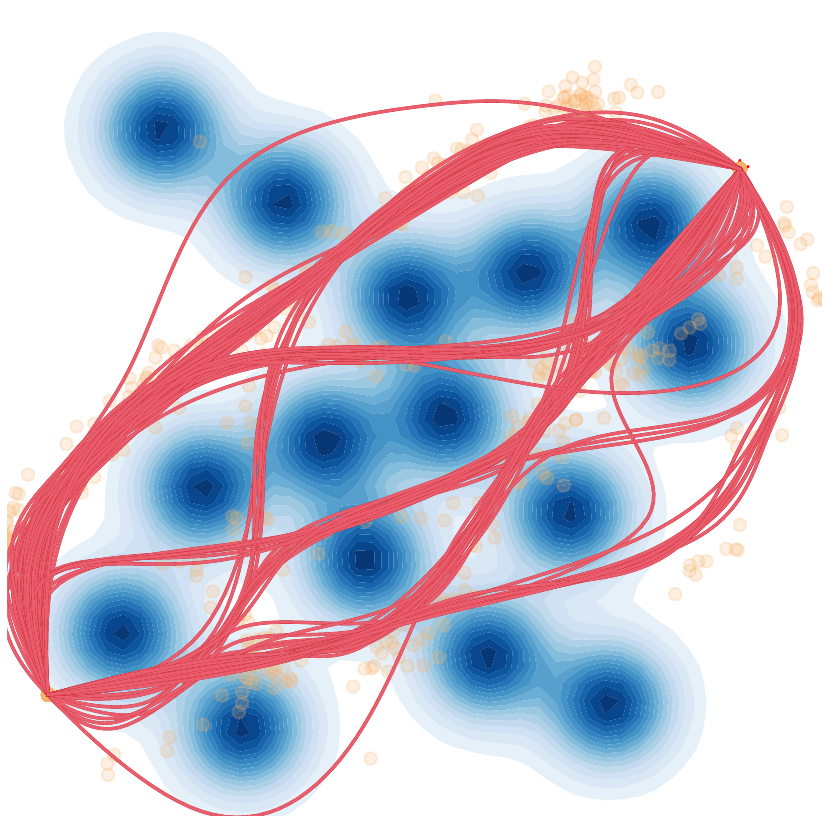}
    }
    \caption{Samples obtained by various methods.
    Gradient-based SVGD (b) captures all target densities effectively, while SV-CMA-ES produces the highest quality samples among gradient-free methods. 
    GF-SVGD struggles on more complex targets, and SV-OpenAI-ES tends to converge slowly due to taking small steps in flat regions of the target.
    }
    \label{fig:samples}
\end{figure}

\begin{figure*}[!btp]
    \centering
    \subfloat[Gaussian Mixture]{
        \includegraphics[width=.21\linewidth]{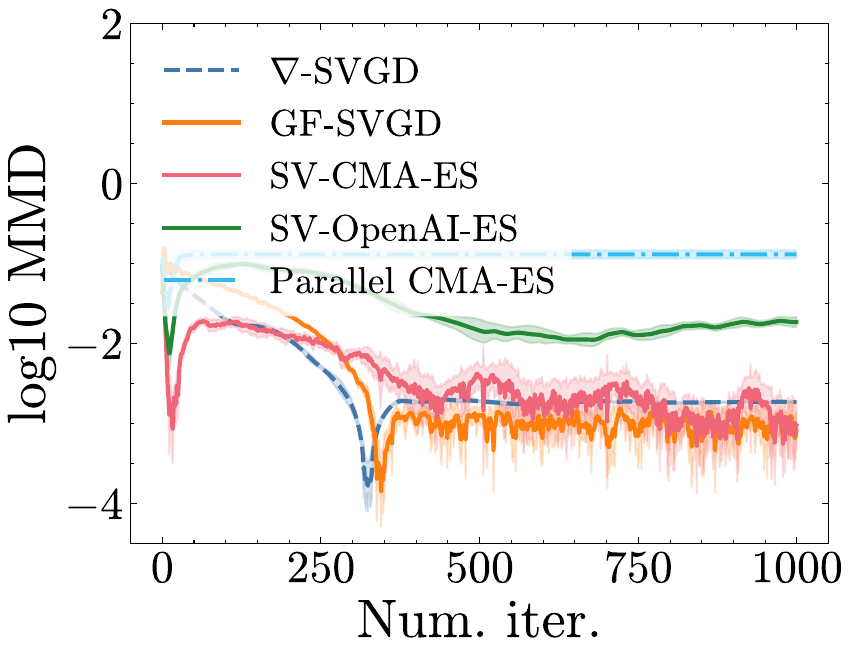}
    }\hfill
    \subfloat[Double Banana]{
        \includegraphics[width=.21\linewidth]{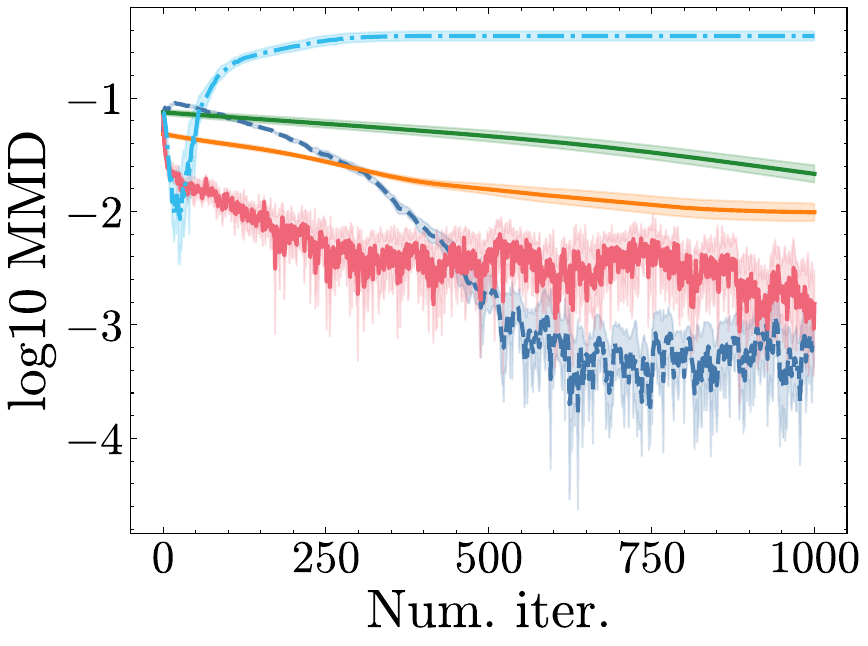}
    }\hfill
    \subfloat[Motion Planning]{
        \includegraphics[width=.225\linewidth]{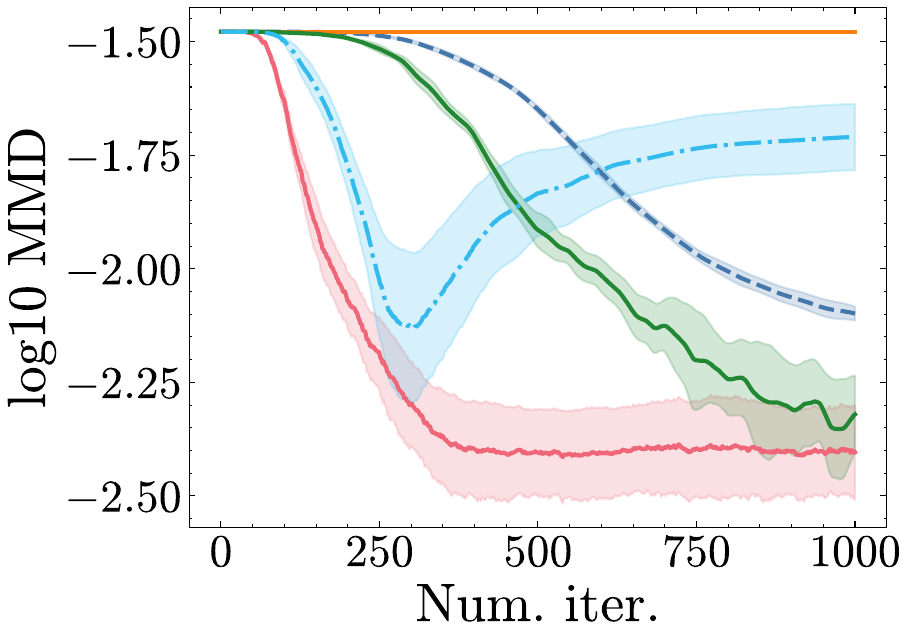}
    }\hfill
    \subfloat[MMD \wrt $\N$-SVGD]{
        \includegraphics[width=.21\linewidth]{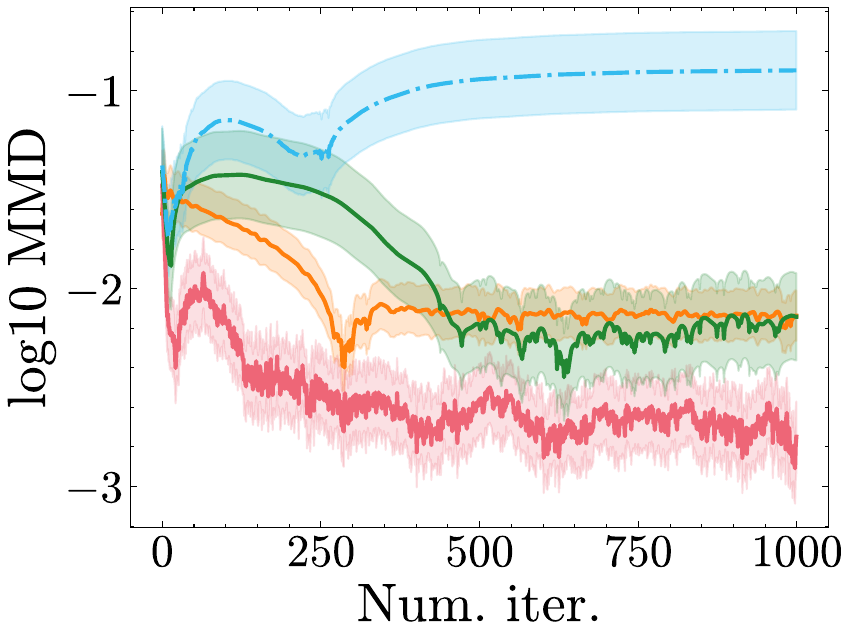}
    }\hfill
    \caption{\textbf{(a)-(c):} MMD \wrt \textit{ground truth}  samples on the synthetic densities depicted in \Cref{fig:samples}. \textbf{(d):} Mean log10 MMD across all three sampling tasks \wrt the \textit{samples obtained by gradient-based SVGD}. 
    All results are averaged across 10 independent runs ($\pm 1.96$ standard error). 
    SV-CMA-ES approximates the ground truth samples and results by gradient-based SVGD (blue line) the best out of all gradient-free methods.
    }
    \label{fig:mmd}
\end{figure*}

\paragraph{Setting} 
We first evaluate our method on multiple synthetic densities to illustrate the quality of the generated samples. 
The closed form pdf for every problem is listed in \Cref{secExpDets}.
We use a total population size of 400 for all methods, which is split across 100 particles for the ES-based algorithms. 
In other words, each particle samples an ES population of 4.
Following common practice in the literature, we quantify sampling performance by evaluating the Maximum Mean Discrepancy \citep[MMD]{gretton2012kernel} of the particles with respect to ground truth samples.
We additionally evaluate the scaling to higher particle numbers in Sec.~\ref{sec:abl}.

\paragraph{Results} 
Figures \ref{fig:samples} and \ref{fig:mmd} display the qualitative and quantitative sampling results.
As expected, $\N$-SVGD generates high-quality samples for all problems.
We find that among the gradient-free methods, SV-CMA-ES performs the best across all problems.
While GF-SVGD generates high-quality samples for the Gaussian mixture, the variance of the generated samples on the double banana density is too high, and the samples for the motion planning problem are of poor quality.
Concurrently, SV-OpenAI-ES performs well on the motion planning problem, but on the others it underestimates the variance (Gaussian mixture) or converges slowly (double banana, also see Fig.~\ref{fig:banana_convergence_iter}).
These results highlight the fast convergence properties of our method, as it employs the automatic step-size adaptation of CMA-ES.
Additionally, \Cref{fig:mmd}~(d) displays the MMD \wrt the samples that were obtained by $\N$-SVGD, aggregated across all sampling tasks.
Our results demonstrate that SV-CMA-ES can indeed quickly converge to a set of samples that approximates the outcomes of $\N$-SVGD well, and better than the two other gradient-free baselines.
We further illustrate these results in \cref{sec:app_full} where we display the sample sets for all sampling tasks.
Moreover, we illustrate the benefit of using the presented algorithm compared to other CMA-ES-based methods in Fig.~\ref{fig:sves-overview} -- since prior CMA-ES methods only maximize likelihood, the diversity of samples is low.

\subsection{Bayesian Logistic Regression}
\paragraph{Setting} Next, we evaluate our method on Bayesian logistic regression for binary classification.
We follow the setup of \citet{langosco2021neural}, which uses a hierarchical prior $p(\theta)$ on the parameters $\theta = [\alpha, \beta]$, where $\beta \sim \NN (0, \alpha^{-1})$ and $\alpha \sim \Gamma (a_0, b_0)$.
Given data $D$, the task is to approximate samples from the posterior
\begin{align*}
    &p(\theta \mid D) = p(D \mid \theta)p(\theta) \quad \text{ with:}\\
    &p(D \mid \theta) = \prod_{i=1}^N \bigl[ y_i \tfrac{\exp (x_i^T \beta)}{1 + \exp (x_i^T \beta)} + (1- y_i) \tfrac{\exp (-x_i^T \beta)}{1 + \exp (-x_i^T \beta)} \bigr].
\end{align*}
We consider the binary \textit{Covtype}, \textit{Spambase}, and the \textit{German credit} datasets from the UCI Machine Learning Repository \citep{asuncion2007uci}, as suggested in prior work \citep{liu2016stein, arenz2020trust, futami2018variational}. 
For all experiments, we use a total population of 256, which is split across 8 particles for the ES-based methods.

\paragraph{Results} 
For each dataset, we report the accuracy and negative log-likelihood (NLL) across the entire particle set, and report the mean performance across 10 runs.
Our results demonstrate that SV-CMA-ES outperforms the remaining gradient-free algorithms.
On both datasets, our method is the fastest converging among the gradient-free methods.
Furthermore, its final performance is considerably better than GF-SVGD on all datasets.
While the performance of $\N$-SVGD is slightly better on the Covtype dataset, SV-CMA-ES is on par with it for the Spam dataset.
Additionally, on the credit data, we find that ES-based methods are both more accurate and exhibit greater stability than the gradient-based SVGD, which underlines the potential of zero-order methods in this context.

\begin{figure}[!ht]
    \centering
    \subfloat{
        \includegraphics[width=.3\linewidth]{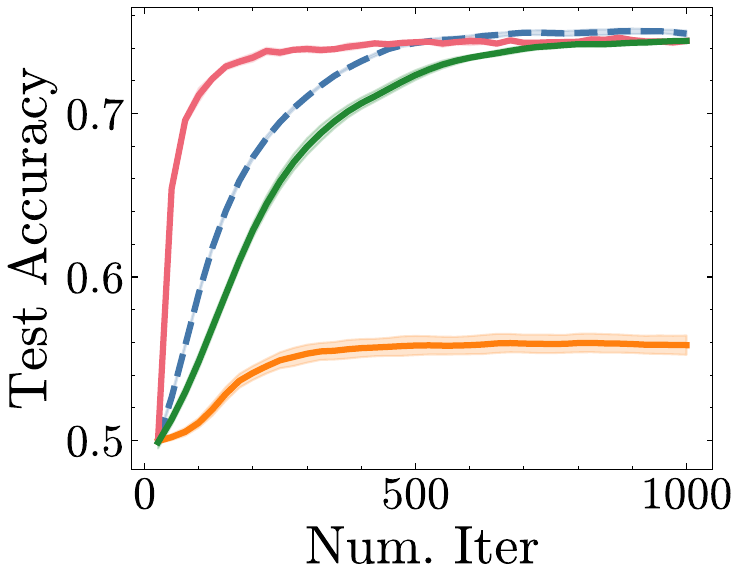}
    }
    \subfloat{
        \includegraphics[width=.3\linewidth]{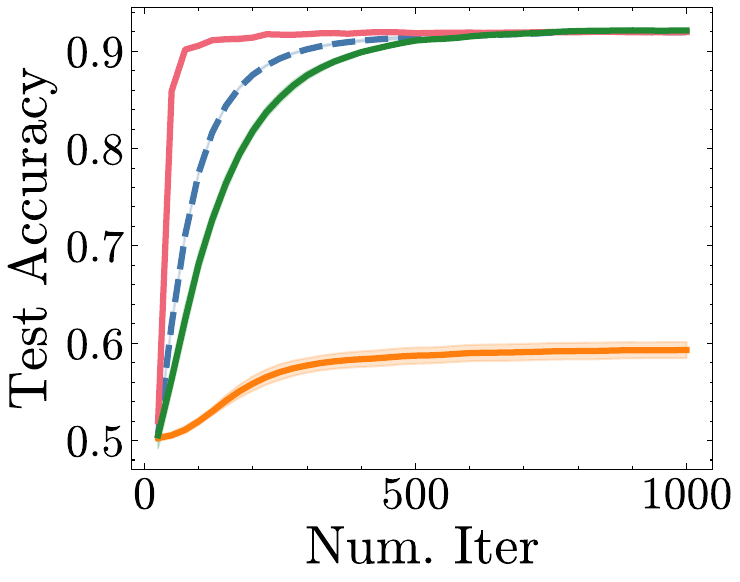}
    }
    \subfloat{
        \includegraphics[width=.3\linewidth]{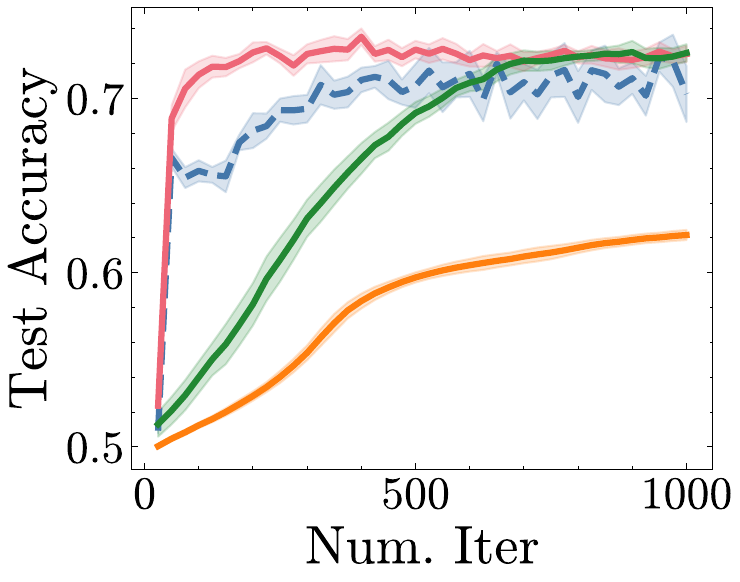}
    }\\
    \setcounter{subfigure}{0}
    \subfloat[Covtype]{
        \includegraphics[width=.3\linewidth]{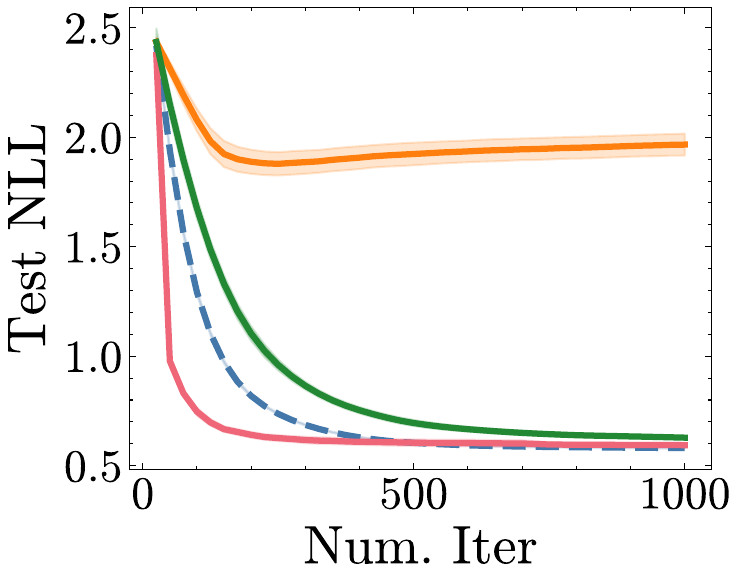}
    }
    \subfloat[Spam]{
        \includegraphics[width=.3\linewidth]{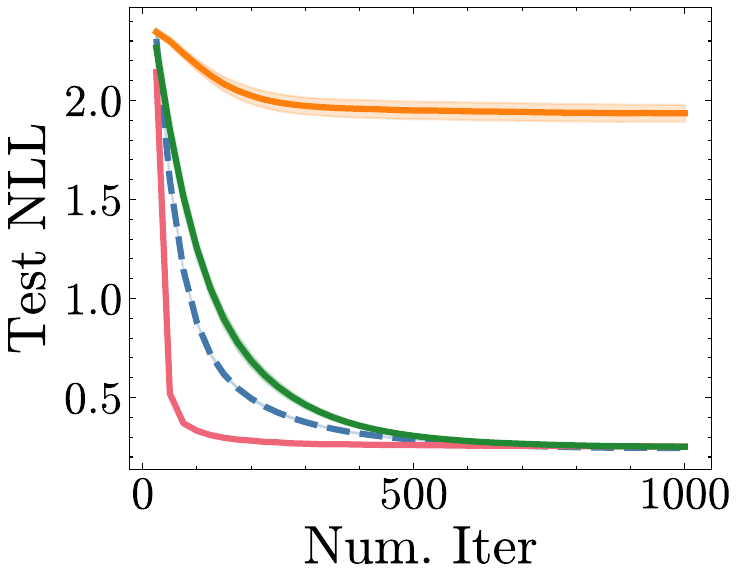}
    }\subfloat[Credit]{
        \includegraphics[width=.3\linewidth]{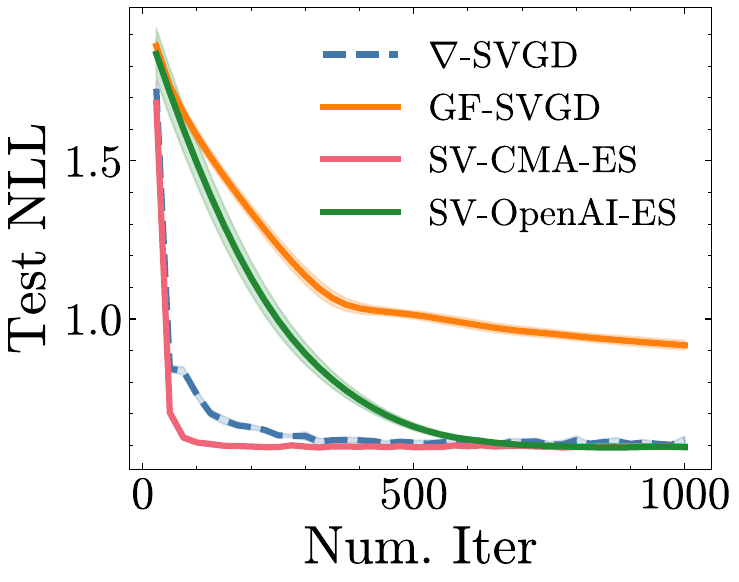}
    }
    \caption{
        Results of Bayesian logistic regression. 
        We report mean ($\pm 1.96$ standard error) across 10 independent runs. 
        SV-CMA-ES converges the faster than other gradient-free methods, and achieves similar performance levels at convergence as gradient-based SVGD (dashed line).
    }
    \label{fig:log_reg}
\end{figure}

\subsection{Reinforcement Learning}\label{secExpRl}
\paragraph{Setting} 
We further assess the performances of the gradient-free SVGD methods on six classic reinforcement learning (RL) problems.
The goal of each RL task is to maximize the expected episodic return $J(\theta)$, where each particle $\theta$ now parametrizes a multi-layer perceptron (MLP).
The corresponding inference objective is to sample policy parameters $\theta$ from the following Boltzmann distribution:
\begin{equation*}
    p(\theta) \propto \exp(J(\theta)), \quad J(\theta) = \EEE_{(s_t, a_t) \sim \pi_{\theta}} \bigl[\sum_{t=1}^T r(s_t, a_t)\bigr]
\end{equation*}
where $(s_t, a_t) \sim \pi_{\theta}$ represent a trajectory sampled from the distribution that is induced by the policy parametrized by $\theta$.
For each problem, we train a 2-hidden layer MLP with 16 units per layer, which implies high-dimensional optimization problems as each MLP has several hundred parameters.
The specific numbers vary across the benchmarks and are listed in \Cref{table:hyperparams} in the Appendix.
We use a total population size of 64 which we split into 4 subpopulations for the ES-based methods and estimate the expected return across 16 rollouts with different seeds.
To make the results comparable to other works on ES for RL, we follow the approach of \citet{lee2023stamp} and extend the optimization by a phase that attempts to find exact optima.
We realize this by fading out the repulsive term via the schedule $\gamma (t) = \log (T / t)$.

\begin{figure}[ht]
    \centering
    \subfloat[Pendulum]{
        \includegraphics[width=.32\linewidth]{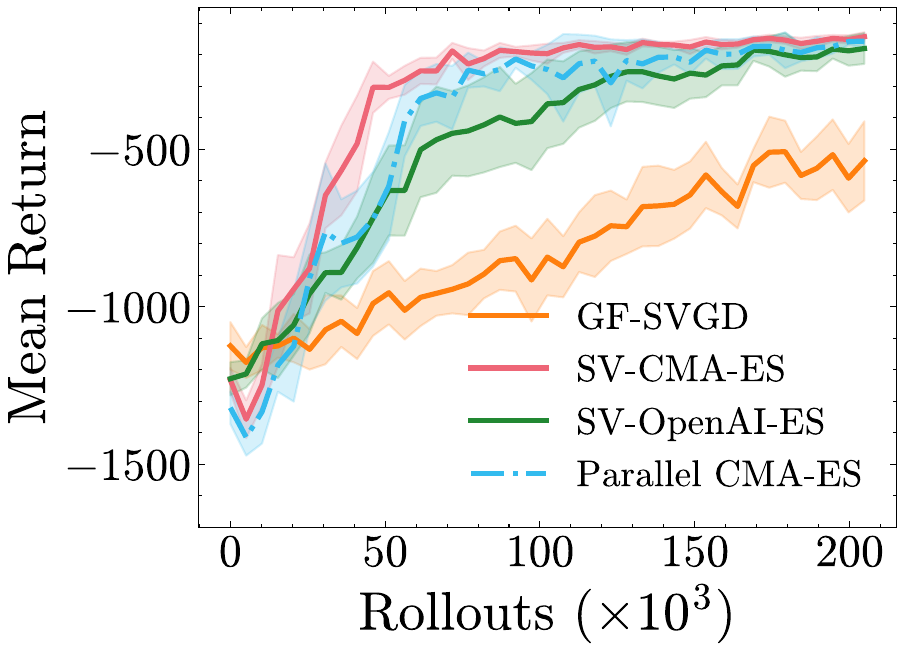}
    }
    \subfloat[CartPole]{
        \includegraphics[width=.3\linewidth]{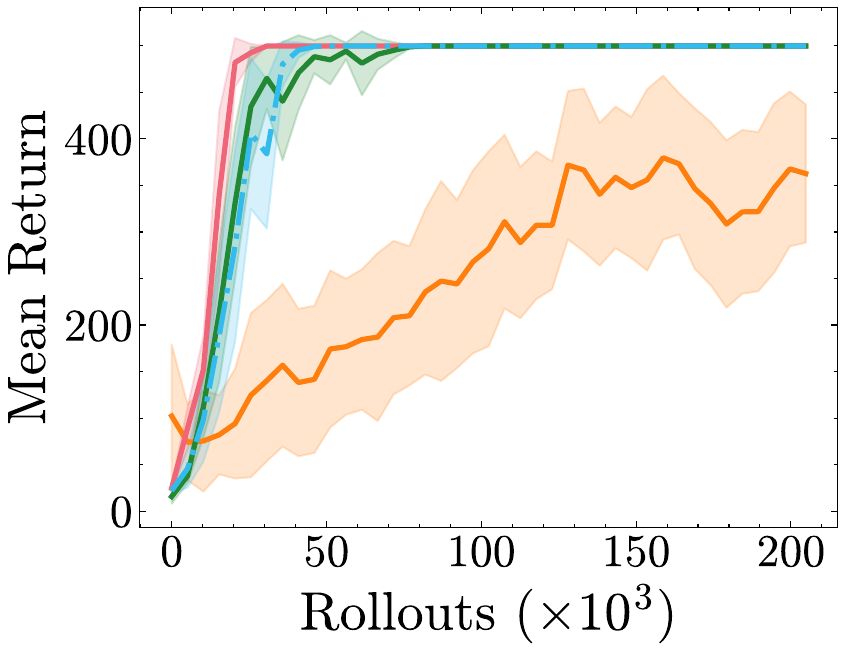}
    }
    \subfloat[MountainCar]{
        \includegraphics[width=.3\linewidth]{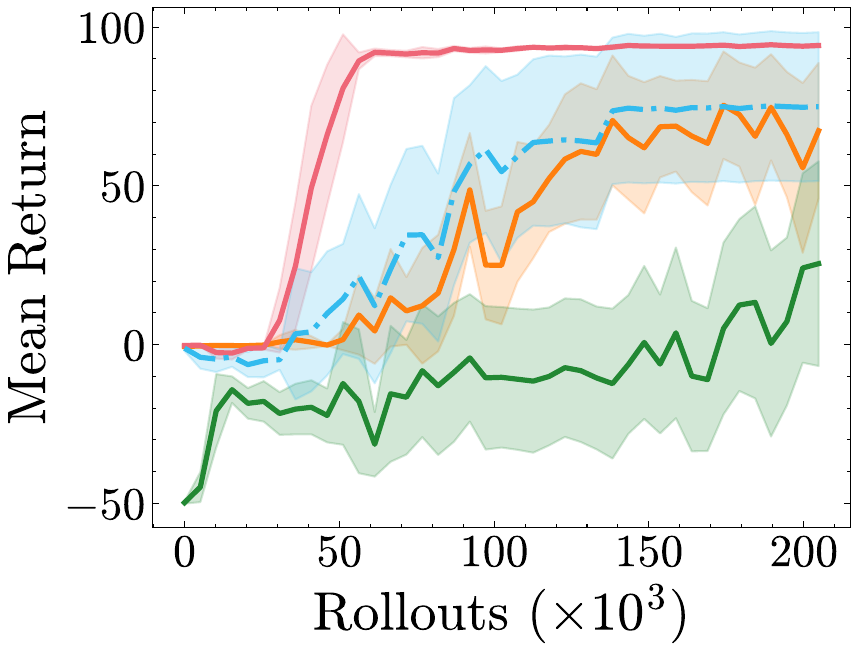}
    }\\[-.5em]
    \subfloat[Halfcheetah]{
        \includegraphics[width=.3\linewidth]{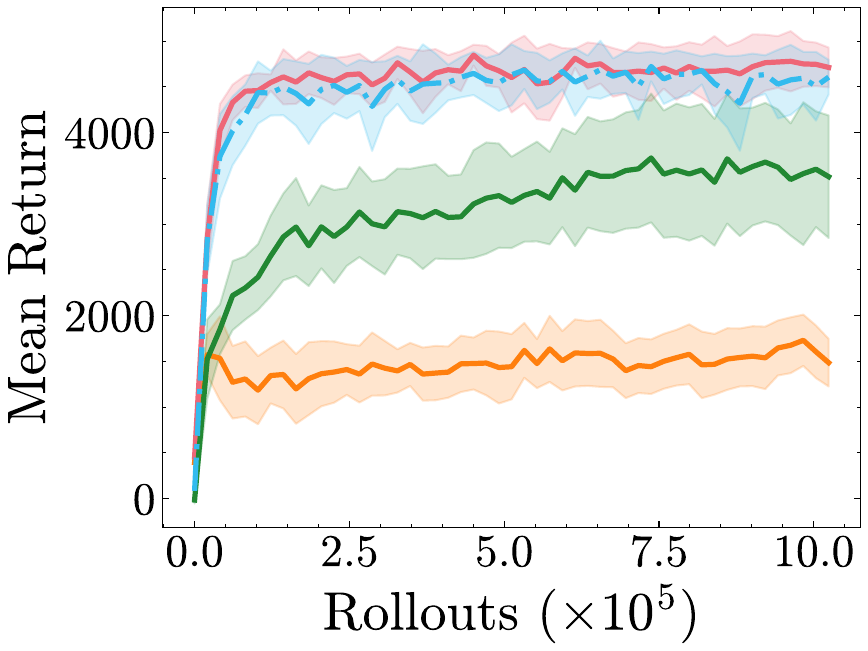}
    }
    \subfloat[Hopper]{
        \includegraphics[width=.3\linewidth]{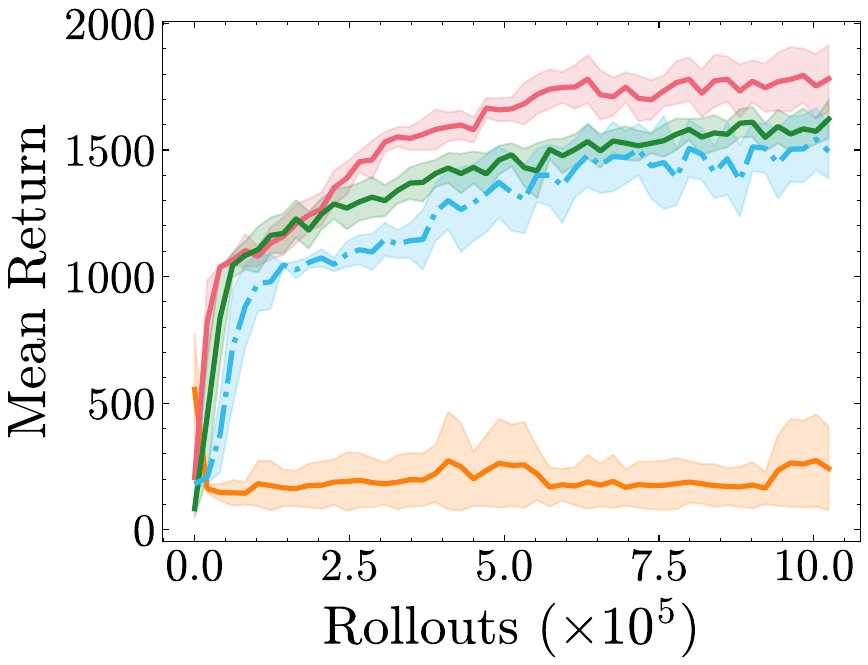}
    }
    \subfloat[Walker]{
        \includegraphics[width=.3\linewidth]{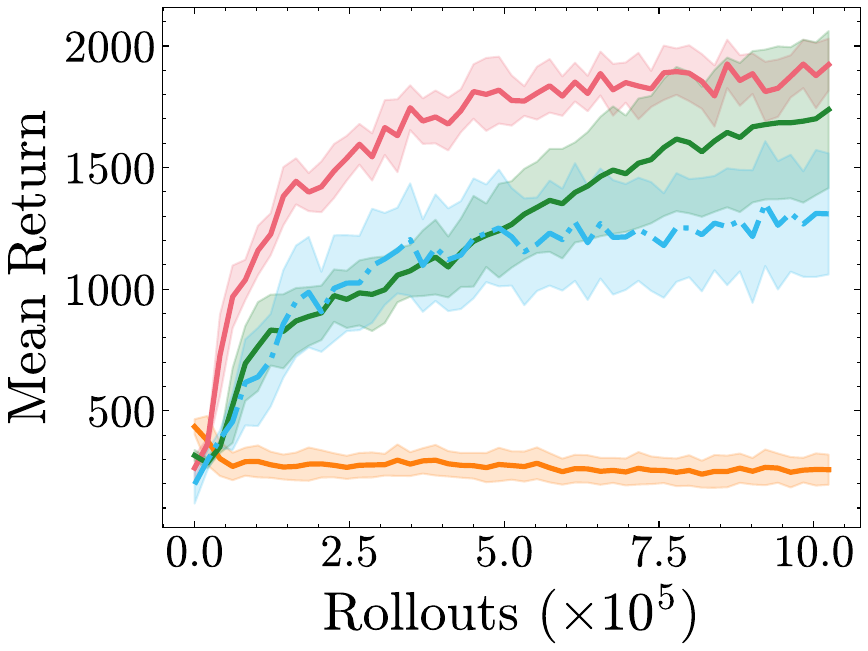}
    }
    \caption{Results of sampling MLP parameters for RL tasks. 
    Plotted is the best expected return across all particles for each method. 
    We report the mean ($\pm 1.96$ standard error) across 10 independent runs.
    SV-CMA-ES performs better than the gradient-free baselines across all tasks.
    }
    \label{fig:rl}
\end{figure}

\paragraph{Results} 
We display the aggregated results across all RL tasks in Fig.~\ref{fig:sves-overview}, and the individual task performances in Fig.~\ref{fig:rl}.
Our results showcase a strong performance of SV-CMA-ES.
In comparison to other gradient-free versions of SVGD, it is the only method that generates high scoring solutions for all problems.
In particular, we observe that SV-CMA-ES is the only method that solves the MountainCar problem consistently, while it is the fastest to converge on Pendulum.
Both of these environments feature a local optimum at which agents remain idle to avoid control costs \citep{eberhard2023pink}.
It is on these problems that GF-SVGD converges to such optima in certain runs, which we further illustrate in Fig.\ \ref{fig:mc_seeds} in the Appendix.
These results illustrate that SV-CMA-ES improves over GF-SVGD by sampling stochastic ES steps, which leads to a higher exploration of the domain.
Interestingly, our results further show that SV-OpenAI-ES may deliver good samples in some runs, but the high standard error on several problems underline its sensitivity to initialization.
These findings confirm that SV-CMA-ES is a strong gradient-free SVGD scheme, capable of sampling from densities and optimizing blackbox objectives.
Further, we would like to note that our final performances are comparable to those reported in prior, gradient-based, work \citep{jesson2024relu}, which again underlines the potential of our method.

Further, we analyze the benefits of the kernel term by comparing our method to uncoordinated parallel runs of CMA-ES.
Overall, we observe a clear performance improvement when using the kernel term.
In particular, in the more challenging Hopper and Walker tasks, the benefits of using SV-CMA-ES over parallel CMA-ES are large.
We extend our analysis of SV-CMA-ES in Appendix \ref{secAblations} where we compare it to vanilla CMA-ES and OpenAI-ES, and conduct additional experiments on sparse reward environments.
This analysis reveals that SV-CMA-ES consistently outperforms competing ES, underscoring its superior performance in environments where effective exploration is essential.

\subsection{Ablation Studies}\label{sec:abl}
\begin{figure*}[ht]
    \centering
    \subfloat[MMD w.r.t.\ ground truth]{
        \includegraphics[width=0.19\linewidth]{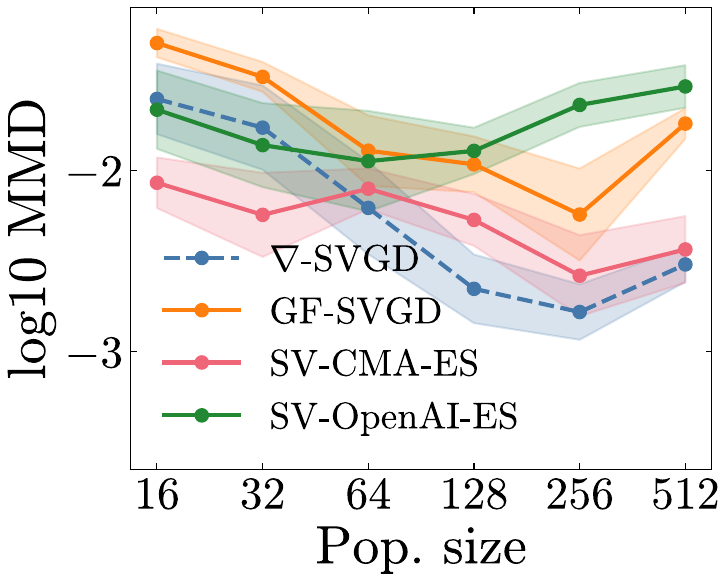}
    }\hfill
    \subfloat[{$\EEE [x]$}]{
        \includegraphics[width=.19\linewidth]{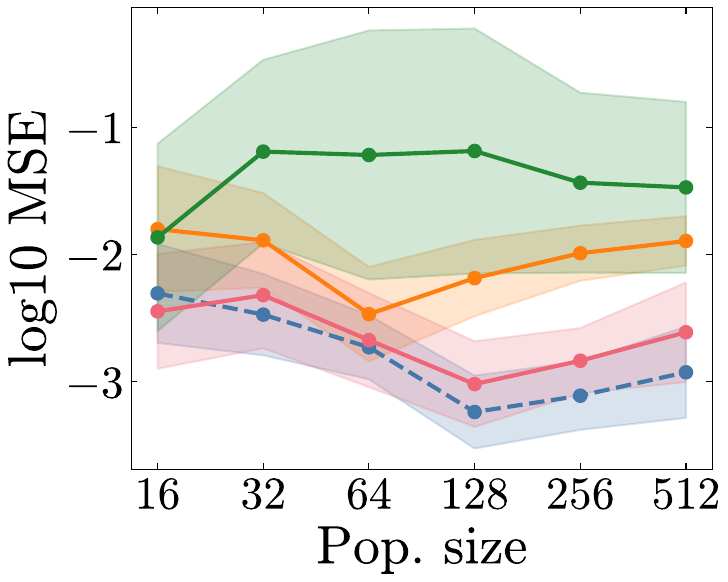}
    }\hfill
    \subfloat[{$\VVV [x]$}]{
        \includegraphics[width=.19\linewidth]{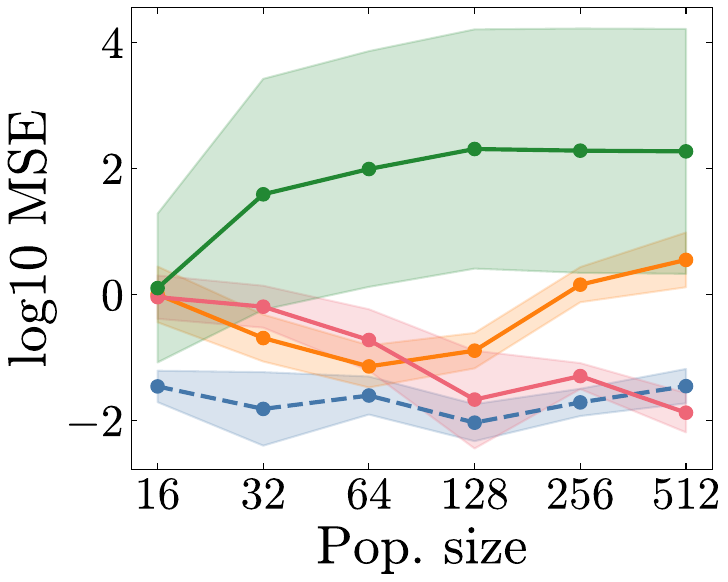}
    }\hfill
    \subfloat[Subpopulation Scaling]{
        \raisebox{.1cm}{\includegraphics[width=.29\linewidth]{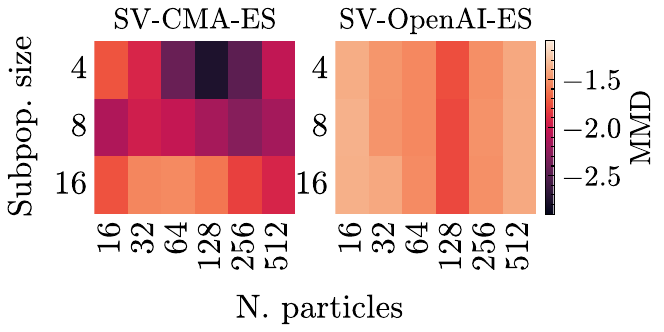}}
    }\hfill
    \caption{ Scaling analysis. Depicted are the final performances for different total population sizes. \textbf{(a)}: MMD vs.\ sample size after 1000 iterations. \textbf{(b)-(c)}: MSE vs.\ sample size when estimating the first two central moments of the ground truth distribution. 
    For ES, we use the same subpop.\ size per particle as in \Cref{fig:mmd}. \textbf{(d)}: Subpopulation size scaling for ES-based SVGD. 
    The results are averaged across 10 independent runs of all synthetic sampling tasks from \Cref{fig:mmd}.
    SV-CMA-ES performs the best out of all gradient-free methods (solid lines) across different particle numbers.
    }
    \label{fig:scaling}
\end{figure*}

\paragraph{Choice of Population Size}
In the experiments above, we investigate the performance for fixed particle numbers and population sizes.
To gain further insights into the scalability of our method, we conduct an additional analysis on the same sampling problems as in \Cref{secExpSynth} using varying particle numbers.
The results of these experiments are displayed in \Cref{fig:scaling}.
In addition to the MMD after $1\,000$ iterations, we report the error when estimating the first two central moments of the target distribution from the generated samples.
We observe the clear trend that SV-CMA-ES performs better than GF-SVGD and SV-OpenAI-ES with increasing particle numbers.
Furthermore, we observe in Fig.~\ref{fig:scaling} (d) that SV-CMA-ES requires fewer samples than SV-OpenAI-ES to estimate good steps.

\paragraph{Choice of Annealing Schedule}
In the experiments above, we used annealed SVGD. 
This decision was made due to its widespread use in the community and many desirable properties. 
However, to assess the quality of our method, it is important to consider the sensitivity to the choice of annealing schedule.
In Table~\ref{table:annealing}, we show the key performance metrics from 10 seeds for SV-CMA-ES with and without annealing.
As we see, performance in all cases is strong, with little difference between the two conditions.

\begin{table}[h]
\centering
\resizebox{\columnwidth}{!}{%
    \begin{tabular}{l|ll}
    \toprule
    \textbf{Task} & \textbf{Annealing} & \textbf{No Annealing} \\
    \midrule
    GMM & -3.03 (0.30) & -2.92 (0.26) \\
    Double Banana & -2.59 (0.16) & -2.83 (0.21) \\
    Motion Planning & -2.40 (0.10) & -2.44 (0.13) \\
    Covertype NLL & 0.59 (0.01) & 0.59 (0.01) \\
    MountainCar RL & 93.68 (0.10) & 93.68 (0.08) \\
    Hopper RL & 1781.31 (132.64) & 1788.13 (79.22) \\
    \bottomrule
    \end{tabular}
}
\caption{Kernel annealing ablation. This table shows SV-CMA-ES performances across 10 seeds with 1.96 standard error in parentheses. All runs use the identical setup, aside from the kernel annealing. No annealing means that we use a constant $\gamma(t)=1$.}\label{table:annealing}
\end{table}

\section{Conclusion}\label{secConc}
\paragraph{Summary} 
We proposed a new gradient-free algorithm that combines elements from evolution strategies and SVGD.
The resulting method, SV-CMA-ES, achieves high computational efficiency by replacing the score term in the SVGD update with an ES step.
On several problems with different characteristics, we demonstrated that SV-CMA-ES outperforms prior gradient-free SVGD-based algorithms consistently.
We could thus confirm our hypothesis that the incorporation of the CMA-ES update enables faster convergence than SVGD with MC gradients, and better overall performance than GF-SVGD across multiple problems.

\paragraph{Limitations}
For stable convergence, we selected a fixed kernel bandwidth via grid search, while prior work used the median heuristic.
Selecting the kernel bandwidth via grid search is costly and thus constitutes a disadvantage.
Furthermore, our approach can be computationally expensive due to the decomposition required for each covariance matrix, leading to a runtime complexity in $\mathcal{O}(\varrho^2d + \varrho d^3)$ in $d$ dimensions with $\varrho$ particles. 
In contrast, SV-OpenAI-ES and GF-SVGD achieve a complexity in $\mathcal{O}(\varrho^2d)$. 
Future work could address this by exploring diagonal covariance matrices, which are commonly used to speed up CMA-ES \citep{ros2008simple}. 
Additionally, we would like to stress that the most time-consuming part of ES is often the fitness evaluation. 
We illustrate this aspect in \Cref{secRuntime} where we present additional plots including empirical runtimes. 
This analysis shows that the wallclock time that SV-CMA-ES requires to produce high-quality solutions is competitive with the baselines.

\paragraph{Future work}
In our experiments, we used the standard RBF kernel, following the convention of many prior works.
Recent work suggested adjusting the size of the considered neighborhood adaptively in the context of particle swarm optimization \citep{zhang2024diffusion}.
One potential extension is to integrate this idea into our approach to improve particle repulsion.
Moreover, we see, for instance in \Cref{fig:mmd}, that our method has higher variance compared to other methods.
Future work could investigate mechanisms to make the optimization more stable.
Finally, we see a potential to scaling up our method to a high number of particles to parallelize ES in an informed way. 
An investigation of scaling laws would be an intriguing avenue of research.

\begin{acknowledgements}
This research was funded by the Amazon Development Center Germany GmbH.
The authors want to thank anonymous reviewers for constructive feedback which helped improve the paper.
\end{acknowledgements}

\bibliographystyle{plainnat}
\bibliography{references}

\newpage
\onecolumn
\makeatletter
\renewcommand{\@author}{}
\makeatother
\title{Supplementary Material}\label{secAppendix}
\maketitle
\appendix
\vspace{-1.5in}
The supplementary material is structured as follows:
\begin{itemize}
    \item \textbf{Appendix \ref{secAlgSuppl}} lists the vanilla CMA-ES algorithm for comparison to our method and the computation of all hyperparameters that we used.
    \item \textbf{Appendix \ref{secExpDets}} presents the full experimental details of our work.
    \item \textbf{Appendix \ref{secSupplRes}} presents additional experimental results. These include ablations of SV-CMA-ES, an empirical runtime analysis, and an empirical convergence analysis.
\end{itemize}

\begin{algorithm}[!ht]
    \caption{The Vanilla CMA-ES Update. Adapted from \citet{hansen2016cma}.}
    \label{alg:cmaes}
    \textbf{Input:} Generation index $t$, search distribution parameters $\vx, \sigma, \vC$; pop.\ size $n$, num.\ elites $m$; learning rates $\epsilon, \alpha_\sigma, \alpha_1, \alpha_{m}, \alpha_c$; damping hyperparam. $d_\sigma$\\
    \begin{algorithmic}[1]
        \State \textbf{Sample and evaluate new population of search points}, for $i = 1, \dots, n$
        \begin{flalign*}
            %
            &\vy_{i} = (\vxi_{i} - \vx) / \sigma, \quad \text{where } \vxi_{i} \sim \NN (\vx, \sigma^2 \vC)&\\
            &f_{i} = f(\vxi_i)&
        \end{flalign*}
        
        \State \textbf{Selection and recombination}
        \Statex Sort $\vy_{i}$ by $f_{i}$ in ascending order, for $i = 1, \dots, n$
        \begin{flalign*}
            &\cma = \sigma \sum_{i=1}^m w_i \vy_{i} &&& \text{where } \sum_{i=1}^m w_i = 1,\text{ and }\forall i \in \{1, \dots, m\}.~ w_i > 0 &\\
            &\vx \leftarrow \vx + \epsilon~\cma&
        \end{flalign*}

        \State \textbf{Cumulative step-size adaptation}
        \begin{flalign*}
            &\vp_{\sigma} \leftarrow (1 - \alpha_{\sigma}) \vp_{\sigma} + \sqrt{\alpha_\sigma (2 - \alpha_\sigma)~ m_{\text{eff}}}~\vC^{-\frac{1}{2}} \cma / \sigma, &&& \text{where }m_{\text{eff}} = (\tsum_{i=1}^m w_i^2)^{-1}&\\
            &\sigma \leftarrow \sigma \times \exp\left(\tfrac{\alpha_{\sigma}}{d_{\sigma}} \left(\tfrac{\lVert \vp_{\sigma}\rVert}{\EEE \lVert \NN (0, \Id) \rVert} - 1\right)\right)&
        \end{flalign*}

        \State \textbf{Covariance matrix adaptation}
        \begin{flalign*}
            &h_\sigma = \mathds{1}\left(\tfrac{\lVert \vp_{\sigma} \rVert}{\sqrt{1 - (1-\alpha_\sigma)^{2(t+1)}}} < (1.4 + \tfrac{2}{d+1})\EEE \lVert \NN (0, \Id) \rVert)\right)&\\ 
            &\bar{w}_j = w_j \times \left( 1 \text{ if } w_j \geq 0 \text{ else } d / \lVert \vC^{-\frac{1}{2}} \vy_j \rVert^2 \right)&\\ 
            &d(h_\sigma) = \mathds{1}\left( \alpha_c (1 - h_\sigma)(2-\alpha_c) \leq 1 \right)&\\ 
            &\vp_c \leftarrow (1 - \alpha_c) \vp_c + h_{\sigma} \sqrt{\alpha_c (2-\alpha_c) m_{\text{eff}}}~ \cma / \sigma&\\
            &\vC \leftarrow (1 + \alpha_1 d(h_\sigma) - \alpha_1 - \alpha_m \tsum_{i=1}^n w_i)\vC + \alpha_1 \vp_c {\vp_c}^T + \alpha_m \sum_{i=1}^n \bar{w}_i \vy_i \vy_i^T&
        \end{flalign*}
    \end{algorithmic}
\end{algorithm}

\section{Vanilla CMA-ES Algorithm}\label{secAlgSuppl}

The listing in Algorithm \ref{alg:cmaes} displays the update step of vanilla CMA-ES at iteration $t$.
Parallel CMA-ES, on which we base our method, uses the same update, but carries it out $\varrho$ times in parallel, once for each search distribution.
We would like to point out that our notation differs from the standard notation of the ES community.
While the number of sampled proposals in each iteration is classically denoted by $\lambda$ and the number of selected elites by $\mu$ \citep{hansen2016cma}, we deviate from this notation in the present work.
Following the main paper, we denote the number of sampled candidates by $n \in \NNN^+$, and the number of elites by $m$.
In our experiments, we tune the elite population size $m \in \NNN^+$, the initial value for $\sigma$ using a grid search (cf. Appendix \ref{secExpDets}) and use the defaults from the evosax codebase \citep{lange2023evosax} for the remaining parameters.
These hyperparameters should coincide with those of \citet{hansen2016cma}.

\subsection{Computation of the Recombination Weights and Hyperparameters}\label{secHyper}
For completeness, we list the computation of the recombination weights and all remaining hyperparameters in this section.
Each population uses the same recombination weights, so $w_{ik} = w_{jk}$ for populations $i$ and $j$.
For simplicity, we therefore drop the population indices in the weights and define them for the single population case.
For this, we use the equations (49)-(58) from \citet{hansen2016cma} with an \textit{elite} population size of $m$. 
Please note that the aforementioned work uses different notation and denotes the population size by $\lambda$, and the number of elites by $\mu$.

Given population size $n$ and problem dimensionality $d$, we compute the set of weights $w_1, \ldots, w_n$ as follows:
\begin{align}
    &\epsilon = 1\\
    &w'_i = \ln\left(\frac{n + 1}{2}\right) - \ln(i), \quad \text{for } i = 1, \ldots, n\\
    &m_{\text{eff}} = \frac{(\sum_{i=1}^m w'_i)^2}{\sum_{i=1}^m {w'}_i^2} \qquad \text{Note that this is equivalent to the previous definition in } \eqref{eq:mueff}\\
    &m_{\text{eff}}^- = \frac{(\sum_{i=m+1}^n w'_i)^2}{\sum_{i=m+1}^n {w'}_i^2}\\
    &\alpha_\sigma = \frac{m_{\text{eff}} + 2}{d + m_{\text{eff}} + 5}\\
    &d_\sigma = 1 + 2 \max \Bigl( 0, \sqrt{\frac{m_{\text{eff}} - 1}{d+1}} - 1\Bigr) + \alpha_\sigma\\
    &\alpha_c = \frac{4 + m_{\text{eff}} / d}{d + 4 + 2m_{\text{eff}} / d}\\
    &\alpha_1 = \frac{2}{(d + 1.3)^2 + m_{\text{eff}}}\\
    &\alpha_m = \min \Bigl(1 - \alpha_1, 2\frac{1 / 4 + m_{\text{eff}} + 1 / m_{\text{eff}} - 2}{(d + 2)^2 + m_{\text{eff}}} \Bigr)\\
    &\beta_m = 1 + \alpha_1 / \alpha_m\\
    &\beta_{m_{\text{eff}}} = 1 + \frac{2 m_\text{eff}^-}{m_\text{eff} + 2} \\
    &\beta_{\text{pos def}} = \frac{1 - \alpha_1 - \alpha_m}{d \alpha_m}\\
    &w_i = \begin{cases}
        \frac{1}{\sum_{i=1}^n \max(w'_i, 0)} w'_i & \text{if } w'_i \geq 0 \\[2mm]
        \frac{\min(\beta_m, \beta_{m_{\text{eff}}}, \beta_{\text{pos def}})}{\sum_{i=1}^n |\min(w'_i, 0)|} w'_i & \text{else} 
    \end{cases}
\end{align}

\section{Experimental Details}\label{secExpDets}

This section lists the full experimental details for this paper.
The code is partially based on \citet{langosco2021neural} and \citet{lange2023evosax}.
For the experiments including existing ES such as vanilla CMA-ES, we use the implementations in $\texttt{evosax}$ \citep{lange2023evosax}.
All the experiments are performed on an internal cluster with eight NVIDIA A40 GPUs.
The code to reproduce our experiments and plots will be made available upon conference publication.

Each experiment is repeated 10 times using randomly generated seeds.
In each plot, we report the mean performance across all 10 runs and 1.96 standard error bars. 
To compute the maximum mean discrepancy \citep[MMD]{gretton2012kernel} we use the RBF kernel for which we select the bandwidth based on the median distance between the ground truth samples \citep{han2018stein}.
For all experiments, we use the standard RBF kernel $k(\vx, \vy) = \exp(-\lVert \vx - \vy \rVert^2 / 2h)$ for which we find the bandwidth $h$ via grid search.
Following \citet{salimans2017evolution} we use a rank transformation for fitness shaping for all OpenAI-ES-based methods.
For GF-SVGD, we follow \citet{han2018stein} in using a Gaussian prior $\NN(0, \sigma^2 \Id)$, where $\sigma^2$ is determined via grid search.

\subsection{Hyperparameter Tuning}
We tune the hyperparameters for each method separately.
For all methods, this includes the kernel bandwidth.
For SVGD, GF-SVGD and SV-OpenAI-ES, we additionally tune the Adam learning rate, while the initial step-size $\sigma$ for SV-CMA-ES is selected analogously.
For GF-SVGD, we follow \citet{han2018stein} in using a Gaussian prior centered at the origin, with isotropic covariance.
The scale of the prior covariance is also determined via grid search.
The ranges over which we search are listed in the following subsections in the corresponding hyperparameter paragraph of each experimental subsection.
For SV-OpenAI-ES, we additionally tune the width $\sigma$ of the proposal distribution, and for SV-CMA-ES we select the elite population size $m$ via grid search.
All remaining hyperparameters for the CMA-ES- and OpenAI-ES-based algorithms are chosen to be the defaults from evosax.
The ranges over which we search the hyperparameters differs across the problems due to their different characteristics.
We list the specific ranges in the following subsections.
The full list of hyperparameters can be found in Table \ref{table:hyperparams}.

\begin{table}[h]
\centering
\caption{Full Hyperparameter Overview}\label{table:hyperparams}
\begin{tabular}{l|l|ll|lll|lll|lll}
\toprule
\textbf{Task} & \textbf{Dim.} & \multicolumn{2}{c|}{\textbf{SVGD}} & \multicolumn{3}{c|}{\textbf{GF-SVGD}} & \multicolumn{3}{c|}{\textbf{SV-CMA-ES}} & \multicolumn{3}{c}{\textbf{SV-OpenAI-ES}} \\
\textit{Hyperparameter} & & \multicolumn{1}{c}{$\epsilon$} & \multicolumn{1}{c|}{$h$} & \multicolumn{1}{c}{$\epsilon$} & \multicolumn{1}{c}{$h$} & \multicolumn{1}{c|}{$\sigma^2$} & \multicolumn{1}{c}{$m$} & \multicolumn{1}{c}{$h$} & \multicolumn{1}{c|}{$\sigma^2$} & \multicolumn{1}{c}{$\epsilon$} & \multicolumn{1}{c}{$h$} & \multicolumn{1}{c}{$\sigma^2$} \\
\midrule
Gaussian Mixture & 2& 0.05 & 0.223 & 1.0 & 0.889 & 2.72 & 2 & 0.889 & 0.50 & 0.50 & 0.001 & 0.10 \\
Double Banana & 2 & 1.0 & $10^{-4}$ & 0.001 & 0.011 & 1.116 & 2 & 0.011 & 0.5 & 0.001 & $10^{-4}$ & 0.15 \\
Motion Planning & 10 & 0.01 & 0.01 & 0.001 & 0.67 & 2.67 & 2 & 0.01 & 0.10 & 0.05 & 0.01 & 0.10 \\
Credit & 22 & 0.1 & $10^{-3}$ & 0.01 & 0.67 & 3.34 & 9 & $10^{-3}$ & 0.35 & 0.005 & $10^{-3}$ & 0.05 \\
Covtype & 55 & 0.01 & 0.45 & 0.05 & 1.0 & 2.28 & 12 & 0.78 & 0.15 & 0.01 & 0.334 & 0.2 \\
Spam & 58 & 0.01 & 0.11 & 0.05 & 1.0 & 0.56 & 9 & 0.45 & 0.20 & 0.01 & 0.11 & 0.2 \\
Pendulum & 353 & -- & -- & 0.05 & 16.67 & 0.34 & 2 & 3.33 & 0.47 & 0.10 & 30.0 & 0.05 \\
CartPole & 386 & -- & -- & 0.10 & 13.33 & 0.45 & 3 & 30.0 & 0.89 & 1.0 & 30.0 & 1.0 \\
MountainCar & 337 & -- & -- & 0.05 & 23.33 & 0.56 & 2 & 30.0 & 0.68 & 1.0 & 30.0 & 0.68 \\
Halfcheetah & 678 & -- & -- & 0.05 & 6.67 & 0.01 & 5 & 16.67 & 0.68 & 0.05 & 30.0 & 0.05 \\
Hopper & 515 & -- & -- & 0.10 & 16.67 & 0.01 & 5 & 3.33 & 0.05 & 0.10 & 30.0 & 0.26 \\
Walker & 662 & -- & -- & 0.10 & 10.0 & 0.01 & 8 & 10.0 & 0.79 & 0.05 & 30.0 & 0.16 \\
\bottomrule
\end{tabular}
\footnotesize{
\begin{tabbing}
\hspace*{.5cm}\= \textbf{SVGD:} \= $\epsilon$ is the Adam learning rate, $h$ is the kernel bandwidth. \\
\hspace*{.5cm}\= \textbf{GF-SVGD:} \= $\epsilon$ is the Adam learning rate, $h$ is the bandwidth, $\sigma^2$ is the scale of the prior covariance. \\
\hspace*{.5cm}\= \textbf{SV-CMA-ES:} \= $m$ is the number of elites, $\sigma^2$ is the init.\ step-size, $h$ is the bandwidth. \\
\hspace*{.5cm}\= \textbf{SV-OpenAI-ES:} \= $\epsilon$ is the Adam learning rate, $\sigma^2$ is the step-size, $h$ is the bandwidth.
\end{tabbing}
}
\end{table}

\subsection{Sampling from Synthetic Densities}

\paragraph{Gaussian Mixture}
We construct a Gaussian Mixture by uniformly sampling $N$ modes $\vmu_i$ over the fixed interval of $[-6, 6]^2$.
Furthermore, we sample the associated weights $w_i \in \RRR$ uniformly over $[0, 10]$. The resulting density is given by:
$$p(\vx) = \frac{1}{K} \sum_{i=1}^N w_i \NN (\vx; \vmu_i, \Id), \qquad K = \tsum_{i=1}^N w_i.$$
In our setting, we set $N = 4$. 
We note that in our experiments one of the sampled weights is close to zero, which is why Fig.\ \ref{fig:samples} depicts only three modes.

\paragraph{Double Banana}
We use the double banana density that was introduced by \citet{detommaso2018stein}:
\begin{gather*}
    p(\vx) \propto \exp \Bigl( -\frac{\lVert x \rVert_2^2}{2 \sigma_1} - \frac{(y - F(\vx))^2}{2 \sigma_2} \Bigr),\\
    \text{where } \vx = [x_1, x_2] \in \RRR^2, \qquad F(\vx) = \log((1-x_1)^2 + 100(x_2-x_1^2)^2).
\end{gather*}
We choose the same hyperparameters as \citet{wang2019matrix}, which are $y = \log (30), \sigma_1 = 1.0, \sigma_2 = 0.09$.

\paragraph{Motion Planning}
The motion planning problem that we sample from was introduced by \citet{barcelos2024path}.
The goal is to sample from the density over $N$ waypoints that induce a trajectory from a predefined start point $t_0$ to a target $t_T$.
The resulting density defines a cost that interpolates obstacle avoidance and smoothness penalties:
\begin{align*}
    &p(\vx) \propto \exp \Bigl( \sum_{t=1}^{W} p_{\text{collision}} (\vx_t) + \alpha \lVert \vx_t - \vx_{t-1} \rVert_2 \Bigr),\\
    \text{where } &p_{\text{collision}}(\vx) = \frac{1}{K} \sum_{i=1}^K \NN (\vx; \vmu_i, \sigma^2 \Id).
\end{align*}
The distribution $p_{\text{collision}}$ induces a randomized 2D terrain for navigation, with the valleys between modes having lower cost. 
We sample a total number of 15 obstacles $\vmu_i \in \RRR^2$ from a Halton sequence following the procedure of \citet{barcelos2024path}.
We choose the hyperparameters $\sigma = 0.25$ and choose $N = 5$.
This results in an optimization problem over $\RRR^{10}$, as each waypoint is two-dimensional.

\paragraph{Ground Truth Samples}
For each task, we sample 256 ground truth samples to compute the MMD.
For the Gaussian Mixtures, samples can be drawn by sampling from the closed form Gaussian mixture.
For the other two problems, ground truth samples are drawn with the help of standard Metropolis-Hastings MCMC.
For this, we sample from one independent chain per sample, using $100{,}000$ burn-in steps.

\paragraph{Hyperparameter Tuning} We carefully tune the hyperparameters of each method by carrying out a grid search over the following hyperparameter ranges.
For the Adam learning rate, we tune over the values proposed by \citet{wang2019matrix}: $[0.001, 0.005, 0.01, 0.05, 0.1, 0.5, 1.0]$.
For the kernel parameter $h$, we use 10 equidistant values from the intervals $[0.001, 1]$, $[0.0001, 0.1]$, $[0.001, 3.0]$ for the Gaussian mixture, double banana and motion planning tasks respectively.
For the elite ratio hyperparameter in SV-CMA-ES, we use 10 equidistant values from the interval $[0.15, 0.5]$, and for the initial value of $\sigma$ that is used in SV-CMA-ES and SV-OpenAI-ES, we use 10 equidistant values from the interval $[0.05, 0.5]$.
For GF-SVGD $\sigma$ refers to the prior covariance scaling which we search over 10 equidistant values from the intervals $[0.1, 6.0]$, $[0.01, 2.0]$, and $[0.01, 4.0]$ for the Gaussian mixture, double banana and motion planning tasks respectively.

\begin{table}[ht]
\centering
\caption{Sampling Task Hyperparameters}
\begin{tabular}{ll|ll}
\toprule
\textbf{Hyperparameter} & \textbf{Value} & \textbf{Hyperparameter} & \textbf{Value} \\
\midrule
Number of iterations & 1000 & Total Pop.\ Size & 400 \\
Number of seeds & 10 & Number of Subpop.\ & 100 \\
\bottomrule
\end{tabular}
\end{table}

\subsection{Logistic Regression}

\paragraph{Datasets}
We follow the implementation of \citet{langosco2021neural}, which uses a hierarchical prior $p(\theta)$ on the parameters $\theta = [\alpha, \beta]$, where $\beta \sim \NN (0, \alpha^{-1})$ and $\alpha \sim \Gamma (a_0, b_0)$.
Following the aforementioned work, we use $a_0 = 1$ and $b_0 = 0.01$.
Given data $D$, the task is to approximate samples from the posterior
\begin{align*}
    &p(\theta \mid D) = p(D \mid \theta)p(\theta)\\
    \intertext{with}
    &p(D \mid \theta) = \prod_{i=1}^N \bigl[ y_i \tfrac{\exp (x_i^T \beta)}{1 + \exp (x_i^T \beta)} + (1- y_i) \tfrac{\exp (-x_i^T \beta)}{1 + \exp (-x_i^T \beta)} \bigr].
\end{align*}
We use three datasets from the UCI Machine Learning repository \citep{asuncion2007uci} in our experiments: Covtype \citep{blackard1998comparison}, Spambase \citep{hopkins1999spambase}, and Credit score data \citep{data23}.
For the Covtype dataset, we create a binary version following \citet{liu2016stein}, which groups the samples into the two groups: \textit{Lodgepole Pine} versus \textit{rest}.
For each task, we partition the data into 70\% for training, 10\% for validation and 20\% for testing and use a batch size of 128.
The dimensionalities of the problems are defined by the number of features in the respective datasets, with an additional degree of freedom for the parameter $\alpha$ of the prior.

\paragraph{Hyperparameter Tuning}
We tune the hyperparameters of each method by carrying out a grid search over the following hyperparameter ranges on the validation set. 
For the Adam learning rate, elite ratio, and initial $\sigma$ values for ES, we tune over the same parameter ranges as for the toy densities.
For the kernel parameter $h$, we use 10 equidistant values from the interval $[0.001, 1.0]$ and for the prior $\sigma$ in GF-SVGD we use the interval $[0.01, 5.0]$.

\begin{table}[h]
\centering
\caption{Logistic Regression Task Hyperparameters}
\begin{tabular}{ll|ll}
\toprule
\textbf{Hyperparameter} & \textbf{Value} & \textbf{Hyperparameter} & \textbf{Value} \\
\midrule
Number of iterations & 1000 & Total Pop.\ Size & 256 \\
Number of seeds & 10 & Number of Subpop.\ & 8 \\
Batch size & 128 & -- & -- \\
\bottomrule
\end{tabular}
\end{table}

\subsection{Reinforcement Learning}

\paragraph{Setup}
The goal of each RL task is to maximize the expected return.
The corresponding objective is to sample from the following posterior:
\begin{equation*}
    p(\theta) \propto \exp(J(\theta)), \quad J(\theta) = \EEE_{(s_t, a_t) \sim \pi_{\theta}} \bigl[\sum_{t=1}^T r(s_t, a_t)\bigr].
\end{equation*}
Here, we estimate the expected return via MC approximation across 16 independent policy rollouts for each iteration at which we evaluate the current particle set performance.
In our experiments, we report the best expected return across all particles.
Every such experiment is repeated 10 times, as with all previous tasks.
We use the standard maximum episode length for all classic tasks as specified in the gym environment implementation, except for MountainCar, where we use a maximum episode length of 500 (instead of the default 1000 steps) to make the task more difficult.
For the brax control tasks, we use a maximum episode length of 1000 steps.
For the classic tasks Pendulum, CartPole and MountainCar, we use the $\texttt{gymnax}$ implementations \citep{lange2022gymnax}, and for the remaining environments, we use the $\texttt{brax-v1}$ implementations \citep{brax2021github}.
We use the continuous version of MountainCar.
Please note that we use the brax legacy dynamics in our experiments.

\paragraph{Hyperparameter Tuning}
We tune the hyperparameters of each method by carrying out a grid search over the following hyperparameter ranges. 
For the Adam learning rate, elite ratio, and initial $\sigma$ values, we tune over the same parameter ranges as for the toy densities.
For the kernel parameter $h$, we use 10 equidistant values from the interval $[0.001, 30.0]$ and for the prior $\sigma$ in GF-SVGD we use the interval $[0.01, 1.0]$.
For brax environments, we tune the hyperparameters for 500 generations and report results for running the selected configurations over 10 seeds for 1000 generations.
Please note that while we use 5 seeds for hyperparameter tuning, all evaluation runs and plots in this paper use 10 runs as indicated.

\begin{table}[ht]
\centering
\caption{RL Task Hyperparameters}
\begin{tabular}{ll|ll}
\toprule
\textbf{Hyperparameter} & \textbf{Value} & \textbf{Hyperparameter} & \textbf{Value} \\
\midrule
Number of iterations & 200 (classic) / 1000 (brax) & Total Pop.\ Size & 64 \\
Number of seeds & 10 & Number of Subpop.\ & 4 \\
MLP Layers & 2 & MC Evaluations & 16 \\
Hidden Units & 16 & Hidden Activation & ReLU \\
-- & -- & Output Activation & Tanh \\
\bottomrule
\end{tabular}
\end{table}

\section{Additional Results}\label{secSupplRes}

\begin{figure}[t]
    \centering
    \begin{minipage}{0.48\linewidth}
        \centering
        \textbf{CMA-ES-based methods}\\
        \subfloat[Pendulum]{
            \includegraphics[width=.31\linewidth]{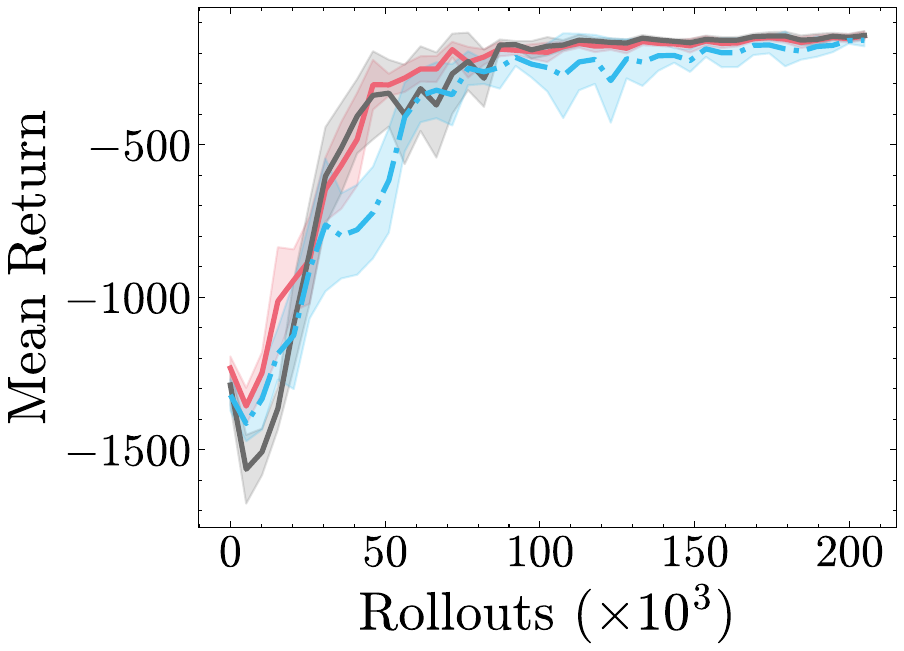}
        }
        \subfloat[CartPole]{
            \includegraphics[width=.3\linewidth]{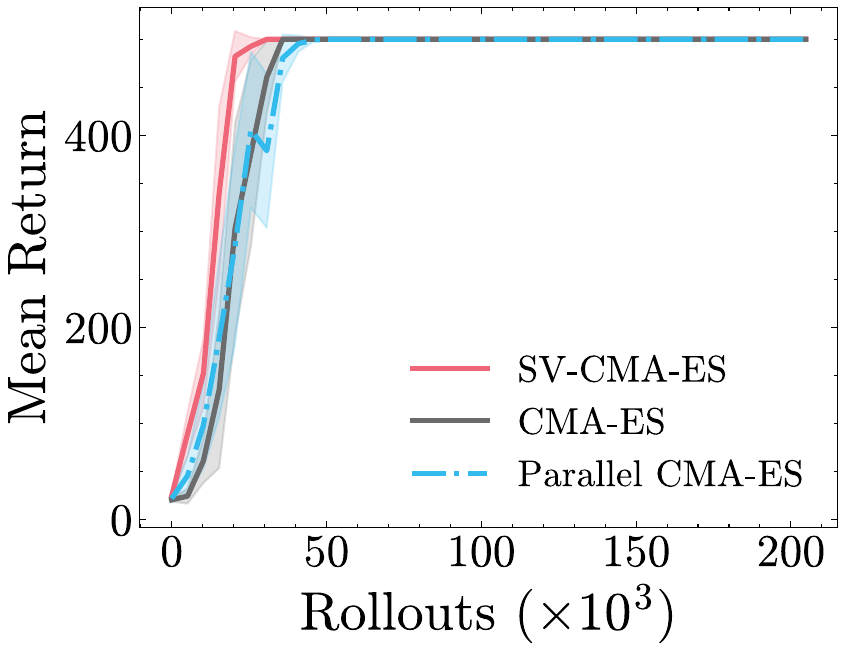}
        }
        \subfloat[MountainCar]{
            \includegraphics[width=.3\linewidth]{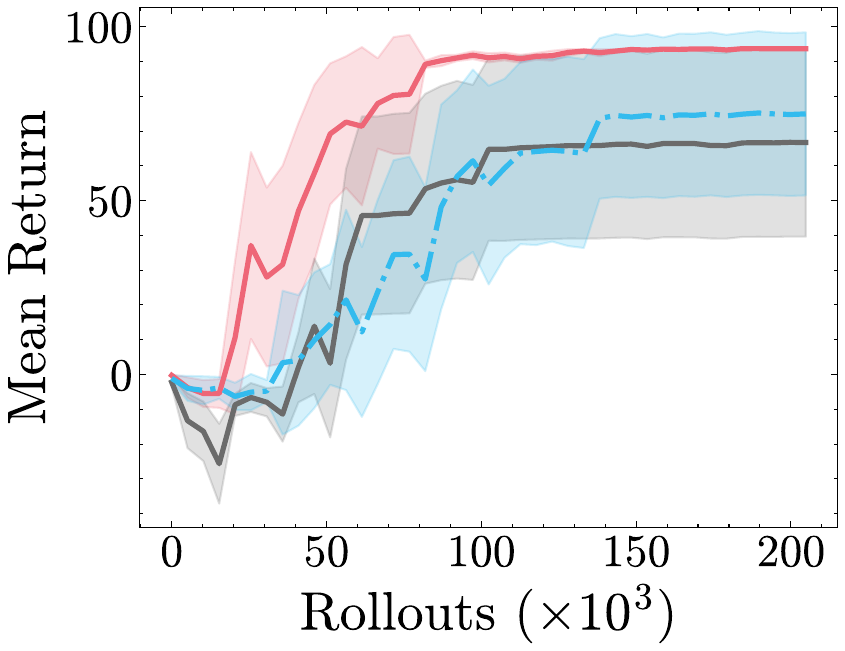}
        }\\
        \subfloat[Halfcheetah]{
            \includegraphics[width=.3\linewidth]{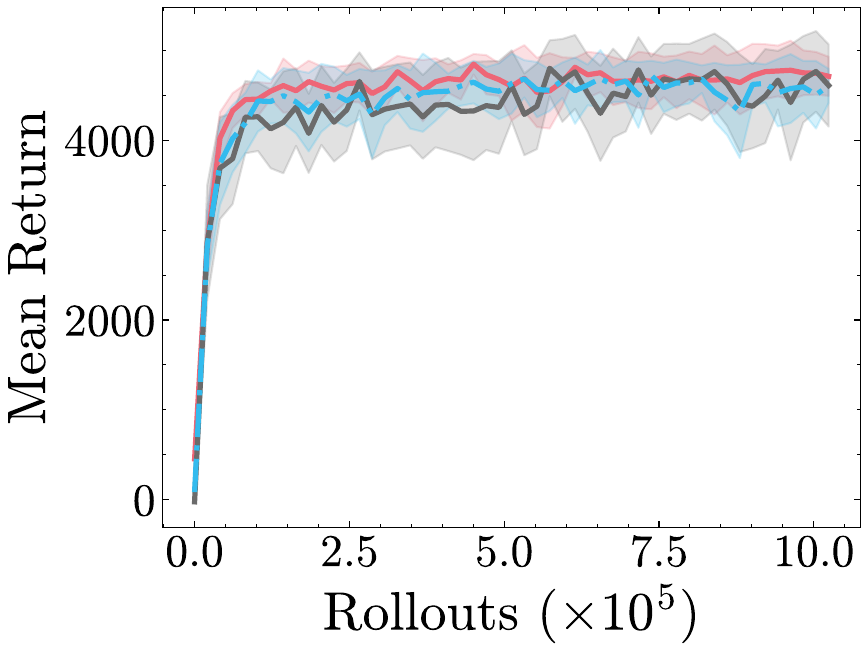}
        }
        \subfloat[Hopper]{
            \includegraphics[width=.3\linewidth]{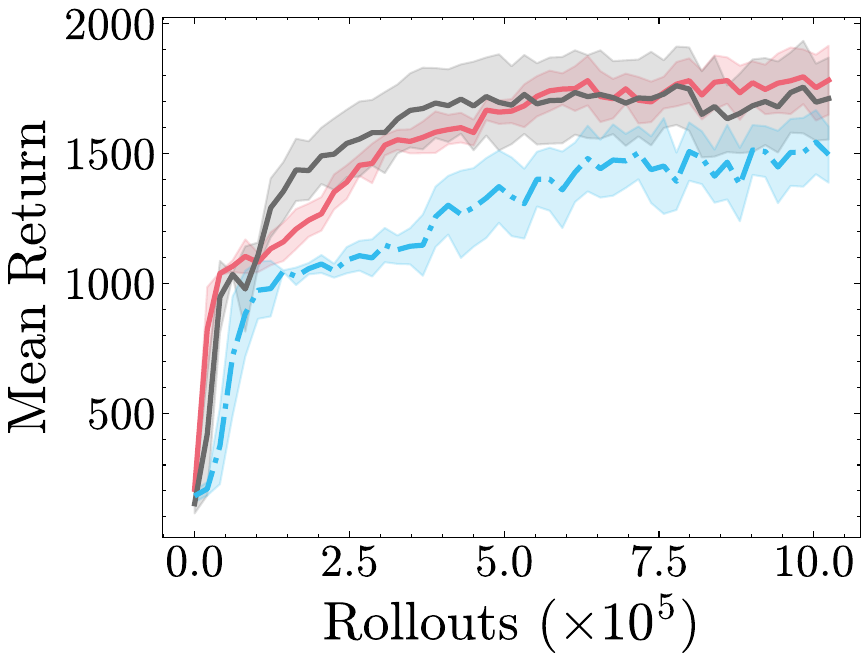}
        }
        \subfloat[Walker]{
            \includegraphics[width=.3\linewidth]{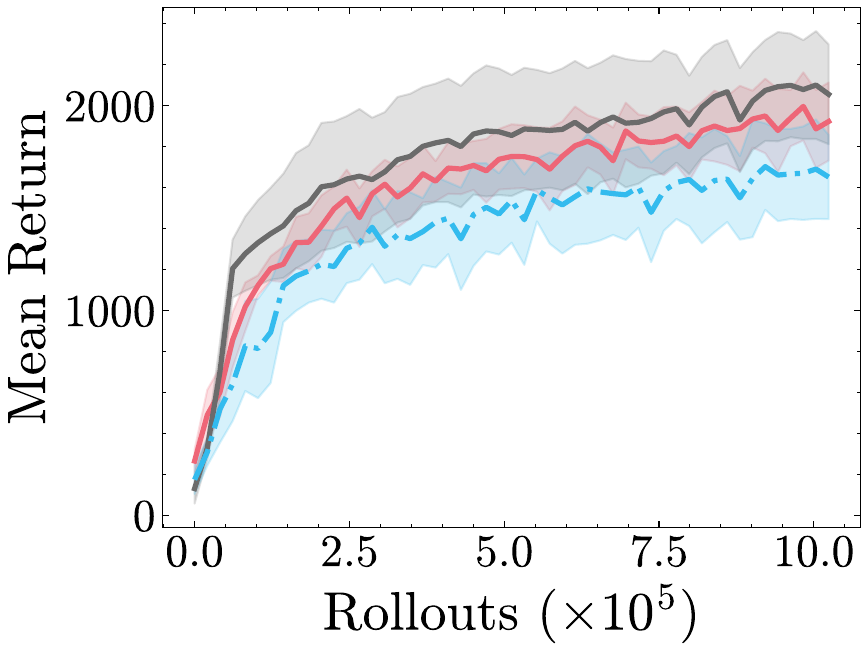}
        }
    \end{minipage}
    \hfill
    \begin{minipage}{0.48\linewidth}
        \centering
        \textbf{OpenAI-ES-based methods}\\
        \subfloat[Pendulum]{
            \includegraphics[width=.3\linewidth]{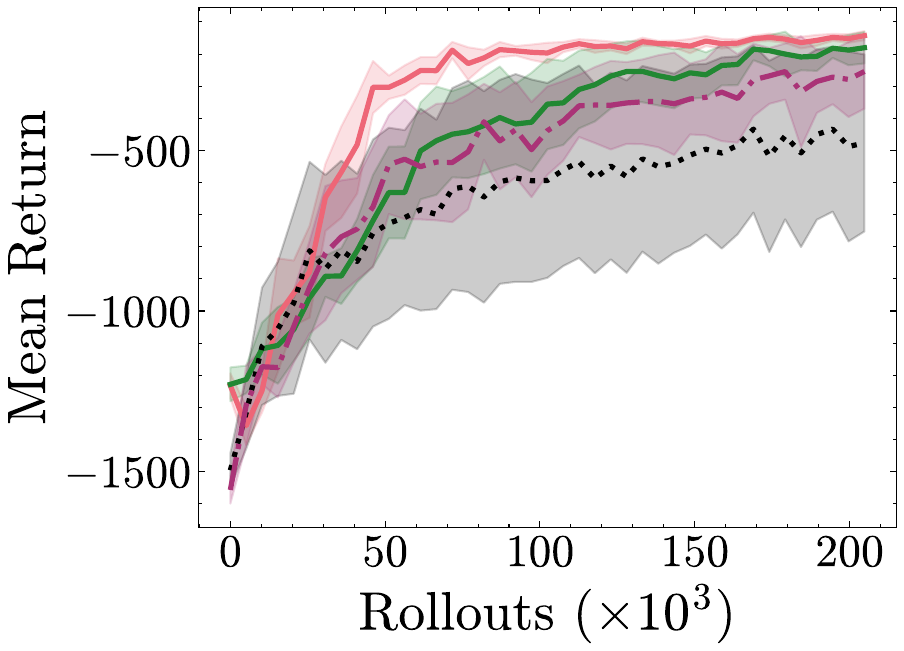}
        }
        \subfloat[CartPole]{
            \includegraphics[width=.3\linewidth]{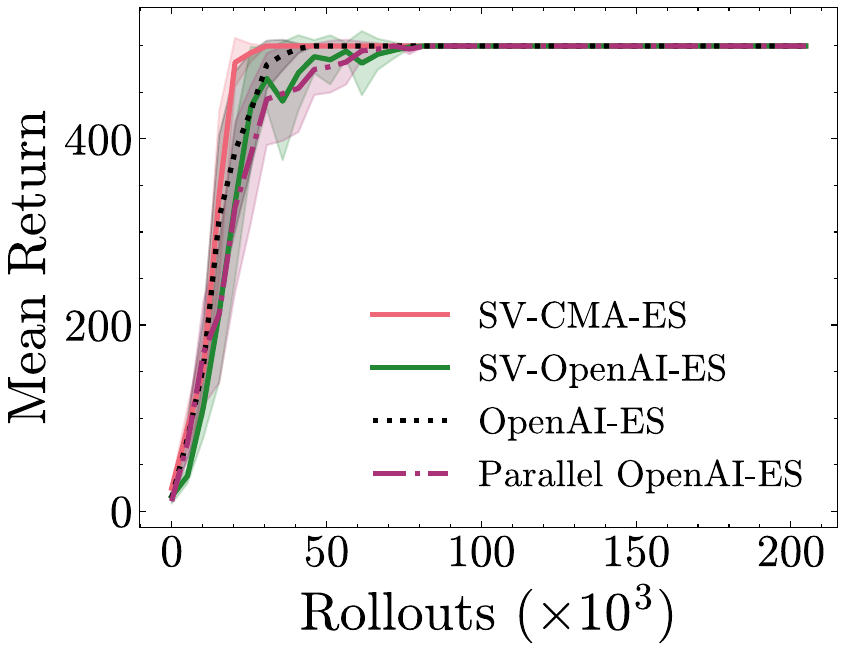}
        }
        \subfloat[MountainCar]{
            \includegraphics[width=.3\linewidth]{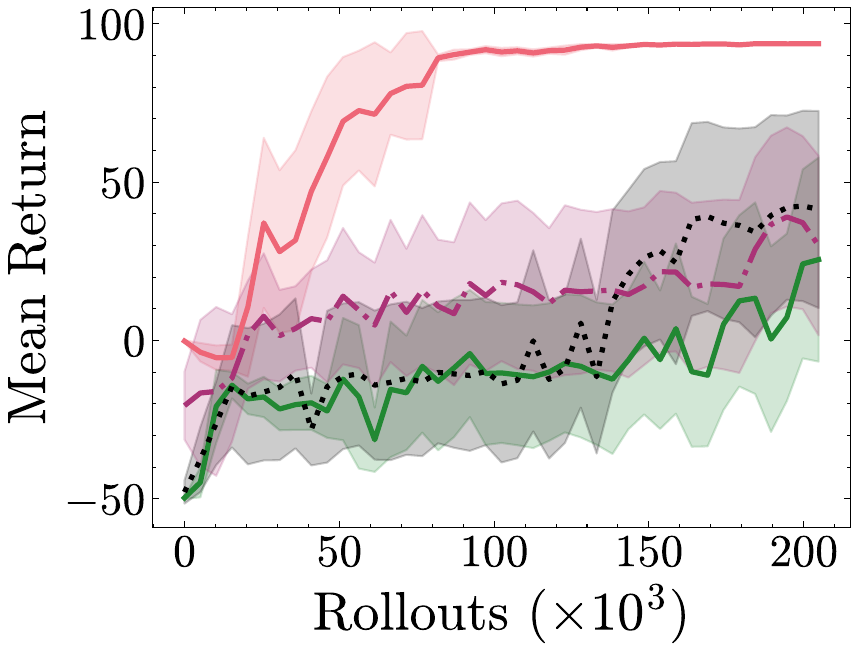}
        }\\
        \subfloat[Halfcheetah]{
            \includegraphics[width=.3\linewidth]{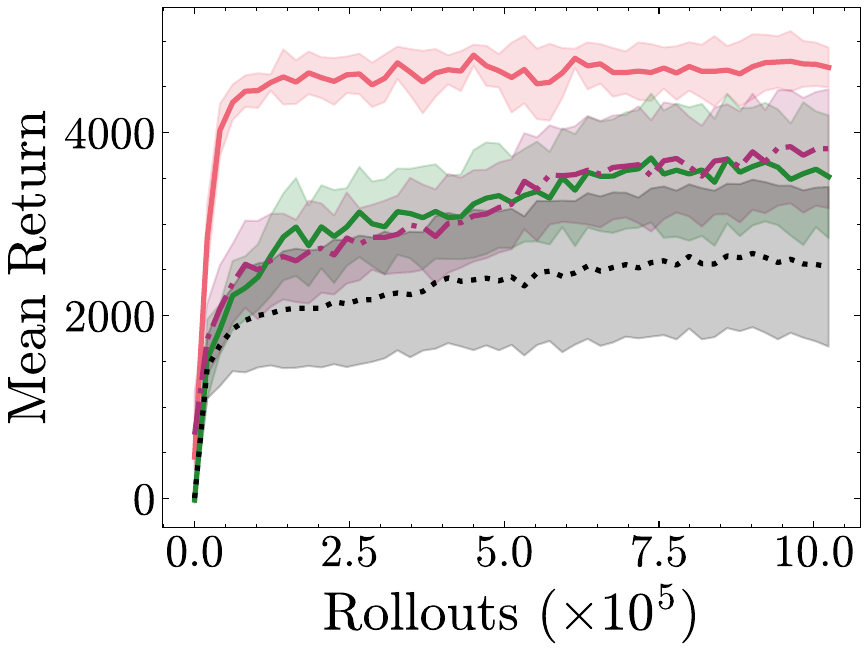}
        }
        \subfloat[Hopper]{
            \includegraphics[width=.3\linewidth]{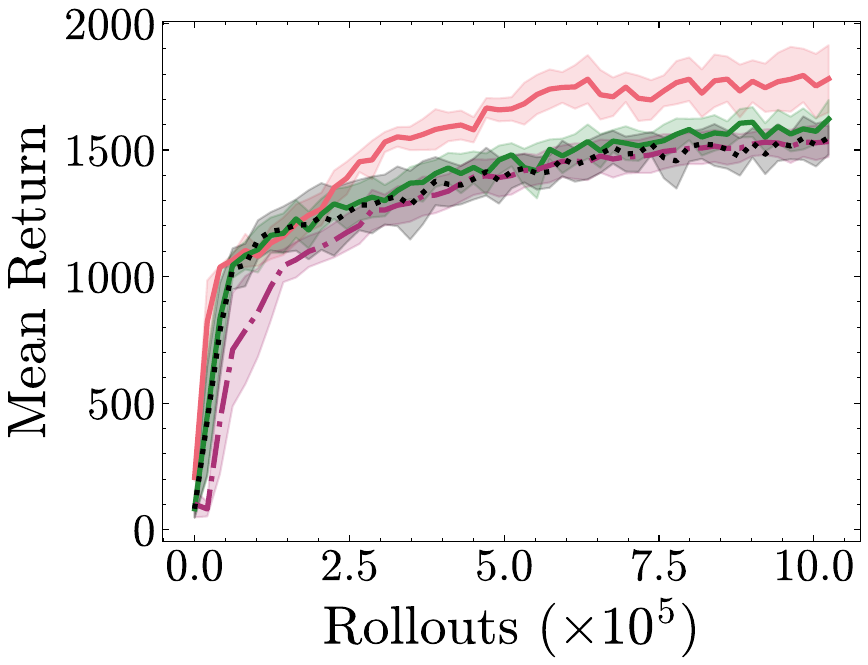}
        }
        \subfloat[Walker]{
            \includegraphics[width=.3\linewidth]{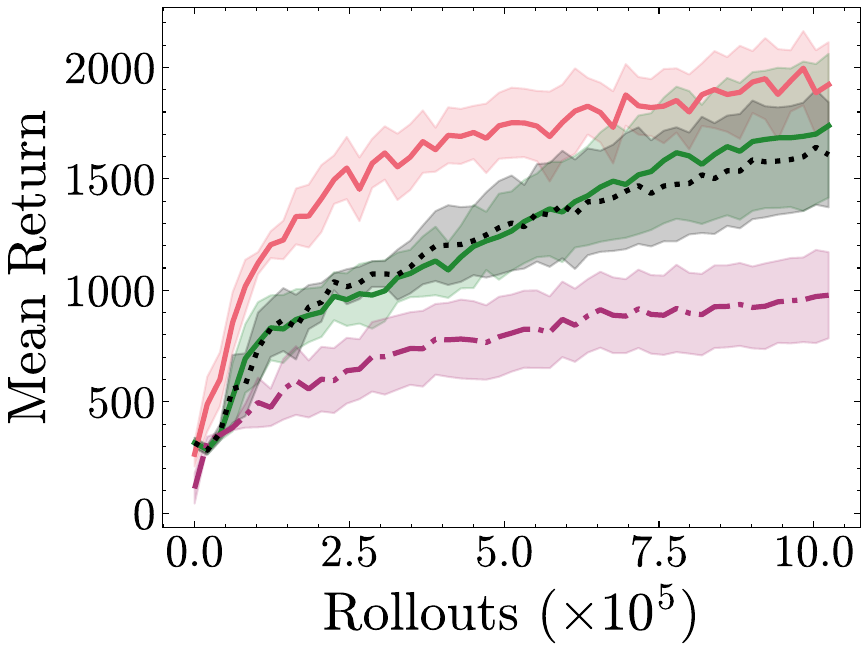}
        }
    \end{minipage}
    
    \caption{Comparison of CMA-ES-based methods (left) and OpenAI-ES-based methods (right). In all experiments, a total population size of 64 is used, split across 4 subpopulations in parallel methods. We report the mean ($\pm 1.96$ standard error) across 10 independent runs.}
    \label{fig:rl-baselines}
\end{figure}

\subsection{Algorithm Ablations}\label{secAblations}

\paragraph{Setting} To evaluate the potential of SV-CMA-ES for pure blackbox optimization, we compare it to other ES-based methods that perform pure maximum-likelihood estimation.
The two natural baselines to compare against are first simple CMA-ES, but with a population size of $n = \text{Num.\ particles} \times \text{Subpopsize}$, which ablates the effect of sampling multiple subpopulations in parallel.
Second, we compare against multiple independent parallel runs of CMA-ES, which ablates the coordination of the runs via the SVGD update.

\paragraph{Results} In Fig.\ \ref{fig:rl-baselines}, we display the results of our evaluation.
When comparing SV-CMA-ES against single population CMA-ES we observe similar performances on most problems. 
The most notable difference that stands out is the better performance of SV-CMA-ES on the MountainCar problem.
What happens in this experiment?
The MountainCar problem is an MDP in which a car must be accelerated to reach a goal on a hill.
Since there are negative rewards for large accelerations, there is a large local optimum at which policies are idle and do not accelerate at all to avoid negative rewards.
However, once the goal is reached, the agent receives a reward of 100.
We speculate due to these results that SV-CMA-ES may be superior to other CMA-ES-based methods on such sparse reward environments.
We will further investigate this hypothesis in Appendix \ref{secEnvAbl}.

Furthermore, we compare the performance to multiple parallel CMA-ES runs.
This experiment essentially ablates the kernel term in our update.
We find that in comparison to multiple parallel CMA-ES runs, our method achieves superior performance on half of the problems, and equal on the rest.
These results are encouraging, as they suggest that our SV-ES is a viable alternative to independent parallelizations of a given ES algorithm.

\begin{wrapfigure}{R}{0.45\textwidth}
    \centering
    \includegraphics[width=\linewidth]{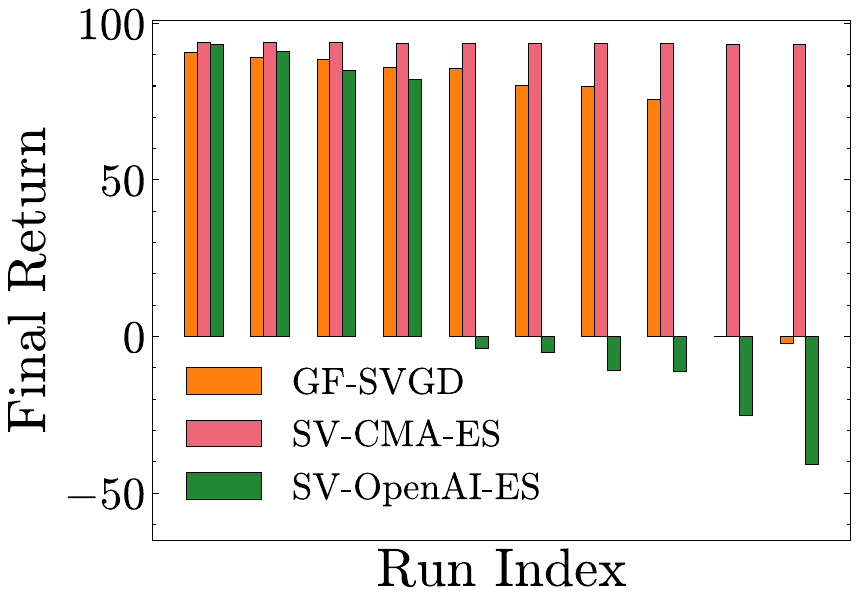}
    \caption{Per-seed performances on the MountainCar task. SV-CMA-ES is the only method to achieve optimal performance across all seeds. GF-SVGD converges to idle agents (i.e., a reward of zero) on two out of the ten seeds.}\label{fig:mc_seeds}
\end{wrapfigure}

Additionally, we include Fig.~\ref{fig:mc_seeds} to display the per-seed results on the MountainCar problem.
The results highlight that GF-SVGD converges to the local optimum of being idle on two out of the ten runs.

\subsection{Environment Ablations}\label{secEnvAbl}

\paragraph{Setting} To investigate the hypothesis that SV-CMA-ES is superior to simple parallel CMA-ES runs or single large-population CMA-ES in sparse reward environments, we construct sparse reward versions of the Walker and Hopper environments following the procedure of \citet{mazoure2019leveraging}.
The goal of the environment ablations is to make the optimization landscape more similar to that of MountainCar.
To achieve this goal, we increase the control cost in the reward and remove the positive rewards for the agent being healthy.
Additionally, the original forward reward is replaced by a positive reward for being upright that is only awarded once the agent has moved beyond a certain position in space.
Essentially, the resulting environments only reward successful locomotion policies once the behavior is learned, instead of rewarding intermediate learning outcomes as well.
In this setting, the reward is negative for policies that move the agent not far enough and close to zero for policies that do not move the agent at all, similar to the MountainCar reward.
The resulting reward function is:
\begin{align*}
    r_{\text{ablated}}(s_t, a_t) &= \alpha \mathds{1} (pos(s_t) \geq 1) - \beta \lVert a_t \rVert_2^2,
\end{align*}
where $pos(s_t) \in \RRR$ is the agent's position at state $s_t$ in the task space.
In our experiments, we choose $\alpha = 2.0$ and $\beta = 0.1$.
Since the resulting ablated Walker problem is significantly more difficult, we run experiments on this benchmark for 1500 instead of 1000 iterations on this problem.

\paragraph{Results} 
Our results demonstrate that SV-CMA-ES outperforms other CMA-ES-based methods considerably in the constructed scenarios.
Both problems are more difficult to solve than the standard problems in Fig.\ \ref{fig:rl-baselines} because a positive reward is only received once the agent moves far enough.
As hypothesized, we observe a similar pattern as for MountainCar.
In other words, SV-CMA-ES outperforms the other methods on sparse reward environments.
These results suggest two things.
First, we find that the coordination of multiple CMA-ES runs via the SVGD step improves the algorithmic performance on problems for which exploration is crucial.
Second, we observe that this effect only comes to fruition if the population updates are coordinated via SVGD, as SV-CMA-ES outperforms parallel CMA-ES runs considerably.
The combination of these findings suggests that SV-CMA-ES constitutes a strong alternative to other CMA-ES-based optimizers if problems require exploration.
We thus believe that analysis of the scalability of the algorithm to a high number of particles beyond our computational capabilities (i.e., several thousands) would be an intriguing direction of future research.

\begin{figure}[hb]
    \centering
    \subfloat[Hopper-sparse]{
        \includegraphics[width=.3\linewidth]{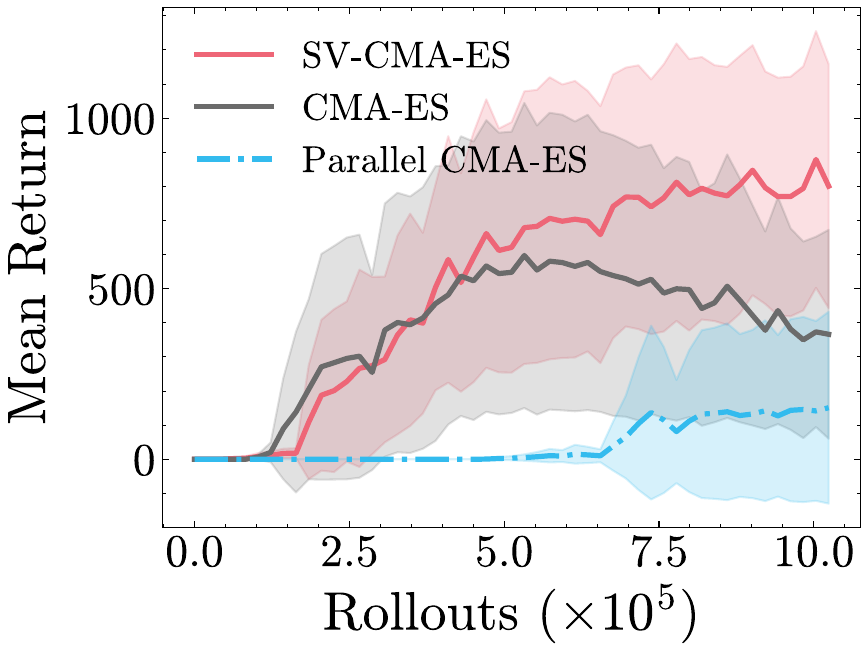}
    }
    \hspace*{5mm}
    \subfloat[Walker-sparse]{
        \includegraphics[width=.3\linewidth]{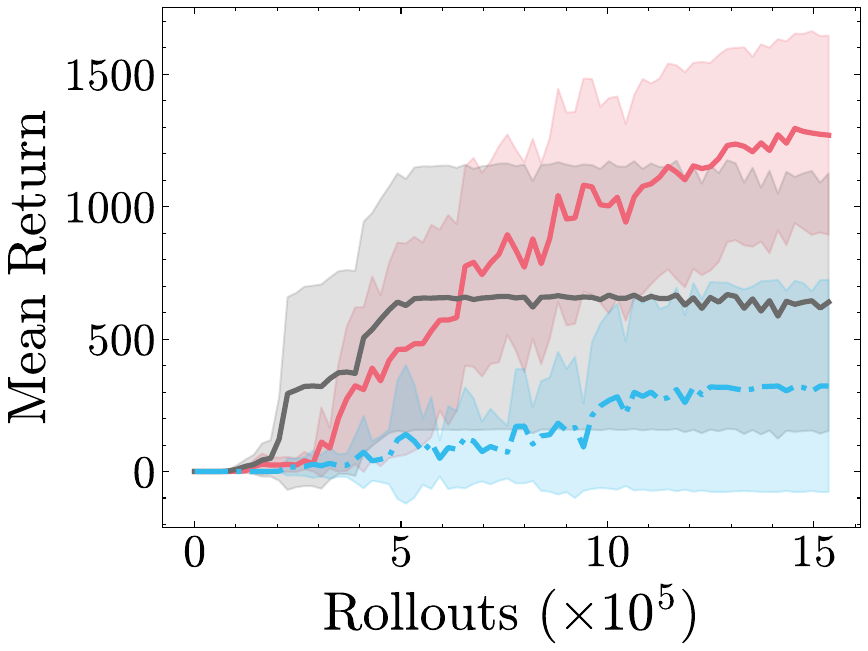}
    }
    \caption{Comparison of CMA-ES-based methods on sparse reward environments. Coordinating parallel ES runs via the SVGD update clearly improves the performance. In all experiments, a total population size of 64 is used, split across 4 subpopulations in parallel methods. We report the mean ($\pm 1.96$ standard error) across 10 independent runs.
    }
    \label{fig:rl-ablated}
\end{figure}

\subsection{Empirical Runtime Analysis}\label{secRuntime}

\paragraph{Setting} Given the higher theoretical runtime complexity of SV-CMA-ES compared to the baselines, we perform an empirical runtime analysis.
The goal of this investigation is to evaluate the significance of the theoretical algorithm properties in practice.
To this end, we select multiple problems from the main paper to illustrate the performances over time.
We select the synthetic Gaussian mixture as a low dimensional and fast-to-evaluate objective, a logistic regression tasks as moderately dimensional tasks with fast-to-evaluate objectives, and the brax RL tasks as the most high-dimensional task of the paper which require expensive simulator calls for evaluations of the ground truth density. 
For each problem, we run 1000 iterations, just as in prior experiments. 
For the RL tasks, we use the GPU accelerated brax simulator.
To provide a fair analysis of wall-clock times, we disable deterministic operations on the GPU, which may lead to slight differences in performances, \wrt to the results reported in the main paper, but is more realistic in terms of runtimes.
We would like to point out, however, that we did not observe any visible differences in the results.
The experiment is carried out on a single NVIDIA A40 GPU and reported times are for all 10 independent runs being run in parallel.

\begin{figure*}[ht]
    \centering
    \subfloat[Gaussian Mixture]{
        \includegraphics[width=.25\linewidth]{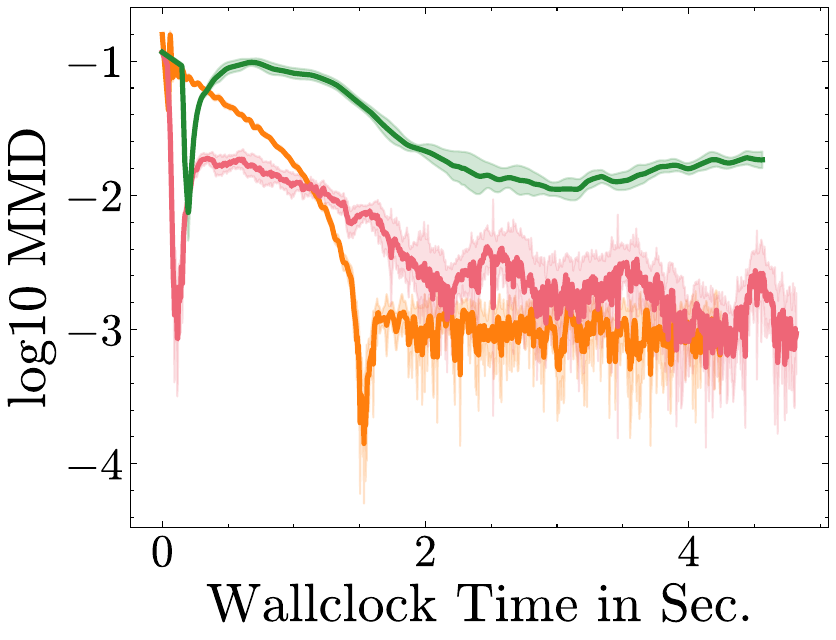}
    }\hspace*{5mm}
    \subfloat[Credit Log.\ Reg.]{
        \includegraphics[width=.25\linewidth]{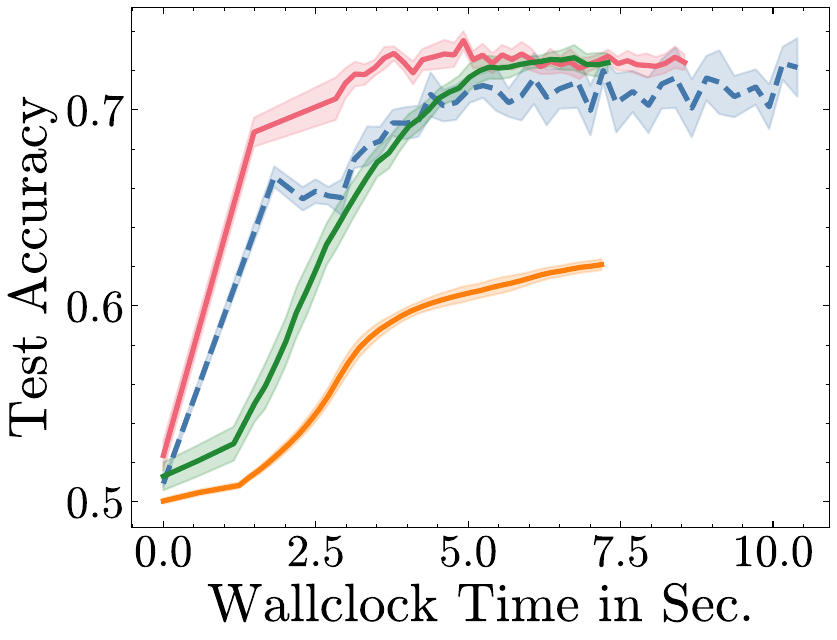}
    }\hspace*{5mm}
    \subfloat[Covtype Log.\ Reg.]{
        \includegraphics[width=.25\linewidth]{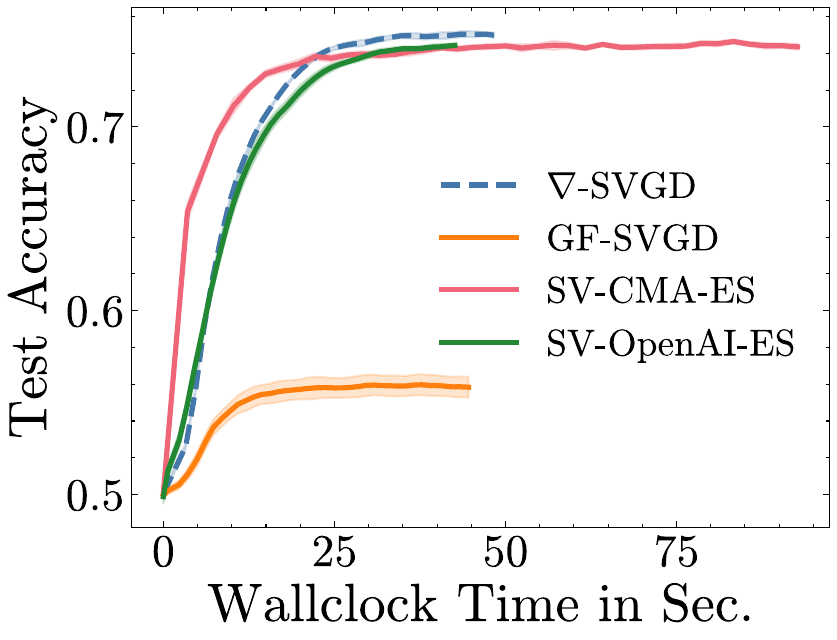}
    }\\
    \subfloat[Halfcheetah RL]{
        \includegraphics[width=.25\linewidth]{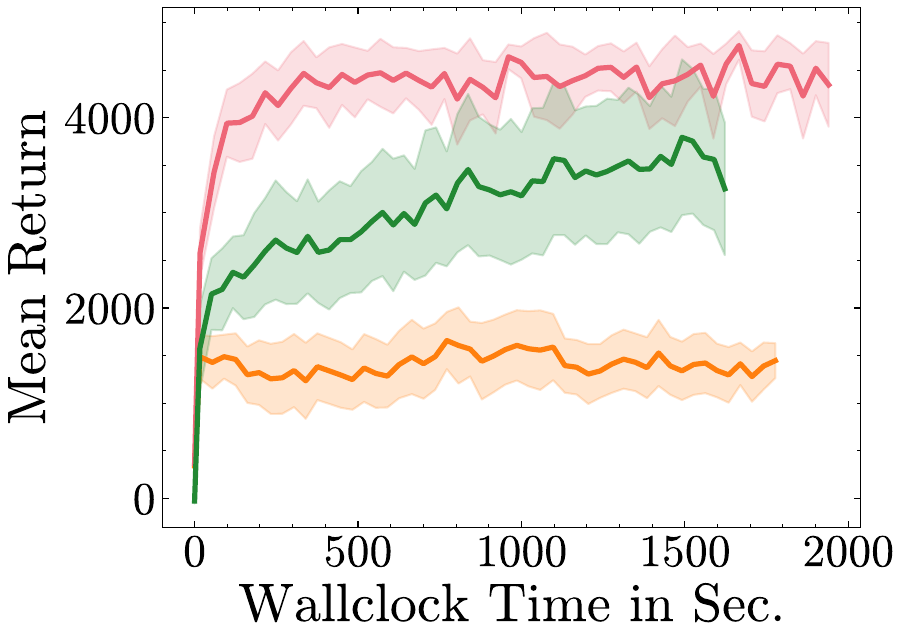}
    }\hspace*{5mm}
    \subfloat[Hopper RL]{
        \includegraphics[width=.25\linewidth]{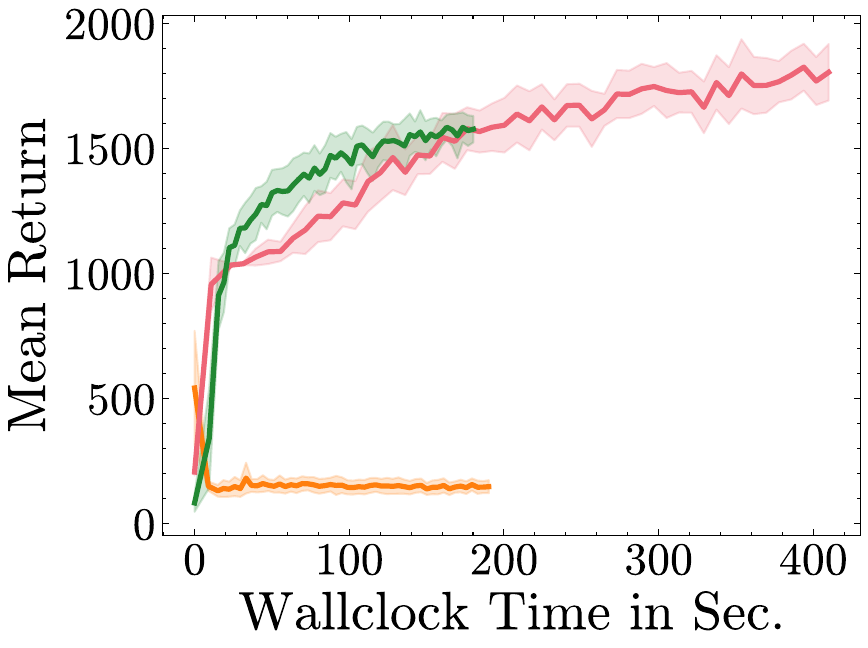}
    }\hspace*{5mm}
    \subfloat[Walker RL]{
        \includegraphics[width=.25\linewidth]{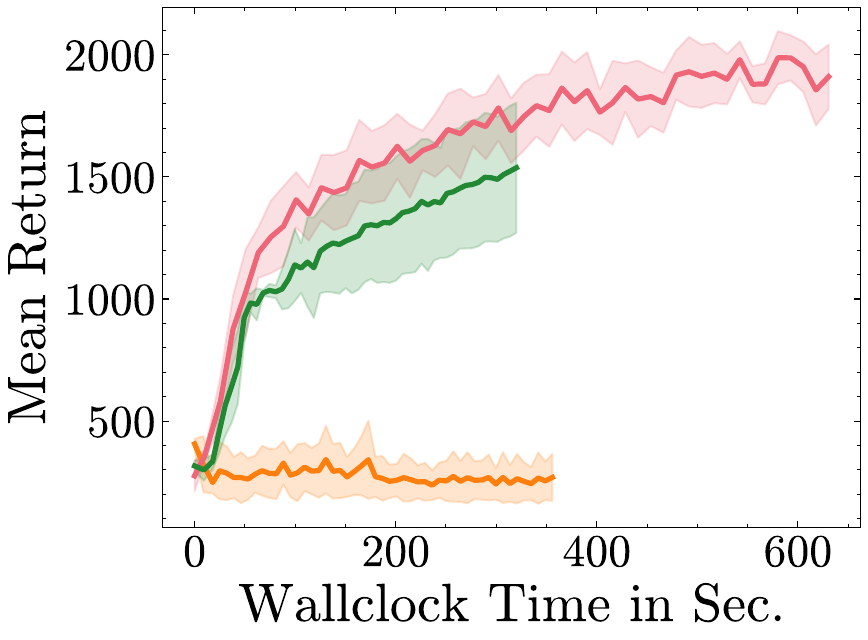}
    }\hfill
    \caption{Performance vs.\ wallclock time. We run all methods for 1000 iterations and display the elapsed wallclock time. The plot shows that SV-CMA-ES also performs well \wrt this metric.
    All results are averaged across 10 independent runs ($\pm 1.96$ standard error).
    }
    \label{fig:timing}
\end{figure*}

\paragraph{Results} The results of the runtime analysis are depicted in Fig.\ \ref{fig:timing}.
Overall, we observe competitive convergence behavior of SV-CMA-ES in terms of wall-clock time. 
While the overall runtime time required to run 1000 iterations is highest for SV-CMA-ES, we do observe that our method obtains good samples efficiently under this metric.

Generally, we observe that the influence of the update computation costs on the overall runtime varies greatly between problems.
While the differences between all methods are relatively small on the synthetic 2d problems, we observe that SV-CMA-ES is slower on high dimensional problems such as the Covtype logistic regression task.
This reflects that the computational cost of SV-CMA-ES increases with the number of problem dimensions, while the cost of the other baselines is dominated by the number of particles.
However, we also observe that the quality of the updates of SV-CMA-ES is still generally better in these cases, as it samples well performing solutions the fastest. 

More importantly, we observe that the main factor that drives overall runtime is the evaluation of the objective function.
Furthermore, we observe smaller runtime differences between the methods on the Halfcheetah task despite it being the most high dimensional task that we evaluate on.
This underlines that the significance of theoretical algorithm complexity depends heavily on the problem and the costs that are associated with target density evaluations. 
Since our method requires fewer iterations than the baselines to reach comparable performance (as shown previously), we find that this demonstrates the practical relevance of the presented approach.

\begin{figure}[ht]
    \centering
    \subfloat[Gaussian Mixture]{
        \includegraphics[width=.3\linewidth]{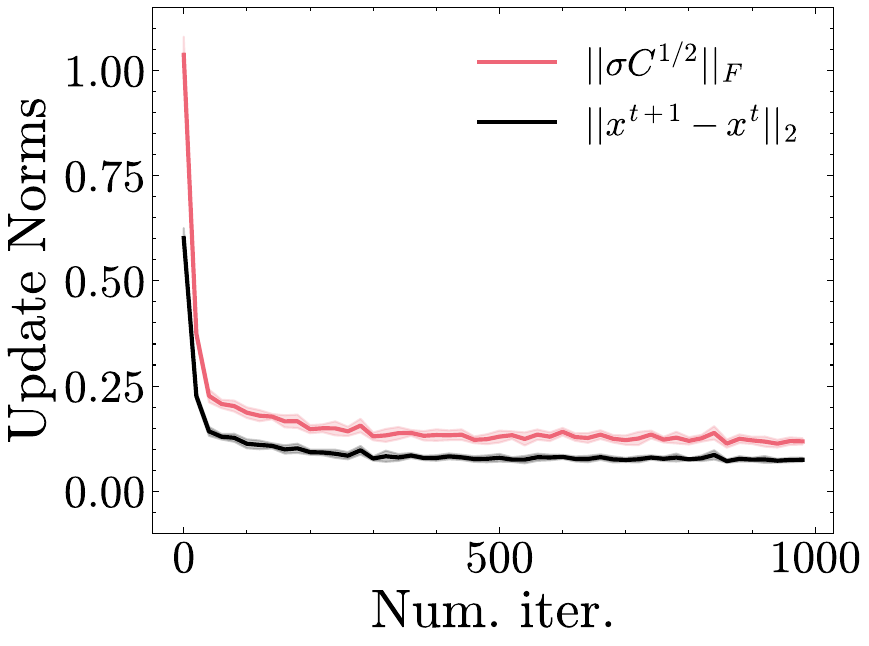}
    }
    \hfill
    \subfloat[Double Banana]{
        \includegraphics[width=.3\linewidth]{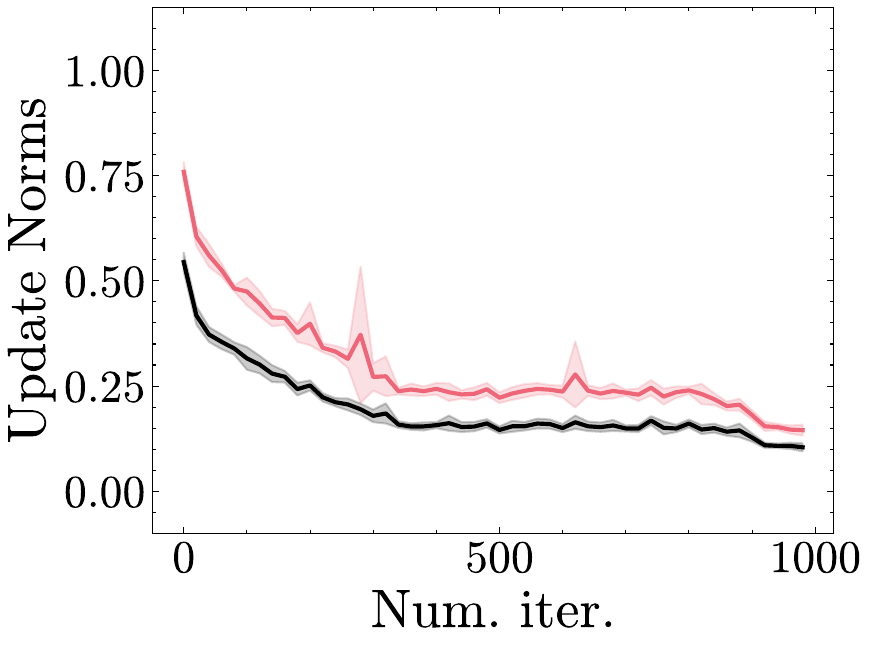}
    }
    \hfill
    \subfloat[Motion Planning]{
        \includegraphics[width=.3\linewidth]{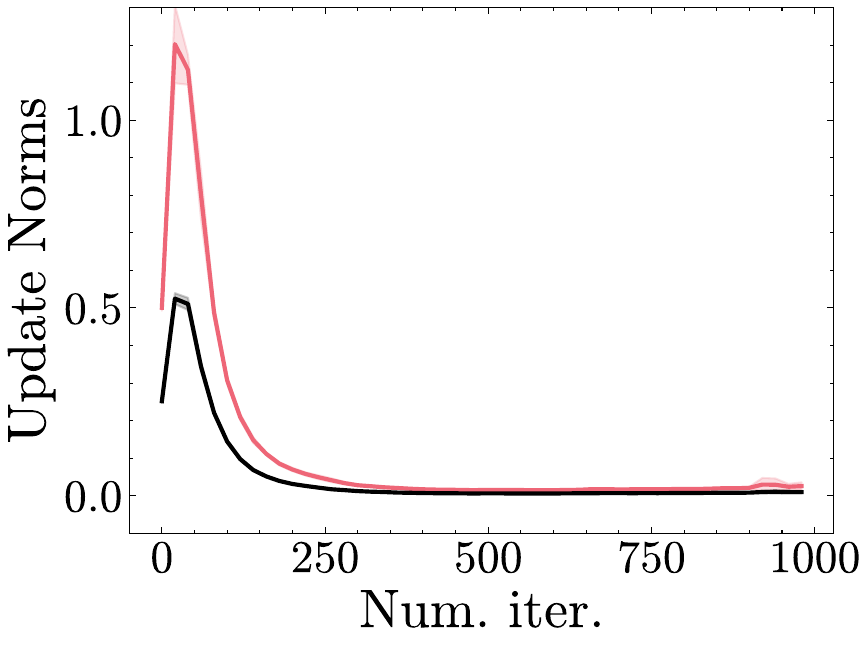}
    }
    \caption{Mean step lengths per iteration. We record the step length every 20 steps across $1\,000$ iterations. 
    The SV-CMA-ES steps and estimated gradient steps by CMA-ES converge to a stable equilibrium. We report the mean ($\pm 1.96$ standard error) across all 100 particles and 10 independent runs.
    }
    \label{fig:convergence}
\end{figure}

\subsection{Empirical Convergence Analysis}\label{sec:app_convergence}
To evaluate the convergence of our method, we expand our analysis beyond the pure MMD computations in Fig. \ref{fig:mmd} and analyze the length of the particle steps.
To this end, we use the same setting as the one in Fig. \ref{fig:mmd} to plot the Frobenius norm of the particle preconditioning matrix $\sigma C^{1/2}$ across iterations in Fig. \ref{fig:convergence}.
In the same figure, we further plot the length of the steps that were actually taken by the particles, i.e., the size of the update in Eq.\ \eqref{eq:asv-cma-final}.
The results show that the algorithm exhibits stable convergence.
A key finding in this context is that for each problem, there exists a stationary point at which the CMA-ES steps that are sampled from $\NN(\vx, \sigma^2 C)$ are counteracted by the kernel gradients, as the total step length is shorter than the matrix norm.
This shows that our method is capable of finding the point of convergence of SVGD where the kernel gradient and objective gradient balance each other out.
We see the fact that the resulting total step length does not reach zero across all problems as an artifact of the stochastic gradient estimates that CMA-ES provides.
We do not see this as a problem, however, as the MMD analysis in Fig. \ref{fig:mmd} shows that the particles only move within areas of high density.
Since the approximation of the target density in SVGD is based on a sum of delta function which approximates the target as $q(x) = \sum_{\vx_i \in X} \delta (\vx - \vx_i) / \varrho$ it permits small particle movements at the equilibrium between repulsive and driving force in the update.



\subsection{Convergence Plots}\label{sec:app_full}
\begin{figure*}[!b]
    \centering
    \subfloat[$\N$-SVGD]{
        \includegraphics[width=.9\linewidth]{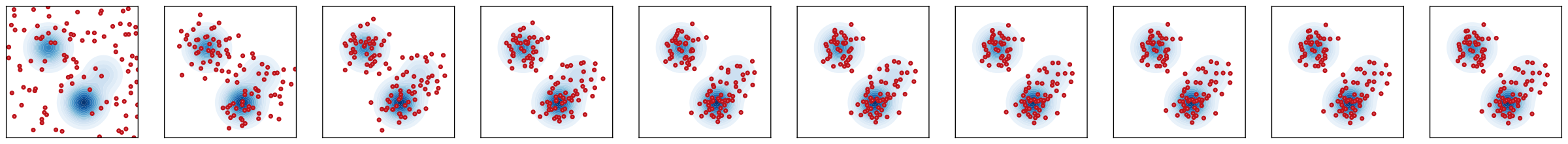}
    }\\
    \subfloat[SV-CMA-ES]{
        \includegraphics[width=.9\linewidth]{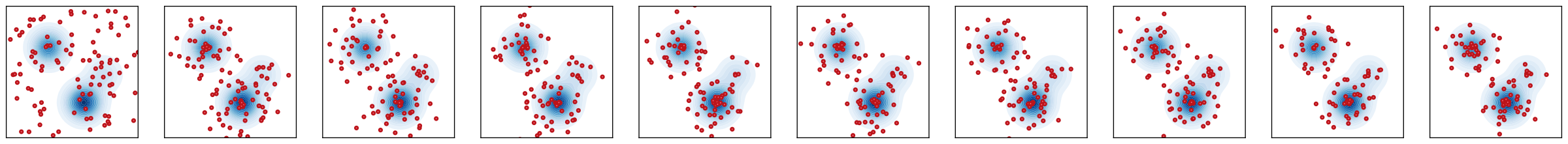}
    }\\
    \subfloat[GF-SVGD]{
        \includegraphics[width=.9\linewidth]{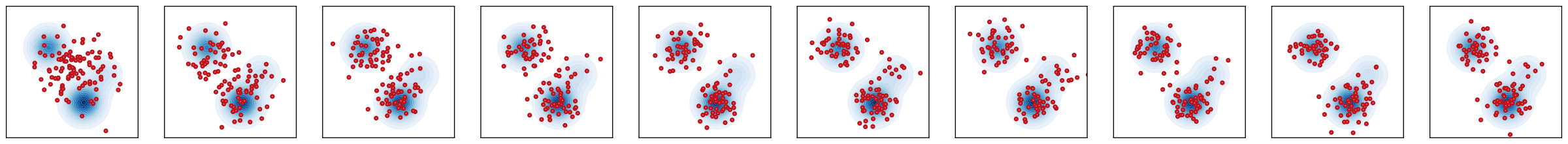}
    }\\
    \subfloat[SV-OpenAI-ES]{
        \includegraphics[width=.9\linewidth]{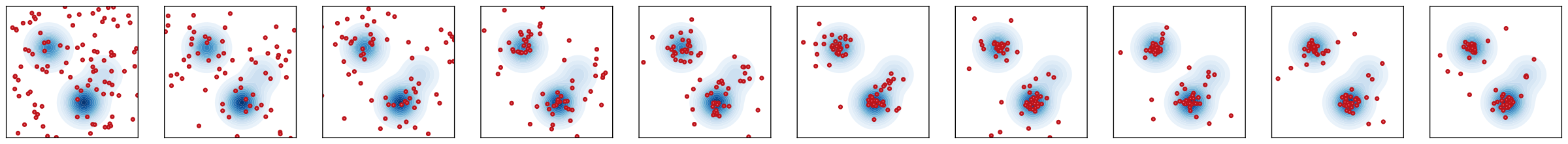}
    }\\
    \begin{tikzpicture}[x=\linewidth/1000, >=stealth, thick, line cap=round, line join=round]
      \draw[->] (0,0) -- (850,0);
      \draw (0, 0.2) -- (0, -0.2) node[below=2pt]{Iteration 1};
      \draw (415, 0.2) -- (415, -0.2) node[below=2pt]{500};
      \draw (850, 0.2) -- (850, -0.2) node[below=2pt]{1000};
    \end{tikzpicture}
    \caption{Convergences on GMM sampling task. For each method, the sample convergence across the full 1000 sampling iterations is displayed. }
    \label{fig:gmm_convergence_iter}
\end{figure*}

\begin{figure*}
    \centering
    \subfloat[$\N$-SVGD]{
        \includegraphics[width=.95\linewidth]{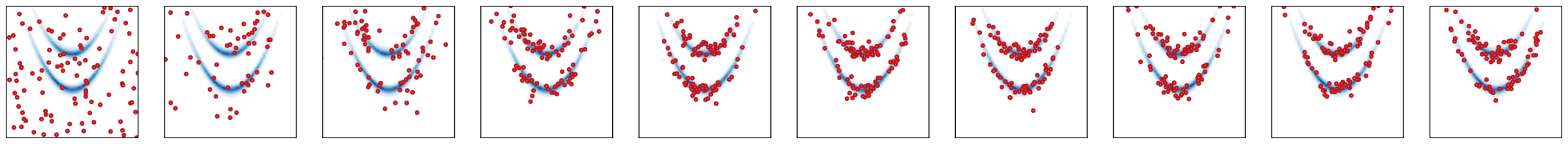}
    }\\
    \subfloat[SV-CMA-ES]{
        \includegraphics[width=.95\linewidth]{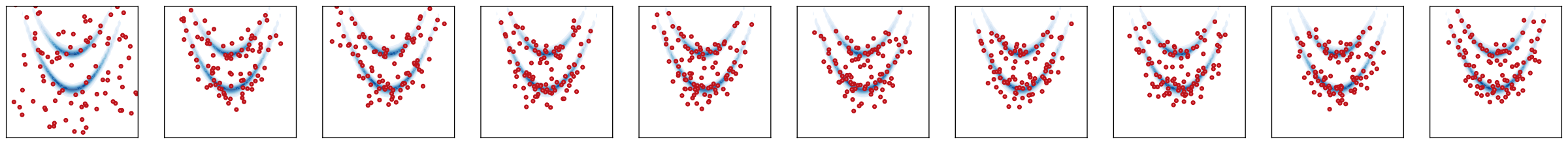}
    }\\
    \subfloat[GF-SVGD]{
        \includegraphics[width=.95\linewidth]{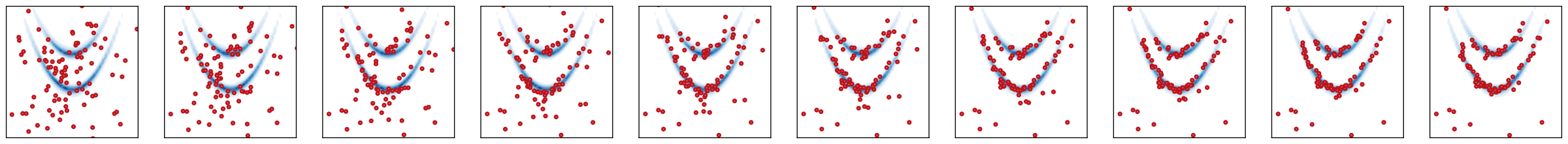}
    }\\
    \subfloat[SV-OpenAI-ES]{
        \includegraphics[width=.95\linewidth]{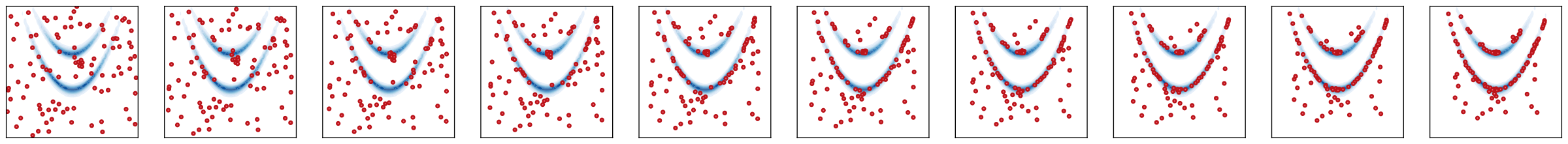}
    }\\
    \begin{tikzpicture}[x=\linewidth/1000, >=stealth, thick, line cap=round, line join=round]
      \draw[->] (0,0) -- (850,0);
      \draw (0, 0.2) -- (0, -0.2) node[below=2pt]{Iteration 1};
      \draw (415, 0.2) -- (415, -0.2) node[below=2pt]{500};
      \draw (850, 0.2) -- (850, -0.2) node[below=2pt]{1000};
    \end{tikzpicture}
    \caption{Convergences on Double banana sampling task. For each method, the sample convergence across the full 1000 sampling iterations is displayed.}
    \label{fig:banana_convergence_iter}
\end{figure*}

\begin{figure*}[t]
    \centering
    \subfloat[$\N$-SVGD]{
        \includegraphics[width=.95\linewidth]{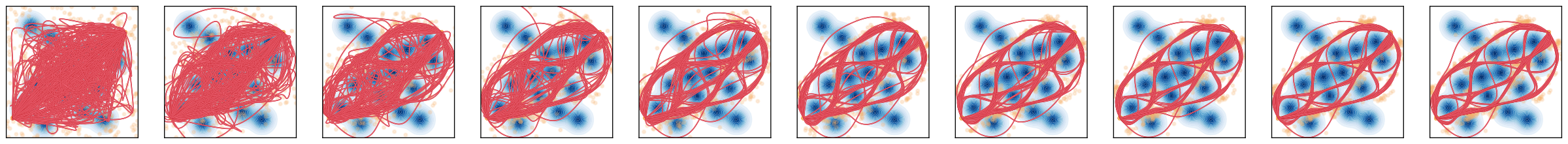}
    }\\
    \subfloat[SV-CMA-ES]{
        \includegraphics[width=.95\linewidth]{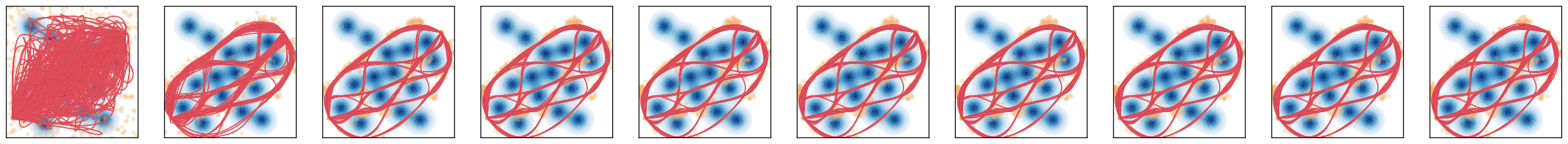}
    }\\
    \subfloat[GF-SVGD]{
        \includegraphics[width=.95\linewidth]{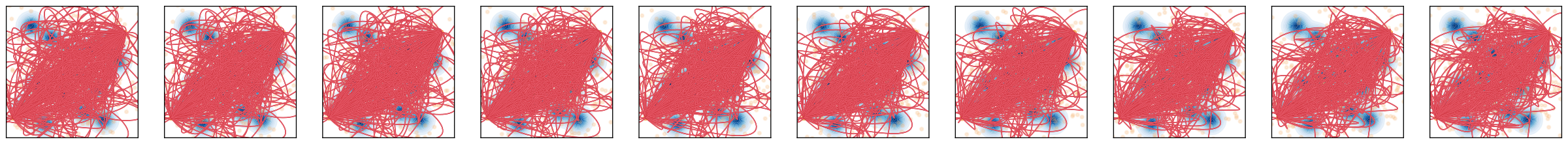}
    }\\
    \subfloat[SV-OpenAI-ES]{
        \includegraphics[width=.95\linewidth]{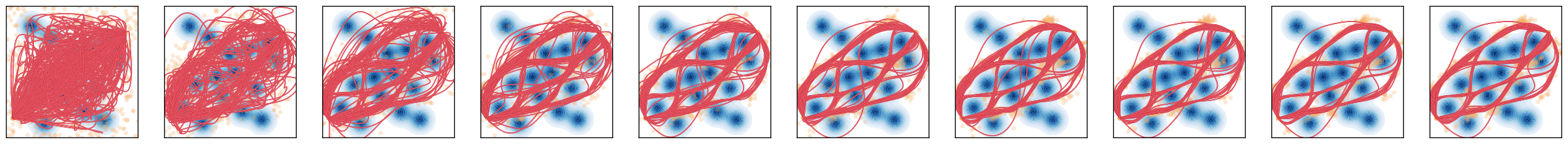}
    }\\
    \begin{tikzpicture}[x=\linewidth/1000, >=stealth, thick, line cap=round, line join=round]
      \draw[->] (0,0) -- (850,0);
      \draw (0, 0.2) -- (0, -0.2) node[below=2pt]{Iteration 1};
      \draw (415, 0.2) -- (415, -0.2) node[below=2pt]{500};
      \draw (850, 0.2) -- (850, -0.2) node[below=2pt]{1000};
    \end{tikzpicture}
    \caption{Convergences on motion planning sampling task. For each method, the sample convergence across the full 1000 sampling iterations is displayed.}
    \label{fig:ramos_convergence_iter}
\end{figure*}

For completeness, we depict the convergence results on the synthetic sampling tasks.
For each task and method combination, we display the evolution of the sample set across the optimization.
To this end, we split the total number of iterations into 10 equally-sized bins and generate a plot for each.
In other words, we plot the sample set every 100 iterations.
The results for the Gaussian mixture are listed in Fig.~\ref{fig:gmm_convergence_iter}, those for the Double Banana density in Fig.~\ref{fig:banana_convergence_iter}, and Fig.~\ref{fig:ramos_convergence_iter} shows the convergence on the trajectory optimization task.
All plots confirm the quantitative results, i.e., SV-CMA-ES converges quickly compared to the baselines.
We note that we refer to convergence once all particles are in high density areas.
Due to the stochastic nature of the update, there is no guarantee that there is an actual stationary distribution of the finite optimization process.
For storage reasons, we include these images as raster images.
We encourage readers to reach out to us in case they are interested in the vector graphics version for these last three plots.

\end{document}